\documentclass{article} % For LaTeX2e
\usepackage{iclr2027_conference,times}

\usepackage{amsmath,amsfonts,bm}

\def\eqref#1{equation~\ref{#1}}
\def\1{\bm{1}}

\DeclareMathAlphabet{\mathsfit}{\encodingdefault}{\sfdefault}{m}{sl}
\SetMathAlphabet{\mathsfit}{bold}{\encodingdefault}{\sfdefault}{bx}{n}

\setcitestyle{citesep={,}}
\usepackage{hyperref}
\usepackage{url}
\usepackage{graphicx}
\usepackage{booktabs}
\usepackage{wrapfig}
\usepackage{capt-of}
\usepackage{longtable}
\usepackage{etoolbox}
\AtBeginEnvironment{itemize}{%
  \settowidth{\leftmargini}{\labelitemi}%
  \addtolength{\leftmargini}{\labelsep}}
\makeatletter
\patchcmd{\LT@makecaption}{\reset@font}{\reset@font\normalsize}{}%
  {\PackageError{paper}{Unable to normalize longtable captions}{Check longtable compatibility.}}
\makeatother
\usepackage{placeins}
\usepackage{needspace}
\usepackage{amsmath}
\usepackage{amssymb}
\usepackage{xcolor}
\DeclareRobustCommand{\opusrevision}[1]{#1}
\DeclareRobustCommand{\astrarevision}[1]{#1}
\hypersetup{hidelinks}
\usepackage{titletoc}

\DeclareRobustCommand{\revised}[1]{#1}
\DeclareRobustCommand{\latestrevision}[1]{#1}
\DeclareRobustCommand{\clarifiedhistory}[1]{#1}
\DeclareRobustCommand{\captionrevision}[1]{#1}
\newsavebox{\appendixtablebox}
\newcommand{\appendixtable}[1]{%
  \sbox{\appendixtablebox}{\footnotesize #1}%
  \ifdim\wd\appendixtablebox>0.90\linewidth
    \resizebox{0.90\linewidth}{!}{\usebox{\appendixtablebox}}%
  \else
    \usebox{\appendixtablebox}%
  \fi}

\renewcommand{\emph}[1]{{\upshape #1}}
\renewcommand{\textit}[1]{{\upshape #1}}

\title{Learn2Play Bench: How Well Do LLM Agents Learn from Experience in Unfamiliar Environments?}

\author{\parbox[t]{\dimexpr\textwidth-2\tabcolsep\relax}{\normalfont\raggedright
\textbf{Yibo~Li}$^{1}$ \enspace \textbf{Jinhang~Qiu}$^{1}$ \enspace \textbf{Zhi~Zheng}$^{1}$ \enspace \textbf{Qianyun~Guo}$^{1}$ \enspace \textbf{Jiaying~Wu}$^{1}$ \enspace \textbf{Shuo~Ji}$^{1}$ \enspace \textbf{Bryan~Hooi}$^{1}$
\\[4pt]
$^{1}$National University of Singapore
\\[2pt]
\texttt{liyibo@u.nus.edu, bhooi@comp.nus.edu.sg}}}

\newcommand{\game}[1]{\textsc{#1}}

\iclrfinalcopy
\begin{document}

\maketitle
\vspace{-10pt}
% Neutral preprint page style: no conference or review banner.
\pagestyle{plain}
\thispagestyle{plain}

% Compact only main-paper vertical glue, preserving fonts and text geometry.
\begingroup
\setlength{\parskip}{4pt}
\begin{abstract}
Learning from experience is essential for LLM agents to adapt to unfamiliar and dynamic environments.
Evaluating this ability is therefore important for understanding how effectively agents acquire and use new knowledge.
Existing benchmarks have sought to evaluate this ability, but they primarily evaluate tasks whose rules are provided in the instructions or already familiar to pretrained models, making it difficult to distinguish learning from interactions from reasoning with existing knowledge.
To address this, we introduce \textbf{Learn2Play Bench}, a benchmark of newly designed text-based games, whose rules are novel or counterintuitive, requiring agents to acquire knowledge through interaction rather than rely solely on pretrained knowledge.
These games provide reproducible feedback and automatic scoring, enabling controlled evaluation of learning across repeated attempts.
We also vary game instances to test whether agents can apply what they have learned to new situations.
Therefore, we evaluate how backbone models, self-evolving methods, and agent harnesses affect agents’ learning ability, revealing three findings: (1) \textbf{Experience retention:} Retaining complete records of actions and feedback can support more effective learning than summarizing these experiences into rules or strategies.
(2) \textbf{Human–agent gap:} Top-performing human players achieve higher peak scores than the evaluated agents. Humans explore more varied strategies, and repeat actions less.
(3) \textbf{Harness matters:} With the backbone fixed, changing the harness can improve performance while reducing estimated inference cost. 
Together, these findings provide insights into how LLM agents learn from experience and suggest directions for future work to improve their learning ability.
Project website: \url{https://liushiliushi.github.io/learn2play-bench-website/}.
\end{abstract}

\iffalse

Reproducible feedback and automatic scoring support controlled evaluation of learning across repeated attempts, while varied game instances test whether acquired knowledge transfers to new situations.
Therefore, 
We evaluate how backbone models, self-evolving methods, and agent harnesses affect agents’ learning ability, revealing three findings:
(1) \textbf{Experience retention:} Retaining complete records of actions and feedback can support more effective learning than summarizing these experiences into rules or strategies.
(2) \textbf{Human–agent gap:} Top-performing human players achieve higher peak scores than the evaluated agents, explore more varied strategies, and repeat actions less.
(3) \textbf{Harness matters:} With the backbone fixed, changing the harness can improve performance while reducing estimated inference cost.
Together, these results provide insights into the roles of experience representation, exploration, and harness design in learning from experience.
Code is available at \url{https://anonymous.4open.science/r/Learn2Play-Bench-F1D8/}.
\fi

\par\vspace{-6pt}
\section{Introduction}
\label{sec:introduction}

The ability of LLM agents to acquire new knowledge through interaction and use it to improve behavior is central to artificial general intelligence (AGI)~\citep{hendrycks2025agi}. While the extensive knowledge acquired during pretraining enables agents to perform complex tasks (e.g., fixing software issues~\citep{yang2024sweagent}, completing web tasks~\citep{zhou2024webarena}, and operating desktop applications~\citep{xie2024osworld}), in realistic deployment,
agents often encounter \textbf{unfamiliar environments} whose rules are not covered by the pretrained knowledge.
For example, an agent struggles to control a robot through an unfamiliar command interface.
It must therefore try commands and observe the resulting motion to learn which joints each command activates.
Agents can then apply what they have learned to make better decisions in subsequent interactions.

% A capable LLM agent should be able to improve when it enters an environment whose rules it does not already know.
% Although pretraining equips agents with broad knowledge for tasks such as software engineering~\citep{yang2024sweagent}, web navigation~\citep{zhou2024webarena}, and desktop interaction~\citep{xie2024osworld}, real deployments can expose agents to \textbf{unfamiliar environments} with new interfaces, conventions, or dynamics.
% Consider a robot controlled through an unfamiliar command interface.
% An agent must try commands, observe the resulting motion, infer which joints they control, and use these discoveries to act more effectively in later attempts.
% The ability to acquire and reuse such knowledge through interaction is an important component of increasingly general agents~\citep{hendrycks2025agi}.

Recent work has therefore developed self-evolving agents that reuse interaction feedback through memories, extracted rules, reasoning patterns, or evolving agent configurations~\citep{zhang2025ace,evotest,ouyang2026reasoningbank}.
However, evaluating whether these systems actually \emph{learn} remains difficult.
Existing methods are often studied with different models, tasks, and evaluation protocols, making their learning behavior hard to compare directly.
More fundamentally, \textbf{better task performance does not necessarily mean that an agent learns more effectively}.
If an agent improves on a familiar task, we do not know whether it learned a new rule from interaction or merely recovered knowledge already encoded during pretraining.

This distinction is often overlooked in existing evaluations of continual or self-evolving agents.
Benchmarks often measure improvement over repeated attempts or task sequences~\citep{evotest,zheng2025lifelong,wei2025evomemory,yan2026agentstream,yang2026pathbench,zhang2025autoenv}, but many tasks are based on familiar real-world settings whose underlying rules may already be represented in pretraining, leaving agents' learning ability insufficiently tested.
We therefore ask:
\begin{center}
\textit{\textbf{How effectively can LLM agents learn from experience in unfamiliar environments?}}
\end{center}

To answer this question, we introduce \textbf{Learn2Play Bench}, a benchmark of newly designed text-based games with unfamiliar or counterintuitive hidden rules.
For example, in one game, stealing a colleague's lunch helps the player get promoted, while working hard does not.
Agents are not given these rules.
They must discover them by acting, observing textual feedback, and carrying useful experience into later episodes, without updating model weights.
Each game provides deterministic feedback and automatic scoring, allowing learning trajectories to be measured reproducibly.
In half of the games, characters, objects, or layouts change across episodes while the underlying rules remain fixed.
This setting tests whether agents can apply knowledge to new situations.

Using this controlled setting, we compare different backbone models, self-evolving methods, and deployed agent harnesses, and contrast their behavior with human players.
Our experiments reveal five main findings:
\textbf{(1) Preserving evidence can matter more than compressing it.}
Raw interaction histories can support stronger learning than methods that summarize experience into rules or strategies, because the original evidence remains available when earlier conclusions need revision (\S4.2).
\textbf{(2) Top-performing human players outperform the best agent.}
Top-performing human players reach higher peak scores than the evaluated agents, while exploring more varied action sequences and recovering more often after setbacks (\S4.4).
\textbf{(3) Better harnesses can improve agent performance and learning while reducing inference cost.}
With the backbone fixed, changing the harness can substantially improve both performance and learning.
In some cases, it also reduces inference cost through fewer model calls and less repeated context (\S4.3, \S4.6).
\textbf{(4) Agents can learn more slowly in changing environments}, even when the underlying rules stay the same. The drop in learning slope tends to be larger for weaker backbones under the same harness (\S4.5).
\textbf{(5) Agents can learn rules from experience but still struggle to use them to develop effective strategies.} For example, in \game{Haunted Inn}, an agent learns how to assign guests to safe rooms but fails to plan ahead, leaving a later guest with nowhere safe to stay (\S4.2).

Our contributions are threefold:
\begin{itemize}
    \item We introduce \textbf{Learn2Play Bench}, a controlled benchmark of newly designed games with hidden rules, deterministic feedback, and automatic scoring. We also design re-shuffled games to evaluate whether agents could transfer the knowledge to new instances.

    \item We systematically compare backbone models, self-evolving methods, and agent harnesses, revealing factors that affect learning from experience and identifying directions for future research.

    \item We conduct a human study on the same games and compare scores and interaction records to examine differences in performance and behavior between human players and LLM agents.
\end{itemize}

\par\vspace{-6pt}
\section{Related Work}
\label{sec:related}

\textbf{Self-evolving agents.} Recent work develops self-evolving agents to learn from experience at test time, adapting their behavior using feedback from earlier attempts~\astrarevision{\citep{evotest,yang2026scaling}}.
Early methods reuse experience without updating model weights through verbal reflections in Reflexion~\citep{shinn2023reflexion}, extracted insights and retrieved trajectories in ExpeL~\citep{zhao2024expel}, or reusable executable skills in Voyager~\citep{wang2023voyager}. Other methods organize experience into workflows and playbooks.
For example, \emph{AWM} induces action workflows~\citep{wang2024awm}, \emph{Dynamic Cheatsheet} curates compact strategies and code snippets~\citep{suzgun2025dynamic}, and \emph{ACE} incrementally maintains structured playbooks rather than raw transcripts~\citep{zhang2025ace}.
\emph{ReasoningBank} and \emph{ERL} distill trajectories into reusable reasoning patterns or heuristics ~\citep{ouyang2026reasoningbank,allard2026erl}.
\astrarevision{JitRL retrieves past trajectories to estimate action advantages and adjusts the model's output logits at inference time without gradient updates~\citep{li2026jitrl}.}
Beyond constructing memories, \emph{GEPA} evolves prompts through trajectory-level reflection~\citep{agrawal2025gepa}, while \emph{Agents of Change} revises both prompts and agent code through specialized collaborative roles~\citep{belle2025agents}.
\emph{EvoTest} extends evolution to the broader agent configuration, including memory and tool-use routines ~\citep{evotest}.
Meta-TTL further evolves the adaptation policy that determines how experience should update the agent~\citep{lou2026metattl}.
Beyond context-level updates, \emph{TT-SI} generates training examples from uncertain cases and fine-tunes the agent at test time~\citep{acikgoz2025ttsi}.
Our goal is not to propose another adaptation method, but to assess how effectively existing methods learn under controlled conditions.
Our evaluation focuses on learning without weight updates: we compare backbones in a common harness and self-evolving methods with each backbone held fixed.

\textbf{Benchmarks for self-evolving agents.} To assess whether these adaptation mechanisms improve later behavior, benchmarks evaluate agents over task sequences or repeated interactions.
\emph{StreamBench} uses input--feedback streams~\citep{wu2024streambench}, while MemoryBench studies learning from accumulated user feedback and Evo-Memory evaluates memory updates across task streams~\citep{ai2025memorybench,wei2025evomemory}.
Related evaluations extend experience reuse to interactive tasks: LifelongAgentBench tests skill accumulation across related tasks~\citep{zheng2025lifelong}, SWE-Bench-CL evaluates learning across successive coding issues~\citep{joshi2025swebenchcl}, and MemoryArena tests memory use across interdependent sessions~\citep{he2026memoryarena}.
These benchmarks test whether experience improves later performance. On familiar tasks, however, it is unclear whether agents improve by learning something new or by using what they already know.
Other benchmarks study learning directly through interaction with environments. AutoEnv~\citep{zhang2025autoenv} also creates new environments to evaluate learning across diverse settings.
Closer to our setting, \emph{J-TTL}~\citep{evotest} measures repeated improvement on Jericho games, but their public content may already be represented in model pretraining. \emph{ARC-AGI-3} tests interaction with unfamiliar visual environments, but performance also depends on visual capabilities~\citep{arcagi3}.
We use newly designed text-based games with hidden rules to examine whether agents can discover previously unknown rules through actions and feedback.
We also compare fixed and changing instances of the same game to test whether agents can reuse experience when the visible situation changes.

\par\vspace{-6pt}
\begin{figure}[!t]
\centering
\includegraphics[width=\linewidth]{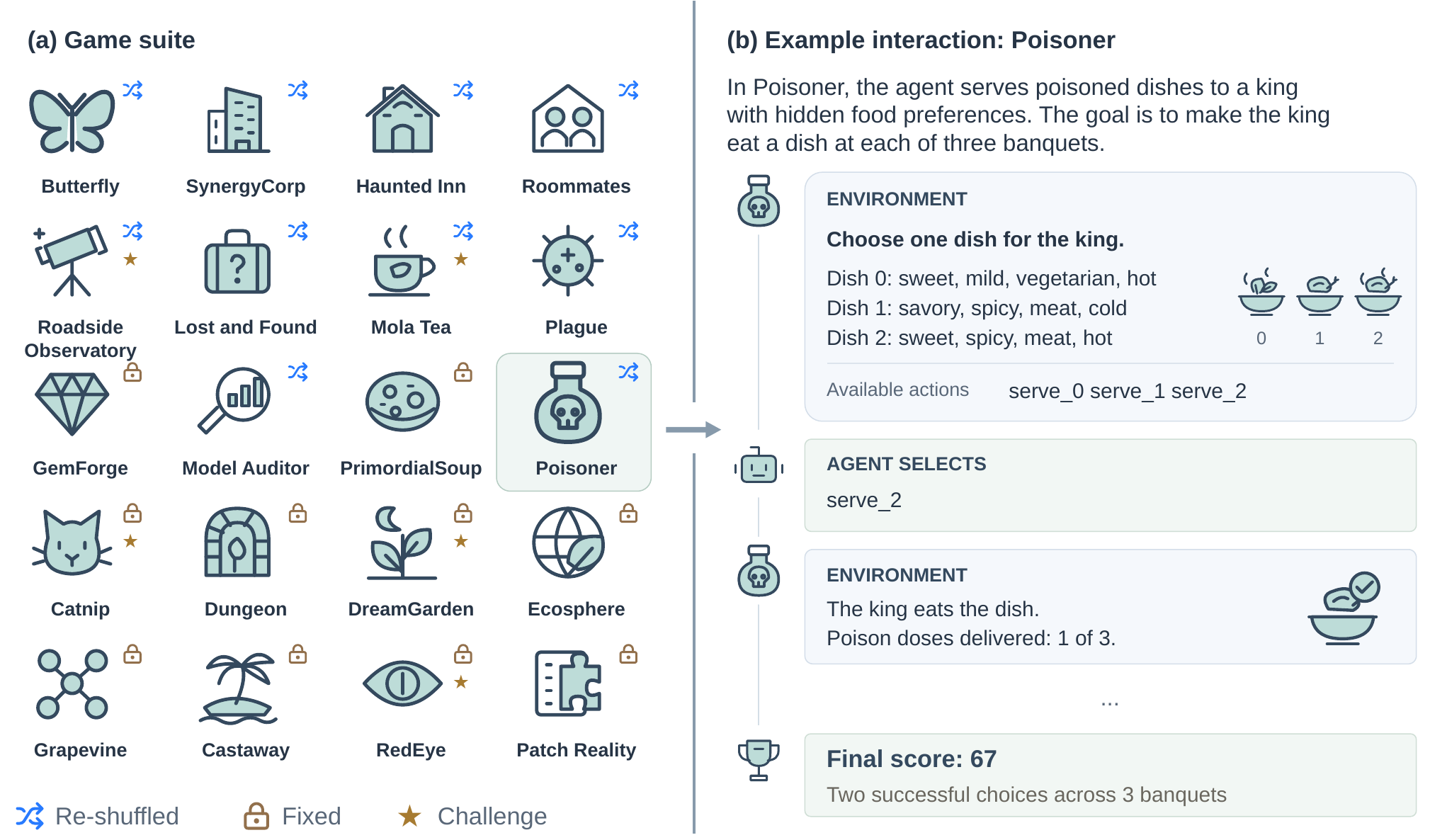}
\caption{Learn2Play Bench overview. The left panel shows 20 manually designed game templates, each of which generates multiple instances using different random seeds. The highlighted \game{Poisoner} template illustrates the interaction protocol on the right: the environment describes the state and enumerates valid actions, the player selects one action, and the environment returns the resulting observation. The episode ends with an automatically computed score (Appendix~\ref{app:game-poisoner}).}
\label{fig:game-interface}
\end{figure}

\section{Evaluation Design}
\label{sec:design}

To measure whether agents improve through experience in unfamiliar environments, we design games that satisfy the following principles:

\begin{itemize}
    \item \textbf{Rule novelty.} We design games with novel hidden rules not found in existing games, including deliberately counterintuitive rules that conflict with common game knowledge.
    For example, in one of our games, instead of hard work, stealing a colleague's lunch helps the player get promoted. The agent must therefore learn which actions lead to promotion from the game's feedback, rather than rely on expectations about real workplaces.
    \item \textbf{Automatic scoring.} Each game automatically assigns a score, enabling large-scale evaluation without human grading.
    \item \textbf{Reproducibility.} Deterministic outcomes allow us to replay interactions, compare agents under the same conditions, and inspect mistakes.
    \item \textbf{Controlled variation across episodes.} To test whether agents transfer learned knowledge, we vary the visible situation across episodes while preserving the core rule.
\end{itemize}

Based on the four principles above, we manually design Learn2Play Bench with 20 game templates, each of which can generate multiple instances that differ in layout, characters, or hidden-rule parameters. Figure~\ref{fig:game-interface} illustrates the game suite and an example interaction. Agents receive a goal and a set of available actions, interact through text to discover the hidden rules, and receive a final score at the end of each episode. They play repeated episodes and can use experience from earlier attempts. \latestrevision{We divide the games along two dimensions. First, ten \textbf{fixed} games repeat the same instance across episodes, whereas ten \textbf{re-shuffled} games change the visible instance while preserving the core rule. Second, we conduct pilot evaluations with several agents to estimate game difficulty and identify five \textbf{challenge} games.\footnote{The challenge games are \game{Catnip}, \game{DreamGarden}, \game{Mola Tea}, \game{RedEye}, and \game{Roadside Observatory}.} We use a ten-episode scoring window for these games and a five-episode window for the remaining \textbf{normal} games.} \revised{Appendix~\ref{app:game-learning-targets} explains what players must learn in each game, and appendix~\ref{app:game-details} provides the corresponding results and archived rule-discovery evidence.}

\par\vspace{-6pt}
\section{Experiments}
\label{sec:exp}

\subsection{Experimental settings}
\label{sec:systems}

\textbf{Datasets.} We evaluate all agents on the same 20 game instances from Learn2Play Bench, using one fixed seed per game template.

\textbf{Protocol.} We evaluate each agent through a black-box command interface. The agent issues actions and observes textual feedback but cannot access the game source code or oracle solvers. We evaluate each game $g$ for $T_g$ episodes: $T_g=5$ for normal games and $T_g=10$ for challenge games, which are harder and therefore require more episodes to learn. The protocol requires three independent trials per game. Each trial starts with no memory of previous trials. Within a trial, the information retained from earlier episodes depends on the harness.

\textbf{Metrics.} We use four metrics to evaluate two complementary aspects of agent behavior across episodes: achieved performance and improvement through experience. Max and Mean measure achieved performance, while Learning Gain (LG) and Learning Slope (LS) measure improvement across episodes. Because score ranges differ across games, we compute these metrics using normalized scores. \revised{Appendix~\ref{app:normalization} explains normalization, trial aggregation, and game weighting.} Let $z_{g,r,t}$ be a normalized terminal score for game $g$, trial $r$, and episode $t$, and let $y_{g,t}$ denote the mean of these scores at episode $t$. We define the four metrics as follows:
\begin{itemize}
\setlength{\itemsep}{1pt}
\setlength{\parskip}{0pt}
\setlength{\parsep}{0pt}
\item \textbf{Max} averages the peak terminal scores $m_{g,r}$ of the $R_g$ scored trial trajectories within the evaluation window: $\mathrm{Max}_g=R_g^{-1}\sum_{r=1}^{R_g}m_{g,r}$. It measures the best performance an agent demonstrates during evaluation.
\item \textbf{Mean} averages performance across all evaluated episodes: $\mathrm{Mean}_g=T_g^{-1}\sum_{t=1}^{T_g}y_{g,t}$. It measures how consistently the agent performs throughout evaluation.
\item \textbf{Learning Gain (LG)} compares average performance in the first and last $k_g=\min(3,\lfloor T_g/2\rfloor)$ episodes: $\mathrm{LG}_g=k_g^{-1}(\sum_{t=T_g-k_g+1}^{T_g}y_{g,t}-\sum_{t=1}^{k_g}y_{g,t})$. It measures the net improvement from the beginning to the end of evaluation.
\item \textbf{Learning Slope (LS)} fits $y_{g,t}=\beta_{0,g}+\beta_{1,g}t+\epsilon_{g,t}$ over all evaluated episodes and reports $\mathrm{LS}_g=\hat{\beta}_{1,g}$. It measures the overall direction and rate of change across the full learning trajectory.
\end{itemize}
We first compute each metric separately for each game and then aggregate it across games as $\sum_g w_gM_g/\sum_g w_g$, assigning weight 3 to challenge games and weight 1 to normal games. The five challenge games and fifteen normal games therefore each receive 50\% of the total weight. To aggregate learning curves from games with different numbers of episodes, we map each game's episodes onto a common 0--100\% progress scale and use linear interpolation only for visualization.

\subsection{RQ1: How effectively do current agents learn?}

We first investigate how learning changes with the choice of baseline model and with the method used to retain experience across episodes. Appendix~\ref{app:self-evolving-method-results} provides additional results and supporting cases from interaction logs.

\astrarevision{\textbf{Baseline models.} We compare eleven backbones using OpenCode~\citep{opencode2026}, a coding agent harness with shell and file tools, under a common agent loop, replay protocol, and memory format. The models are GPT-6 Astra, Claude Opus~5.5, Claude Opus~5, Claude Sonnet~5, Gemini~3.5 Flash, Gemini~3.7 Flash, GPT-5.6-SOL, GPT-5 Mini, DeepSeek V4 Pro, Kimi~K3, and Qwen 3.8 Max.}

\textbf{Methods.} To examine how different ways of retaining and reusing experience affect learning, we compare self-evolving methods separately for Kimi~K3, Claude Opus~5, and GPT-5 Mini, holding the backbone fixed within each comparison.
We evaluate the following methods:
\textbf{\textsc{Naive}} retains the dialogue within the current episode but discards it before the next, providing no cross-episode experience.
\textbf{\textsc{Memory}} cumulatively adds the complete, unmodified interaction histories from preceding episodes to each new episode's prompt, dropping the oldest episodes only when the history exceeds the context budget.
\textbf{Reflexion}~\citep{shinn2023reflexion} reflects on each completed episode to derive concise lessons about discovered rules, successful and failed actions, and next steps, then adds the accumulated lessons to the prompt for later episodes.
\textbf{EvoTest}~\citep{evotest} uses earlier interaction histories and outcomes to evolve a guiding prompt and a state-extraction program that steer action selection in subsequent episodes.
\textbf{ReasoningBank}~\citep{ouyang2026reasoningbank} turns each successful or failed episode into a small set of reusable reasoning lessons and adds the accumulated bank to the prompt in later episodes.
\textbf{Agent Workflow Memory (AWM)}~\citep{wang2024awm} induces reusable multi-step workflows from successful interaction histories and includes the resulting workflows in later episode prompts. Failed episodes are not used.

To assess how these backbones and methods learn across episodes, we plot learning curves for the backbone comparison in Figure~\ref{fig:learning-comparisons}(A) and the method comparison with Claude Opus~5 in Figure~\ref{fig:learning-comparisons}(B).
Table~\ref{tab:rq1-comparisons} reports the backbone, human, and matched-method results, including all six methods with Kimi~K3 and Claude Opus~5.
Table~\ref{tab:rq1-supplementary-methods} reports the GPT-5 Mini method comparison and three-backbone averages. \revised{Appendix~\ref{app:game-details} groups each game's learning curves, leaderboard, and rule-discovery evidence in one entry.}
\revised{The annotation criteria and archive coverage are specified in Appendix~\ref{app:rq1-target-audit}.}

\begin{figure}[!t]
\centering
\begin{minipage}[t]{0.40\linewidth}
\vspace{0pt}
\centering
\setlength{\parskip}{0pt}
\includegraphics[width=\linewidth]{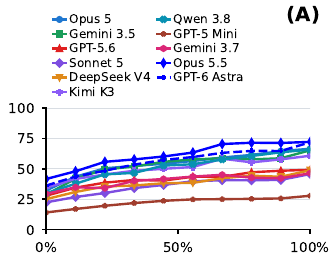}
\par\nointerlineskip
\includegraphics[width=\linewidth]{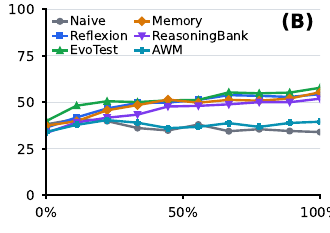}
\par\nointerlineskip
\includegraphics[width=\linewidth]{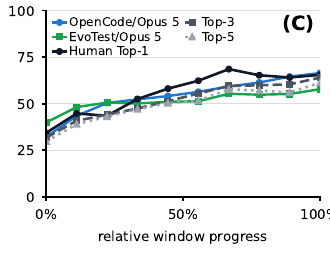}
\setlength{\abovecaptionskip}{2pt}
\captionof{figure}{Normalized learning curves across all 20 games for (A) OpenCode backbones, (B) self-evolving methods with Opus~5, and (C) human Top-$k$ groups and selected agents.}
\label{fig:learning-comparisons}
\end{minipage}\hfill
\begin{minipage}[t]{0.58\linewidth}
\vspace{0pt}
\centering
\captionof{table}{\captionrevision{Performance and learning across backbone models, self-evolving methods, and humans. Backbone comparisons use OpenCode; method comparisons hold the backbone fixed. Human Top-$k$ groups are selected by Max. \opusrevision{Bold marks the best agent value in each group.}}}
\label{tab:rq1-comparisons}
\scriptsize
\renewcommand{\arraystretch}{1.31}
\resizebox{\linewidth}{!}{% Opus 5.5 added by tools/update_learn2play_opus55.py; original rows retained.
% Generated by tools/gen_paper_tables.py from
% results/_analysis/summary_metrics.json derived from canonical_curves.json -- do not edit by hand.
\setlength{\tabcolsep}{2.5pt}
\begin{tabular}{@{}llrrrr@{}}
\toprule
\textbf{Backbone} & \textbf{Method} & \textbf{Max} & \textbf{Mean} & \textbf{LG} & \textbf{LS} \\ 
\midrule
\opusrevision{Claude Opus~5.5} & \opusrevision{OpenCode} & \opusrevision{\textbf{80.1}} & \opusrevision{\textbf{61.0}} & \opusrevision{+23.0} & \opusrevision{+5.22} \\
\astrarevision{GPT-6 Astra} & \astrarevision{OpenCode} & \astrarevision{76.2} & \astrarevision{56.1} & \astrarevision{+23.8} & \astrarevision{+5.88} \\
Claude Opus~5 & OpenCode & \opusrevision{74.5} & \opusrevision{53.8} & +21.3 & +5.13 \\
Qwen 3.8 Max & OpenCode & 72.5 & 51.1 & \textbf{+25.2} & \textbf{+6.30} \\
Gemini~3.5 Flash & OpenCode & 72.0 & 52.2 & +19.8 & +5.37 \\
Kimi~K3 & OpenCode & 68.5 & 50.2 & +17.1 & +4.24 \\
Gemini~3.7 Flash & OpenCode & 58.3 & 39.9 & +10.7 & +2.47 \\
GPT-5.6-SOL & OpenCode & 57.0 & 41.1 & +14.5 & +3.71 \\
Claude Sonnet~5 & OpenCode & 55.9 & 35.6 & +15.7 & +3.72 \\
DeepSeek V4 Pro & OpenCode & 54.2 & 38.2 & +14.7 & +3.66 \\
GPT-5 Mini & OpenCode & 34.0 & 22.3 & +9.2 & +1.83 \\
\midrule
Kimi~K3 & \textsc{Memory} & \textbf{63.3} & \textbf{48.0} & \textbf{+16.9} & \textbf{+4.69} \\
 & EvoTest & 62.4 & 48.0 & +1.8 & +1.01 \\
 & Reflexion & 61.6 & 40.4 & +14.8 & +3.35 \\
 & ReasoningBank & 59.9 & 40.1 & +8.3 & +2.59 \\
 & AWM & 56.1 & 37.0 & +4.8 & +0.62 \\
 & \textsc{Naive} & 54.7 & 31.9 & +0.3 & +0.42 \\
Opus~5 & EvoTest & \textbf{66.1} & \textbf{51.3} & +9.4 & +2.52 \\
 & Reflexion & 64.8 & 48.8 & +11.6 & \textbf{+2.93} \\
 & \textsc{Memory} & 60.9 & 48.2 & +11.4 & +2.87 \\
 & ReasoningBank & 60.2 & 45.3 & \textbf{+12.3} & +2.85 \\
 & \textsc{Naive} & 53.4 & 36.6 & -5.1 & -1.08 \\
 & AWM & 50.7 & 37.8 & +0.7 & +0.43 \\
\midrule
\multicolumn{2}{l}{Human Top-1} & \revised{84.3} & \revised{53.5} & \revised{+21.6} & \revised{+7.33} \\
\multicolumn{2}{l}{Human Top-3} & \revised{79.1} & \revised{50.1} & \revised{+18.9} & \revised{+5.54} \\
\multicolumn{2}{l}{Human Top-5} & \revised{75.1} & \revised{47.1} & \revised{+17.1} & \revised{+4.62} \\
\bottomrule
\end{tabular}
}
\end{minipage}
\end{figure}

\subsubsection{Backbone comparison}

% Appendix~\ref{app:rq1-target-audit} then audits whether game-specific rules are visibly recovered and used in repeated policies.
% Because a missing statement does not prove that knowledge is absent, this evidence remains qualitative rather than being converted into a coverage percentage.

\astrarevision{\textbf{Opus~5.5 leads Max and Mean, while Qwen 3.8 Max leads LG and LS.} The model that scores highest is not necessarily the one that improves most with experience.
As shown in Table~\ref{tab:rq1-comparisons}, Opus~5.5 achieves the highest peak and average scores, whereas Qwen 3.8 Max shows greater improvement across episodes.}

\textbf{Stronger models use feedback to adjust their decisions, whereas weaker models may repeat the same actions despite new observations.}
We inspected the interaction logs to understand how models use experience.
Opus~5 and Qwen 3.8 Max test their understanding of the hidden rules and update it when new observations do not match their expectations.
In contrast, weaker models sometimes repeat actions that worked before, even when they no longer work.
They may also change several factors at once, making it hard to tell which factor affects the outcome.
For example, in \game{Poisoner}, agents choose dishes for a king and learn about his food preferences from whether he accepts or rejects them.
In one GPT-5 Mini run, the agent keeps choosing dishes by their position on the menu, without using feedback from earlier choices.
Kimi~K3 instead uses that feedback to choose which food attribute to test next.

\subsubsection{Self-evolving method comparison}

\textbf{Simply using past interaction histories can outperform more elaborate self-evolving methods.}
\textsc{Memory} retains raw histories without reflection, rule extraction, or workflow induction, yet leads all four metrics with Kimi~K3.
This is because reflection and memory processing can turn an incomplete understanding into rules or workflows that later episodes continue to follow, while raw histories keep the original evidence available for revising those interpretations.
For example, in \game{GemForge}, agents combine ingredients to make gems. After one combination fails, an EvoTest--Opus~5 run adds a rule to its prompt telling it to avoid many other combinations, including one that would succeed.

\textbf{OpenCode achieves better performance than the evaluated self-evolving methods when using the same backbone.}
OpenCode is more flexible during play.
OpenCode lets the model decide when to review earlier feedback, run calculations, or check a plan before taking an action.
By contrast, the evaluated self-evolving methods follow preset steps for choosing actions and updating experience.
When an unexpected result occurs, OpenCode gives the model room to investigate and revise its plan before continuing, rather than only updating its memory for later episodes.

\subsubsection{What limits further improvement?}

\textbf{Agents may stop exploring better strategies after finding one that partly works.}
The game-specific audit shows agents repeating known strategies instead of testing alternatives.
For example, in \game{GemForge}, players combine ingredients to make increasingly valuable gems.
After several failed attempts, one Opus~5 run concludes that higher-value gems cannot be made and keeps repeating known recipes, limiting further improvement.

\textbf{Agents may know the rules but fail to plan ahead.}
In an Opus~5 \game{Haunted Inn} run, the agent states the relevant taboos but assigns rooms greedily, leaving no safe room for a later guest.
In the next episode, it reserves rooms for guests with fewer safe options.
The failure lies in choosing an action that works now without considering what it prevents later.

% \textbf{In the inspected method traces, reflection can identify mistakes without preventing their recurrence.}
% In \game{Model Auditor}, ReasoningBank--Opus~5 recalls the need to verify contradictions but uses self-generated tool responses as evidence to dismiss them.
% In \game{Haunted Inn}, Reflexion--Opus~5's final reflection sees only the current episode and rejects a previously supported rule without revisiting earlier evidence.
% These cases suggest that articulating a lesson does not ensure reliable evidence checking: reflection can reproduce or introduce reasoning errors.

\subsection{RQ2: How does the agent harness affect learning?}
\label{sec:harness}

% Fixed columns keep the prose outside the table at every page position.
\noindent
\begin{minipage}[t]{0.63\linewidth}
\vspace{0pt}
To isolate the effect of the harness from the choice of model, we hold the backbone fixed and compare OpenCode with Claude Code for Opus~5, and with Codex for GPT-5.6-SOL. All configurations follow the same evaluation protocol. \revised{Appendix~\ref{app:agent-harness-results} specifies trial isolation, within-trial history, and the matched harness configurations.}

\textbf{Claude Code and Codex yield higher performance and stronger learning than OpenCode when paired with the same backbone.}
As shown in Table~\ref{tab:rq1-systems}, Claude Code outperforms OpenCode on all four metrics with Opus~5, as does Codex with GPT-5.6-SOL.
This is because models are trained on their harnesses and thus would perform better, and these harnesses are also adapted to the models' capabilities~\citep{openaiCodexPromptingGuide,anthropicSeeingLikeAnAgent}.
%This alignment helps explain why GPT-5.6-SOL performs better with Codex and Opus~5 performs better with Claude Code than with OpenCode.}
\end{minipage}\hfill
\begin{minipage}[t]{0.35\linewidth}
\vspace{0pt}
\centering
\captionof{table}{Same-backbone harness comparison.}
\label{tab:rq1-systems}
\scriptsize
\setlength{\tabcolsep}{3pt}
\resizebox{\linewidth}{!}{\begin{tabular}{@{}lrrrr@{}}
\toprule
Harness & Max & Mean & LG & LS \\
\midrule
\multicolumn{5}{c}{Claude Opus~5} \\
OpenCode & 74.5 & 53.8 & +21.3 & +5.13 \\
Claude Code & \textbf{81.8} & \textbf{62.3} & \textbf{+27.1} & \textbf{+6.59} \\
\midrule
\multicolumn{5}{c}{GPT-5.6-SOL} \\
OpenCode & 57.0 & 41.1 & +14.5 & +3.71 \\
Codex & \textbf{74.3} & \textbf{54.8} & \textbf{+22.2} & \textbf{+5.66} \\
\bottomrule
\end{tabular}
}
\end{minipage}\par

\subsection{\latestrevision{RQ3: How do humans and agents learn differently?}}

We compare humans and agents to understand how their use of experience differs when both must discover unfamiliar rules. We recruited college students for the human study. They use the same game interface, with 21--30 participants per game. \latestrevision{For each game, we select the Top-1, Top-3, and Top-5 records by Max. Their aggregate metrics appear in Table~\ref{tab:rq1-comparisons}.} Appendix~\ref{app:human-model-full} provides full results, including the mean across all participants, behavioral statistics, and cases from interaction logs. Appendix~\ref{app:game-details} reports per-game scores.
\opusrevision{To compare their learning across episodes, Figure~\ref{fig:learning-comparisons}(C) shows these human curves alongside OpenCode--Claude Opus~5 and EvoTest--Opus~5.}

\clarifiedhistory{\textbf{Human Top-1 achieves higher Max than the models and self-evolving methods evaluated in our experiments.}}
This may be because humans explore more broadly, whereas agents exploit discovered strategies more consistently. To examine this behavioral difference beyond the single best record, we compare observed episodes from Human Top-3 with the models and self-evolving methods. In fixed games, consecutive action sequences have an average similarity of 0.54 for humans and 0.66 for agents. After a lower-scoring episode, humans reach a new personal best in the next episode 33\% of the time, compared with 22\% for agents. These observations suggest that humans more often change strategies and recover from temporary failures, while agents more often repeat a reliable but potentially suboptimal strategy.

% Evidence checked 2026-09-12. Human source: results/_analysis/learning/human/app.db.
% Fixed-game adjacent-sequence similarity over observed episode pairs (SequenceMatcher): Human Top-3=0.538.
% pooled six Opus-5 methods=0.657. After a score decrease, next-episode new-best:
% Human Top-3=13/39 (33.3%). Pooled six Opus-5 methods=52/238 (21.8%).

\opusrevision{\latestrevision{\textbf{Top-performing humans perform better when success requires finding new strategies, whereas agents perform better when success requires executing long action sequences or doing precise calculations.}}}
\opusrevision{Human Top-1 scores higher in \game{GemForge} and \game{Grapevine}, while OpenCode--Opus~5.5 scores higher in \game{RedEye} and \game{Model Auditor}. In \game{RedEye}, successful agents combine clues and carry out all the steps needed to rescue the passengers. In \game{Model Auditor}, they use programs to infer the hidden rules and check their predictions. Some human players in this game reduce or stop testing in later episodes, suggesting that the effort required to keep investigating may partly explain their lower scores.}

\subsection{\latestrevision{RQ4: How well does learned knowledge transfer to new game instances?}}

We assess how well agents transfer what they learn to new instances in re-shuffled games. We test seven backbone models using the same OpenCode harness. We also test Claude Code with Opus~5 and Codex with GPT-5.6-SOL. For each re-shuffled game, we create a same-instance control: the agent encounters the same instance in every episode, while the re-shuffled version changes the visible instance but preserves the underlying rule. We compare LS over ten episodes for games evaluated in both conditions, with three independent trials required per game. Appendix~\ref{app:transfer-full} also reports the six methods with Opus~5. Appendices~\ref{app:evidence-transfer} and~\ref{app:evidence-feedback} provide examples from interaction logs of applying learned rules and repeating menu positions.

\noindent
\begin{minipage}[t]{0.43\linewidth}
\vspace{0pt}
\textbf{When game instances change each episode, Sonnet 5 and the Gemini Flash models improve much more slowly than Opus 5 and Kimi K3.}
Opus 5 and Kimi K3 show smaller drops in learning slope (Table~\ref{tab:transfer}). In one \game{SynergyCorp} run, Opus~5 applies a rule learned from earlier bosses to new bosses, whereas Sonnet~5 and Gemini~3.5 Flash test each new boss's preferences again.

\textbf{Past success can mislead transfer when an agent remembers what worked but not why it worked.}
Agent systems with higher re-shuffled LS can identify why an action worked and find where the same conditions hold in a new instance.
Systems with lower LS sometimes repeat episode-specific choices, such as a menu position, even after the menu changes.
\end{minipage}\hfill
\begin{minipage}[t]{0.55\linewidth}
\vspace{0pt}
\centering
\captionof{table}{\captionrevision{Learning slopes when the same game instance repeats (Control) or changes across episodes (Re-shuf.), with the underlying rules held fixed. LS drop is Control minus Re-shuf.; positive values indicate slower learning after re-shuffling.}}
\label{tab:transfer}
\footnotesize
\setlength{\tabcolsep}{3pt}
\begin{tabular}{@{}l@{\hspace{3pt}}r@{\hspace{3pt}}r@{\hspace{3pt}}r@{}}
\toprule
\textbf{Agent} & \textbf{Control} & \textbf{Re-shuf.} & \textbf{LS drop} \\
\midrule
\multicolumn{4}{l}{\textbf{OpenCode backbones}} \\
Claude Opus~5 & 4.10 & 3.51 & 0.59 \\
Qwen 3.8 Max & 3.16 & 3.57 & -0.41 \\
Gemini~3.5 Flash & 4.50 & 2.76 & 1.73 \\
Kimi~K3 & 4.19 & 3.31 & 0.88 \\
Gemini~3.7 Flash & 4.49 & 2.34 & 2.15 \\
GPT-5.6-SOL & 3.72 & 1.78 & 1.94 \\
Claude Sonnet~5 & 4.90 & 2.23 & 2.67 \\
\midrule
\multicolumn{4}{l}{\textbf{Deployed agents}} \\
Claude Code (Opus~5) & 4.30 & 3.77 & 0.52 \\
Codex (GPT-5.6-SOL) & 2.51 & 2.71 & -0.19 \\
\bottomrule
\end{tabular}

\end{minipage}\par

\subsection{RQ5: Do higher inference costs lead to better performance?}

We compare agents' Max scores with their mean cost per episode in Figure~\ref{fig:rq4-cost}.
\revised{Table~\ref{tab:cost} reports the numerical cost results for each configuration.} Appendix~\ref{app:cost-estimation} provides calculation details, token-use breakdowns, and supporting examples from interaction logs.

\begin{figure}[!t]
\centering
\includegraphics[width=\linewidth]{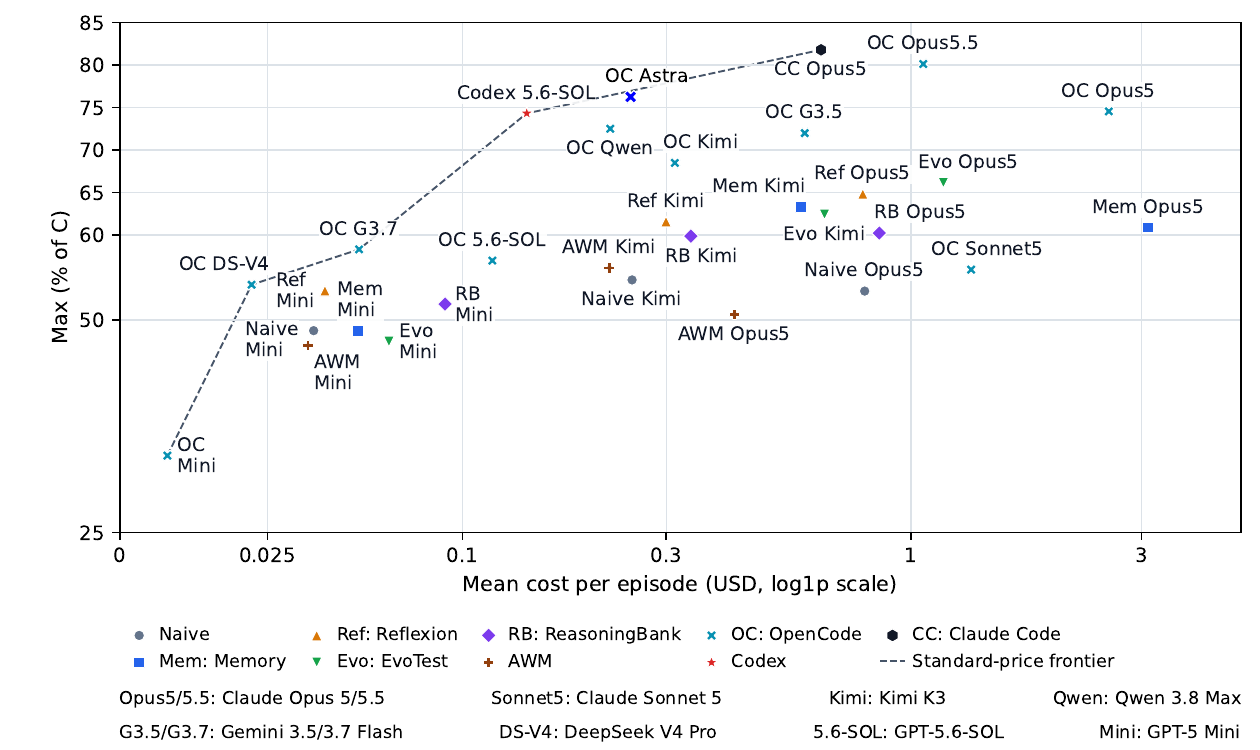}
\caption{\captionrevision{\astrarevision{Max score versus mean cost per episode. The dashed line shows the performance--cost frontier for configurations with standard-price estimates.}}}
\label{fig:rq4-cost}
\vspace{-10pt}
\end{figure}

\textbf{Under the same harness, models with higher token prices can achieve better performance at a lower overall cost.}
With OpenCode, Kimi~K3 achieves higher performance at a lower overall cost than Claude Sonnet~5, despite having higher input and output token prices. Kimi~K3 also uses fewer total tokens per episode: 0.56M on average, compared with 0.73M for Claude Sonnet~5 (Table~\ref{tab:cost}). Higher token prices therefore do not imply higher overall cost.

\textbf{With the model fixed, a better-suited harness can improve performance while reducing cost.}
OpenCode with Kimi~K3 achieves higher Max at lower estimated cost than \textsc{Memory}, EvoTest, and ReasoningBank. Claude Code with Opus~5 also outperforms these self-evolving methods and OpenCode at lower estimated cost. For Kimi~K3, OpenCode uses 56\% fewer input tokens than \textsc{Memory} and 71\% and 41\% fewer generated tokens than EvoTest and ReasoningBank, respectively, as reported in Table~\ref{tab:kimi-token-breakdown}. These self-evolving methods call the model for each action and repeatedly supply past interactions or stored memories. OpenCode instead executes several actions from one model response and removes repeated game feedback, reducing model calls and repeated input.

% \subsection{RQ5: Failure modes}
% \label{sec:failure}

% To further understand why agents fail, we investigate the logs of all the agents and human, to distinguish what they discover, retain, and apply. We have found the following failure modes:

% \textbf{Agents may stop exploring better strategies after finding one that partly works.}
% In \game{GemForge}, players combine ingredients to make increasingly valuable gems.
% After several failed attempts, one Opus~5 run concludes that higher-value gems cannot be made and keeps repeating recipes it already knows.
% This premature conclusion prevents further tests of potentially better recipes.

% \textbf{Agents may predict outputs from the current input alone, ignoring previous inputs.}
% In \game{Model Auditor}, players test a hidden machine, then predict its outputs for a new sequence of inputs.
% The machine remembers earlier inputs, so the same input can produce different outputs at different points in the sequence.
% Resetting before every test erases this memory.
% An agent that learns only from these tests may predict correctly just after a reset but fail when inputs arrive one after another without a reset.

\par\vspace{-6pt}
\section[Discussions and Future Directions]{Discussions and Future Directions}

Building on our findings, we discuss open questions in how to improve the learning ability of LLM agents and outline directions for future research.

\textbf{How should agents balance exploration and exploitation?} In our comparison of human players and agents, Human Top-1 achieves a higher Max score than every agent configuration we evaluate, as shown in Tables~\ref{tab:rq1-comparisons} and~\ref{tab:rq1-systems}. This is because humans try more varied strategies across episodes, while agents more often reuse strategies that have worked before (Appendix~\ref{app:evidence-human-statistics}). Future work should better balance exploration and exploitation, allowing agents to test new strategies even when their current strategy already earns high scores.

\textbf{How should agents store learned rules?} With Kimi~K3, \textsc{Memory} outperforms the other self-evolving methods overall, as shown in Table~\ref{tab:rq1-comparisons}. Unlike those methods, it retains full interaction histories rather than first distilling them into rules. This distinction matters when agents draw rules from limited evidence. In \game{GemForge}, EvoTest tried one gem combination and failed, but its later prompt instructed the agent not to test other combinations it had never tried (Appendix~\ref{app:evidence-memory}). The prompt thus turned incomplete evidence into a definitive rule. Future work can explore ways to store experience, such as linking each learned rule to the original interactions behind it so agents can revisit the evidence and revise the rule when new feedback arrives.

\textbf{How can agents learn what stays constant as environments change?} Our transfer evaluation shows that changing surface details, such as layouts or characters, can reduce agents' learning slopes even when the underlying rules remain the same (Table~\ref{tab:transfer}). In real-world tasks, agents also regularly encounter changes in their surroundings, available resources, and the people they interact with. This raises a broader question: how can agents adapt to these changes in their environments? Future methods could help agents compare experiences across different situations, distinguish shared rules from instance-specific details, and check when those rules still apply.

\vspace{-10pt}
\section{Conclusion}
\vspace{-4pt}

In this study, we investigate how effectively LLM agents learn from experience in unfamiliar environments.
Existing evaluations make it difficult to distinguish newly acquired knowledge from prior knowledge. To address this issue, we introduce Learn2Play Bench, a benchmark of newly designed text-based games with hidden rules that agents discover through interaction without updating their model weights.
We conducted comprehensive experiments comparing backbone models, self-evolving methods, deployed agent harnesses, and human players.
Our findings show that more elaborate self-evolving methods do not always improve learning, \clarifiedhistory{human players explore more varied strategies, with top-performing humans achieving higher peak scores than the models and self-evolving methods we evaluate}, and agent harnesses can improve performance while reducing task costs.
We hope this work provides a foundation for evaluating and improving how agents learn from experience.

\par\endgroup

% Ethics and reproducibility statements.
\begingroup

\endgroup

\bibliography{references}
\bibliographystyle{iclr2027_conference}

\clearpage
\appendix

\startcontents[appendix]

\section*{List of Appendices}
\printcontents[appendix]{}{1}{}
\clearpage

\raggedbottom
\AddToHook{env/tabular/begin}{\footnotesize\setlength{\tabcolsep}{3pt}\renewcommand{\arraystretch}{1.0}}
\AddToHook{env/longtable/begin}{\footnotesize}
\section{Games and Evaluation Settings}
\label{app:game-learning-targets}
\subsection{Games and learning tasks}
\revised{The games test whether players can discover useful rules from feedback and apply them in later episodes. The two tables below describe what must be learned, rather than assigning separate ability scores. Detailed results for each game appear in Appendix~\ref{app:game-details}.}

\revised{In re-shuffled games, the visible situation changes but the underlying rule remains the same. Players must use what they have learned with new menus, guests, or other instances. In fixed games, the same instance repeats, so players can refine a map, recipe, or action sequence.}
\begin{table}[!htbp]
\caption{Re-shuffled games: knowledge that must carry over to new instances.}
\label{tab:reshuffled-games}
\begin{center}
\scriptsize
\setlength{\tabcolsep}{3pt}
\renewcommand{\arraystretch}{1.0}
\begin{tabular}{p{0.25\linewidth}p{0.68\linewidth}}
\toprule
\textbf{Game} & \textbf{What the player needs to learn} \\
\midrule
\game{Poisoner} & Infer taste constraints from accepted and rejected dishes and apply them to new menus. \\
\game{SynergyCorp} & Learn which character traits predict preferred, weak, and taboo actions. \\
\game{Haunted Inn} & Discover guest--room, adjacency, and count taboos through controlled placements. \\
\game{Roommates} & Infer pairwise and nonlinear grouping rules from room-level feedback. \\
\game{Roadside Observatory} & Learn social protocols and resource requirements for new encounters. \\
\game{Lost and Found} & Learn how to verify ownership without revealing private information. \\
\game{Mola Tea} & Infer the adoption chain and when each promotion becomes effective. \\
\game{Plague} & Locate asymptomatic carriers from spread patterns and isolate them early. \\
\game{Butterfly} & Infer inheritance laws to breed new target combinations. \\
\game{Model Auditor} & Learn how to probe a new stateful machine and predict its outputs. \\
\bottomrule
\end{tabular}
\end{center}
\end{table}
\FloatBarrier

\begin{table}[!htbp]
\caption{Fixed games: knowledge accumulated about a recurring instance.}
\label{tab:fixed-games}
\begin{center}
\scriptsize
\setlength{\tabcolsep}{3pt}
\renewcommand{\arraystretch}{1.0}
\begin{tabular}{p{0.25\linewidth}p{0.68\linewidth}}
\toprule
\textbf{Game} & \textbf{What the player needs to learn} \\
\midrule
\game{PrimordialSoup} & Discover reaction recipes and plan synthesis within a resource budget. \\
\game{GemForge} & Discover multi-tier recipes and manage limiting ingredients. \\
\game{Catnip} & Learn action--target matches and when to exit for the best reward. \\
\game{Dungeon} & Remember rooms, keys, hazards, and treasures to refine an escape route. \\
\game{DreamGarden} & Infer delayed ecological effects and plan actions across turns. \\
\game{Ecosphere} & Infer food-web, competition, and waste dynamics to plan stocking. \\
\game{Grapevine} & Infer a cascade network and choose effective seed combinations. \\
\game{Castaway} & Remember the map, recipes, hazards, and steps needed to escape. \\
\game{RedEye} & Combine character evidence and execute the rescue steps in order. \\
\game{Patch Reality} & Remember diagnosis--repair mappings, decoys, and prerequisites. \\
\bottomrule
\end{tabular}
\end{center}
\end{table}
\FloatBarrier

\subsection{Evaluation and scoring}
\label{app:normalization}
\revised{We first normalize each episode's raw score, then compute metrics within each game, and finally average across games. This section specifies those steps so that aggregate scores can be reproduced. Human record selection is described in Appendix~\ref{app:human-model-full}. Cost calculations are in Appendix~\ref{app:cost-estimation}.}

\subsubsection{\revised{Instances, episodes, and trials}}
\revised{All configurations use the pinned game instances listed in Table~\ref{tab:normalization-references}. A trial is one sequence of episodes with no experience carried over from another trial. The protocol requires three trials per configuration and game. We evaluate each trial over five episodes for normal games and ten episodes for challenge games.}
Appendix~\ref{app:trial-variability} reports aggregate results with trial standard deviations. Appendix~\ref{app:game-details} gives the corresponding per-game results.

\subsubsection{\revised{From raw scores to per-game metrics}}
\revised{Let $x_{g,r,t}$ be the raw score for game $g$, trial $r$, and episode $t$, and let $C_g$ be that game's reference score. We convert it to $z_{g,r,t}=100x_{g,r,t}/C_g$. This puts different games on a common scale.}

\revised{Max first takes the peak normalized terminal score $m_{g,r}$ in each trial's scoring window and then averages those peaks across the $R_g$ scored trial trajectories: $\mathrm{Max}_g=R_g^{-1}\sum_{r=1}^{R_g}m_{g,r}$. For example, trial peaks of 40, 60, and 80 give Max $=60$, not 80. The other metrics use the episode-wise mean of terminal scores, $y_{g,t}$: Mean averages its scores, LG compares its late and early scores, and LS measures its linear trend per episode. Section~\ref{sec:systems} defines the formulas. Appendix~\ref{app:human-model-full} specifies human score construction.}

\subsubsection{\revised{Averaging across games and plotting curves}}
\revised{Each challenge game has weight 3 and each other game has weight 1. Across the 20 games, both groups have total weight 15: $5\times3=15\times1$. For a metric $M_g$, the aggregate is $\sum_g w_gM_g/\sum_g w_g$ over games with a defined value. The human selection rule gives all four metrics a value in every one of the 20 games.}

\revised{For aggregate figures only, a game's curve is interpolated to ten equally spaced positions across its scoring window. Thus 100\% on the horizontal axis means episode 5 for a normal game and episode 10 for a challenge game. Interpolation is not used for Max, Mean, LG, or LS. All three panels use the same 20 games throughout each curve. Per-game figures show the original episode axis and raw scores.}

\subsubsection{\revised{Reference scores}}
\revised{An exact reference is supported by an oracle or proof. A partial reference is the available proved value for the pinned instance, without a complete optimality result. An estimated reference comes from a legal reference search or a score cap. These evidence levels describe the reference, not the agent's knowledge of the game.}
% Generated by tools/gen_normalization_table.py -- do not edit by hand.
\begin{table}[!htbp]
\caption{\captionrevision{Per-game reference scores used for normalization and their supporting evidence.}}
\label{tab:normalization-references}
\begin{center}
\scriptsize
\setlength{\tabcolsep}{3.2pt}
\begin{tabular}{@{}lrrll@{}}
\toprule
\textbf{Game} & \textbf{Seed} & \textbf{$C$} & \textbf{Source} & \textbf{Normalized score} \\
\midrule
\game{PrimordialSoup} & 1 & 110 & exact & $100x_t/110$ \\
\game{GemForge} & 1 & 791 & exact & $100x_t/791$ \\
\game{Catnip} & 1 & 320 & estimated & $100x_t/320$ \\
\game{Dungeon} & 2 & 54 & exact & $100x_t/54$ \\
\game{SynergyCorp} & 2 & 137 & exact & $100x_t/137$ \\
\game{DreamGarden} & 1 & 6300 & estimated & $100x_t/6300$ \\
\game{Ecosphere} & 1 & 339 & exact & $100x_t/339$ \\
\game{Grapevine} & 1 & 16 & exact & $100x_t/16$ \\
\game{Haunted Inn} & 2 & 200 & partial & $100x_t/200$ \\
\game{Castaway} & 1 & 241 & partial & $100x_t/241$ \\
\game{Roadside Observatory} & 2 & 1000 & exact & $100x_t/1000$ \\
\game{Lost and Found} & 1 & 1000 & exact & $100x_t/1000$ \\
\game{Mola Tea} & 1 & 1000 & estimated & $100x_t/1000$ \\
\game{Plague} & 2 & 100 & exact & $100x_t/100$ \\
\game{Patch Reality} & 1 & 960 & estimated & $100x_t/960$ \\
\game{Poisoner} & 2 & 100 & exact & $100x_t/100$ \\
\game{Roommates} & 1 & 7 & exact & $100x_t/7$ \\
\game{RedEye} & 2 & 100 & exact & $100x_t/100$ \\
\game{Butterfly} & 1 & 439 & exact & $100x_t/439$ \\
\game{Model Auditor} & 1 & 100 & exact & $100x_t/100$ \\
\bottomrule
\end{tabular}
\end{center}
\end{table}

\FloatBarrier

\section{\revised{Additional Results by Research Question}}
\label{app:additional-results}
\revised{The following subsections follow RQ1--RQ5 in the main text. Each brings together the relevant setup, full results, and supporting interaction evidence. Appendix~\ref{app:game-details} provides a separate entry for each game.}

\subsection{\revised{RQ1: How Effectively Do Agents Learn?}}
\label{app:self-evolving-method-results}
\begin{figure}[!htbp]
\centering
\includegraphics[width=0.70\linewidth]{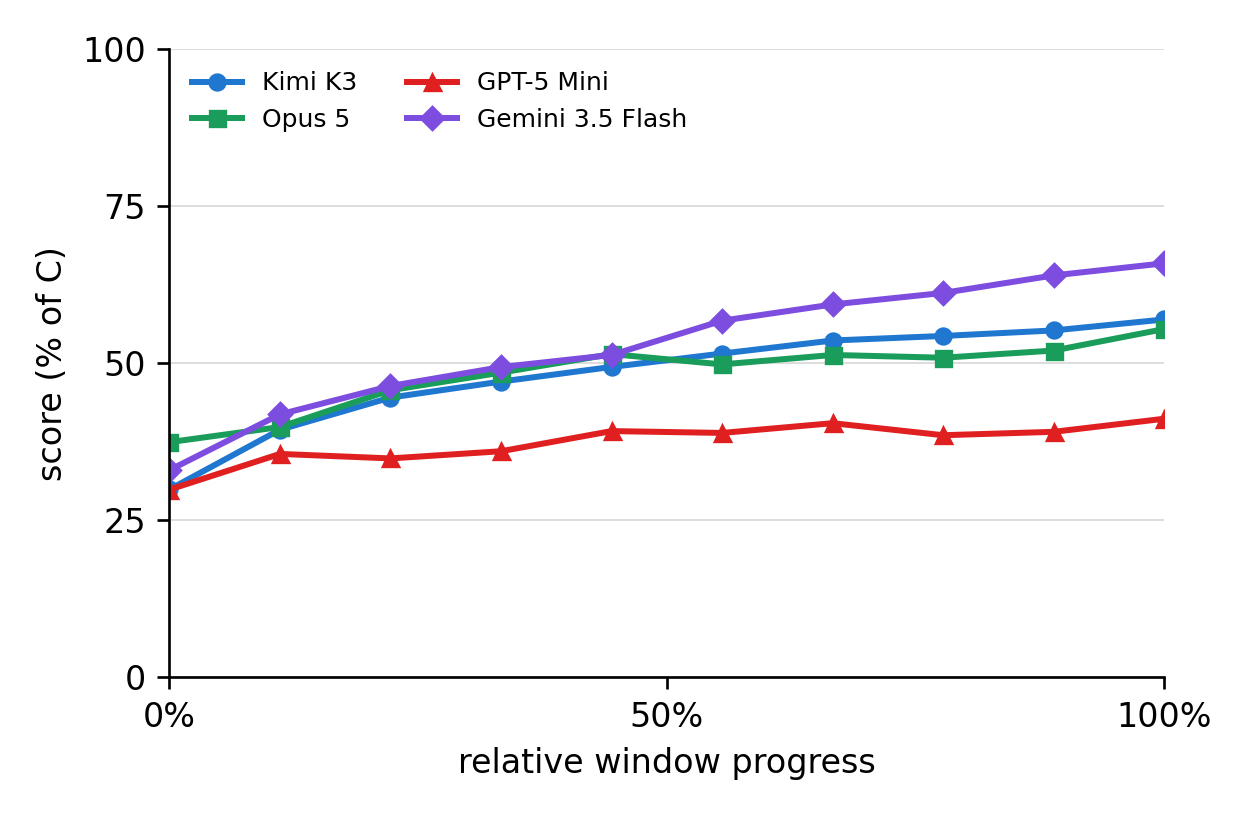}
\caption{\captionrevision{Learning curves for \textsc{Memory} with four backbone models. Scores use the shared normalization and game weights. Relative progress aligns the five- and ten-episode scoring windows.}}
\label{fig:memory-backbones}
\end{figure}

This appendix supplements Table~\ref{tab:rq1-comparisons} with the GPT-5 Mini method comparison and three-backbone averages in Table~\ref{tab:rq1-supplementary-methods}, as well as learning curves and behavioral evidence. Figure~\ref{fig:memory-backbones} compares all four Memory backbones. \revised{We also examine interaction records to distinguish repeated success from discovery of the underlying rules.}
\begin{table}[!htbp]
\centering
\caption{Self-evolving methods with GPT-5 Mini and their average performance across Kimi~K3, Opus~5, and GPT-5 Mini. Bold marks the best value within each group. The last six rows average each metric equally across the three backbones.}
\label{tab:rq1-supplementary-methods}
\appendixtable{% Generated by tools/gen_paper_tables.py from
% results/_analysis/summary_metrics.json derived from canonical_curves.json -- do not edit by hand.
\setlength{\tabcolsep}{2.5pt}
\begin{tabular}{@{}llrrrr@{}}
\toprule
\textbf{Backbone} & \textbf{Method} & \textbf{Max} & \textbf{Mean} & \textbf{LG} & \textbf{LS} \\ 
\midrule
GPT-5 Mini & Reflexion & \textbf{53.5} & 34.4 & +3.6 & +0.97 \\
 & ReasoningBank & 51.9 & 32.5 & -0.4 & +0.00 \\
 & \textsc{Naive} & 48.7 & 30.5 & -0.1 & +0.36 \\
 & \textsc{Memory} & 48.7 & \textbf{37.3} & +5.9 & +1.56 \\
 & EvoTest & 47.5 & 31.8 & \textbf{+7.4} & \textbf{+1.61} \\
 & AWM & 47.0 & 28.9 & -2.9 & -0.98 \\
\midrule
3-backbone avg. & Reflexion & \textbf{60.0} & 41.2 & +10.0 & +2.42 \\
 & EvoTest & 58.7 & 43.7 & +6.2 & +1.71 \\
 & \textsc{Memory} & 57.6 & \textbf{44.5} & \textbf{+11.4} & \textbf{+3.04} \\
 & ReasoningBank & 57.3 & 39.3 & +6.7 & +1.82 \\
 & \textsc{Naive} & 52.3 & 33.0 & -1.6 & -0.10 \\
 & AWM & 51.3 & 34.6 & +0.8 & +0.02 \\
\bottomrule
\end{tabular}
}
\end{table}
\FloatBarrier

\subsubsection{\revised{Evidence of rule discovery}}
\revised{Appendix~\ref{app:game-details} reports each game's target definitions and evidence labels. We retain these per-game records without aggregating the labels into overall percentages, because the archived criteria mix complete-rule discovery with successful execution. Appendix~\ref{app:rq1-target-audit} explains the labels and annotation scope.}

\subsubsection{\revised{Exploration and planning cases}}
\label{app:evidence-rq1}
\label{app:behavioral-evidence}
\revised{The following cases show how a policy changes across episodes. They use recorded actions, visible narration, and saved notes, not private reasoning. They illustrate possible mechanisms rather than their frequency. These supplementary cases draw on interactions beyond the scoring windows and do not enter the reported scores.}

\textbf{Stopping exploration in GemForge.} In OpenCode--Opus~5 trial 1, a failed fusion is followed by the statement ``never combine tier-3 gems.'' The agent then treats its recipe map as complete and repeats a plan scoring 426. This illustrates how an incomplete belief can stop further tests.

\textbf{Knowing taboos without reserving rooms.} OpenCode--Opus~5 trial 1 records both the actor--scholar adjacency taboo and the mirror-room restriction on lantern carriers. Greedy placement leaves no safe room for the last actor. The agent then says ``I need to reserve buffer positions proactively'' and changes its allocation plan. The scores are 100 and 115, but the guest lists also change: the evidence is the revised allocation policy, not a controlled score improvement.

\subsubsection{\revised{How processed experience can restrict exploration}}
\label{app:evidence-memory}
\revised{\textbf{One failed fusion becomes a broad rule.} EvoTest--Opus~5 trial 3 scores 46, 81, 83, 83, and 83 in GemForge's scoring window. An earlier prompt records that the only fusion tried, jade with opal, produced slag. The episode-5 prompt then instructs the agent never to attempt tier-2 or tier-3 fusion, ruling out combinations it has not tested. This case shows how a processed prompt can generalize beyond the observed failure. It does not establish that compression causes the aggregate difference from Memory.}

\subsubsection{Using feedback instead of repeating menu positions}
\label{app:evidence-feedback}
In \game{Poisoner}, OpenCode--GPT-5 Mini trial 3 (seed 2) serves dish 0 at all three banquets in episodes 2, 3, and 5, even though dishes at that position are both accepted and rejected. Its saved notes describe a policy of always serving the same index. By contrast, OpenCode--Kimi~K3 trial 1 (seed 2) compares accepted dishes across episodes 1 and 2 and identifies sourness as irrelevant. In episode 3, an accepted non-hot dish leads it to rule out temperature and then test saltiness. These cases use recorded actions and visible narration within the first five episodes. They illustrate different uses of feedback, not the frequency of either behavior across all trials.
% Evidence: results/poisoner/opencode_gpt-5-mini_t3/seeds/seed2.events.jsonl and seed2.json;
% results/poisoner/opencode_kimi-k3_t1/seeds/seed2.events.jsonl and seed2.json.

\subsection{\revised{RQ2: How Does the Harness Affect Learning?}}
\label{app:agent-harness-results}
\revised{The comparison holds the backbone fixed: OpenCode and Claude Code both use Opus~5, while OpenCode and Codex both use GPT-5.6-SOL. Table~\ref{tab:rq1-systems} reports their scores, and the right-hand panel of each game entry in Appendix~\ref{app:game-details} shows their learning curves.}

\revised{Each trial starts without experience from another trial. Within a trial, earlier episode interactions remain available in the ongoing session, subject to the harness's context handling. Claude Code disables persistent auto-memory. Codex uses \texttt{--ephemeral}. OpenCode starts a new session rather than resuming an unrelated one. These settings separate trials without discarding all within-trial experience. Other reasoning settings use the CLI or provider defaults.}

\revised{The per-game evidence matrices in Appendix~\ref{app:game-details} include the four harness--model configurations used in this comparison. The two OpenCode reference rows appear in each matrix's backbone block. These labels are reported only for individual games, not as overall percentages.}

\subsection{\latestrevision{RQ3: How Do Humans and Agents Learn Differently?}}
\label{app:human-model-full}
\revised{Humans use the same games and interface as agents, with 21--30 participants per game. We compare both their scores and their recorded behavior. This subsection specifies participant selection, score construction, and the behavioral statistics used in the main text.}

\textbf{Participants and incentives.} We recruited college students to play the games. Participants did not use AI tools or code during gameplay. They were paid for participating, with additional rewards for achieving the highest scores.

\subsubsection{\revised{Selecting human records}}
\revised{For each game, we select the Top-1, Top-3, and Top-5 players by their highest observed score within the scoring window. All records must have a completed first episode. If Max is tied, we prefer records with at least two scored episodes, then break remaining ties by the first episode attaining Max, Mean, LG, and LS. This tie-breaking rule supports analysis of learning while preserving the selected Max values.}

\revised{Observed scores take precedence over replay values. Re-shuffled-game metrics use observed episode scores at their original episode indices.}

\subsubsection{\revised{Calculating scores and learning curves}}
\revised{Max averages the selected players' individual peaks. The other metrics use their episode-wise mean score curve. LG compares the last and first $\min(3,\lfloor n/2\rfloor)$ scores, LS fits a line at the original episode indices. The selection rule ensures at least two scored positions in every selected game-level mean curve. All four metrics cover all 20 games.}

\latestrevision{For reference, averaging across all eligible participants rather than selecting the Top-$k$ records gives Max $=43.6$, Mean $=31.2$, LG $=+10.4$, and LS $=+2.72$.}

\revised{Figure~\ref{fig:learning-comparisons}(C) compares human and agent learning curves across all 20 games. The score tables also cover all 20 games.}

\subsubsection{\revised{Comparing recorded behavior}}
\label{app:evidence-human-statistics}
\revised{The behavioral comparison uses Human Top-3 and the six Opus~5 methods on fixed games, within the same five- or ten-episode scoring windows. Human actions and scores come directly from completed episodes. Prior-best replay values do not contribute to the behavioral statistics. We pool the three agent trials for Naive, Memory, Reflexion, EvoTest, ReasoningBank, and AWM.}

\revised{\textbf{Action repetition.} We compare the ordered action sequences in consecutive episodes using SequenceMatcher with \texttt{autojunk=False}. Similarity is $2M/(|a|+|b|)$, where $M$ is the total length of matched blocks. The mean is 0.5379 across 150 human episode pairs and 0.6575 across 918 agent pairs. This measures repeated actions, not whether the actions represent the same strategy or good exploration.}

\revised{\textbf{Recovery after a setback.} For three consecutive observed episodes, we first identify a score decrease, $s_t<s_{t-1}$, then ask whether the next score exceeds every earlier observed score in that record. This happens after 13 of 39 human decreases (33.3\%) and 52 of 238 agent decreases (21.8\%). Both statistics average observations without game weighting. Repeated observations from the same participant or run are not independent, so these are descriptive comparisons rather than causal tests of willingness to explore.}

\revised{\textbf{Record selection.} Human behavior is measured from completed episodes with recorded actions and scores. Each agent trial must contain both actions and scores throughout the scoring window. If multiple records match a trial, we select the one with the most recorded scores, using the most recently modified record to break ties.}

\par\medskip\noindent\begin{minipage}{\linewidth}
\centering
\centering
\captionof{table}{\revised{Human and agent results ranked by Max. All four metrics use the same 20 games and challenge weights. Human selection and score construction are specified above. Bold marks the best value in each column.}}
\label{tab:rq2-human-full}
{\begin{tabular}{@{}llrrrr@{}}
\toprule
\textbf{System} & \textbf{Model} & \textbf{Max} & \textbf{Mean} & \textbf{LG} & \textbf{LS} \\
\midrule
Human Top-1 & Human & \revised{\textbf{84.3}} & \revised{53.5} & \revised{+21.6} & \revised{\textbf{+7.33}} \\
Claude Code & Claude Opus~5 & 81.8 & \revised{\textbf{62.3}} & \textbf{+27.1} & \revised{+6.59} \\
Human Top-3 & Human & \revised{79.1} & \revised{50.1} & \revised{+18.9} & \revised{+5.54} \\
Human Top-5 & Human & \revised{75.1} & \revised{47.1} & \revised{+17.1} & \revised{+4.62} \\
OpenCode & Claude Opus~5 & 74.5 & 53.8 & +21.3 & +5.13 \\
Codex & GPT-5.6-SOL & 74.3 & 54.8 & +22.2 & +5.66 \\
OpenCode & Qwen 3.8 Max & 72.5 & 51.1 & +25.2 & +6.30 \\
OpenCode & Gemini~3.5 Flash & 72.0 & 52.2 & +19.8 & +5.37 \\
OpenCode & Kimi~K3 & 68.5 & 50.2 & +17.1 & +4.24 \\
EvoTest & Claude Opus~5 & 66.1 & 51.3 & +9.4 & +2.52 \\
Reflexion & Claude Opus~5 & 64.8 & 48.8 & +11.6 & +2.93 \\
\textsc{Memory} & Kimi~K3 & 63.3 & 48.0 & +16.9 & +4.69 \\
EvoTest & Kimi~K3 & 62.4 & 48.0 & +1.8 & +1.01 \\
Reflexion & Kimi~K3 & 61.6 & 40.4 & +14.8 & +3.35 \\
\textsc{Memory} & Claude Opus~5 & 60.9 & 48.2 & +11.4 & +2.87 \\
ReasoningBank & Claude Opus~5 & 60.2 & 45.3 & +12.3 & +2.85 \\
ReasoningBank & Kimi~K3 & 59.9 & 40.1 & +8.3 & +2.59 \\
OpenCode & Gemini~3.7 Flash & 58.3 & 39.9 & +10.7 & +2.47 \\
OpenCode & GPT-5.6-SOL & 57.0 & 41.1 & +14.5 & +3.71 \\
AWM & Kimi~K3 & 56.1 & 37.0 & +4.8 & +0.62 \\
OpenCode & Claude Sonnet~5 & 55.9 & 35.6 & +15.7 & +3.72 \\
\textsc{Naive} & Kimi~K3 & 54.7 & 31.9 & +0.3 & +0.42 \\
OpenCode & DeepSeek V4 Pro & 54.2 & 38.2 & +14.7 & +3.66 \\
Reflexion & GPT-5 Mini & 53.5 & 34.4 & +3.6 & +0.97 \\
\textsc{Naive} & Claude Opus~5 & 53.4 & 36.6 & -5.1 & -1.08 \\
ReasoningBank & GPT-5 Mini & 51.9 & 32.5 & -0.4 & +0.00 \\
AWM & Claude Opus~5 & 50.7 & 37.8 & +0.7 & +0.43 \\
\textsc{Naive} & GPT-5 Mini & 48.7 & 30.5 & -0.1 & +0.36 \\
\textsc{Memory} & GPT-5 Mini & 48.7 & 37.3 & +5.9 & +1.56 \\
EvoTest & GPT-5 Mini & 47.5 & 31.8 & +7.4 & +1.61 \\
AWM & GPT-5 Mini & 47.0 & 28.9 & -2.9 & -0.98 \\
\latestrevision{Human Mean} & \latestrevision{Human} & \latestrevision{43.6} & \latestrevision{31.2} & \latestrevision{+10.4} & \latestrevision{+2.72} \\
OpenCode & GPT-5 Mini & 34.0 & 22.3 & +9.2 & +1.83 \\
\bottomrule
\end{tabular}
}
\end{minipage}\par\medskip

\subsubsection{\revised{Cases of execution and continued investigation}}
\label{app:evidence-human-cases}
\revised{\textbf{Combining clues in RedEye.} OpenCode--Opus~5 trial 1 first fails, then combines documentary evidence with information about the owner, engineer, and coast guard. Its notes specify the rescue order, and episodes 9 and 10 score 100. This illustrates executing a multi-step solution in one run.}

\revised{\textbf{Using programs in Model Auditor.} OpenCode--Opus~5 trial 1 switches to programmatic coefficient search in episode 4 and adds Gaussian elimination in episode 5 to test whether its fitted model is unique. Both episodes score 100. The fitted models are recorded hypotheses, not necessarily the true hidden implementation.}

\revised{\textbf{Reduced human probing.} In Model Auditor, the two highest-Max human records use 40, 17, 17, 1, 0 and 17, 40, 0, 0, 0 probe actions across episodes 1--5. These counts exclude resets and predictions. They show reduced testing, but do not identify whether the cause is effort, confidence, frustration, or something else.}

\Needspace{12\baselineskip}
\subsection{\latestrevision{RQ4: How Well Does Learned Knowledge Transfer to New Game Instances?}}
\label{app:transfer-full}
\textbf{Matched-game protocol.} Table~\ref{tab:transfer} reports the OpenCode backbones and deployed agents, while Table~\ref{tab:transfer-methods} reports the six methods with Opus~5. The methods include \textsc{Naive}, which retains no cross-episode memory. All configurations are compared on matched games from the ten-game re-shuffled set, using the same rule seeds in both conditions. Each same-instance control trial repeats instance~1 for ten episodes in an independent native session. The re-shuffled condition reuses archived ten-episode trials with changing instances. LS is fitted to the mean normalized trajectory within each game, then aggregated with the shared challenge weights. This supplementary comparison uses the full ten-episode window for both conditions, rather than the five-episode reporting window used for normal games elsewhere.

\textbf{Included trajectories.} Three independent trials are required per game and configuration. We calculate LS from complete ten-episode trajectories and use the same matched games in both conditions of each row. The Qwen row covers nine games. All other rows cover ten.

\begin{center}
\centering
\captionof{table}{\captionrevision{Learning slopes for six methods with Opus~5 when the same instance repeats (Control) or changes across episodes (Re-shuf.), with the rules held fixed. LS drop is Control minus Re-shuf.}}
\label{tab:transfer-methods}
\begin{tabular}{@{}lrrr@{}}
\toprule
\textbf{Method} & \textbf{Control} & \textbf{Re-shuf.} & \textbf{LS drop} \\
\midrule
\textsc{Naive} & -0.49 & -0.41 & -0.08 \\
\textsc{Memory} & 3.76 & 2.88 & 0.88 \\
Reflexion & 3.33 & 2.70 & 0.63 \\
ReasoningBank & 2.48 & 2.09 & 0.40 \\
AWM & -0.11 & 0.26 & -0.37 \\
EvoTest & 1.92 & 1.07 & 0.85 \\
\bottomrule
\end{tabular}

\end{center}

\textbf{Budget sensitivity.} Table~\ref{tab:transfer-full} compares the standard-budget control LS with a calculation that also includes nine extended-budget control trajectories. Both columns use the same games, so differences do not come from changes in which games are included.

\begin{center}
\centering
\captionof{table}{\captionrevision{Learning slopes in the same-instance control, excluding or including extended-budget trials. Both estimates use the same games. The extended-budget results are supplementary and do not replace the standard-budget estimates in Table~\ref{tab:transfer}.}}
\label{tab:transfer-full}
\begin{tabular}{lrr}
\toprule
\textbf{System / model} & \textbf{Control LS} & \textbf{Control LS incl. extra} \\
\midrule
OpenCode Claude Opus~5 & 4.10 & 4.07 \\
OpenCode Qwen 3.8 Max & 3.16 & 3.34 \\
OpenCode Kimi~K3 & 4.19 & 4.33 \\
\bottomrule
\end{tabular}

\end{center}

\subsubsection{\revised{Applying a rule to a new instance}}
\label{app:evidence-transfer}
The following cases illustrate rule reuse rather than estimate its frequency or explain aggregate differences in LS.

\revised{\textbf{Matching a rule to a new menu.} OpenCode--Opus~5 trial 1 recognizes that Poisoner's menus change while the king's preferences stay fixed. It compares accepted and rejected dishes, then selects dishes matching an inferred five-feature condition. It repeatedly scores 100 by applying the condition to new menus. This supplementary case draws on interactions beyond the scoring window and illustrates reuse of a condition rather than a dish position.}

\textbf{Using a learned clue rather than probing again.} In \game{SynergyCorp} with seed~2, trial~1, episode~10, three OpenCode agents start from the same game instance under a 30-action budget. Opus~5 uses a mapping learned in earlier episodes from visible desk traits to preferred actions, so it often selects an effective action immediately and scores 137. Sonnet~5 tests actions to find each new boss's preferences and scores 67. Gemini~3.5 Flash uses one action as a diagnostic probe for each boss and scores 84. The records illustrate how using an observed clue can save actions.
% Checked against results/synergycorp/opencode_{claude-opus-5,claude-sonnet-5,gemini-3.5-flash}_t1/seeds/seed2.{json,events.jsonl}.

\subsection{\latestrevision{RQ5: Do Higher Inference Costs Lead to Better Performance?}}
\label{app:cost-estimation}
\revised{We estimate cost from recorded token usage so that performance can be compared with the resources used to obtain it. The estimate distinguishes fresh input, cached input, and output tokens. \astrarevision{GPT-6 Astra instead uses the confirmed evaluation bill; other costs are standard-price estimates.}}

\subsubsection{\revised{Usage and pricing}}
\label{app:cost-usage}
\revised{We sum OpenCode's per-step token reports, use Claude Code's run-level usage summaries, and read Codex's final turn-completion report. For self-evolving methods, we use logged input and output counts and reconstruct cache reuse from repeated prompt prefixes, using a five-minute cache lifetime and each model's minimum cacheable length. Older runs use the observed prefix split for the same method. The EvoTest--Kimi~K3 estimate averages the available token logs.}

\revised{\opusrevision{\astrarevision{For models other than Opus~5.5 and GPT-6 Astra, prices use}} the 30 August 2026 snapshot~\citep{openrouter2026pricing,anthropic2026caching}, with separate rates for fresh input, cache reads, cache writes, and output. We report average cost and token usage per episode.}

\opusrevision{For Opus~5.5, we use \href{https://platform.claude.com/docs/en/about-claude/pricing}{Anthropic's standard API prices} accessed on 7 October 2026.}

\astrarevision{For GPT-6 Astra, the experiment owner confirmed a total evaluation bill of \$150 on 8 October 2026. Dividing this bill by 600 completed episodes gives \$0.25 per episode, as reported in Figure~\ref{fig:rq4-cost} and Table~\ref{tab:cost}. Charges for preflight, retries, and superseded attempts are not separately itemized in this total. This billed amount is distinct from the standard-price estimates for other configurations, so Astra is excluded from their cost frontier and efficiency comparisons.}

\astrarevision{The billing confirmation does not resolve missing token reports. Native session/message identifiers deduplicate event exports and database records, including continuations; 13 of 4,253 unique records lack usable token counts. The token total therefore remains a lower bound, and unknown usage is not treated as zero. The archived standard-price ledger retains its separate assumptions for reasoning, caching, and long-context rates; it is not used as the reported billed cost.}

Table~\ref{tab:kimi-token-breakdown} separates input from generation for the Kimi~K3 comparison. Input includes fresh input, cache reads, and cache writes. Generation includes visible output and reasoning. Separately reported reasoning tokens are added without double counting. Each token and call total is divided by the number of recorded episodes. The Updates column is the portion of generation spent on cross-episode reflection, strategy evolution, memory extraction, or workflow induction.

\begin{table}[!htbp]
\centering
\caption{Kimi~K3 token and call breakdown per episode. Generated tokens include reasoning. Updates are a subset of Generated. M and K denote millions and thousands of tokens. USD is the estimated total cost per episode.}
\label{tab:kimi-token-breakdown}
\appendixtable{{\begin{tabular}{@{}lrrrrrr@{}}
\toprule
System & Max & Input (M) & Generated (K) & Updates (K) & Calls & USD \\
\midrule
OpenCode & 68.5 & 0.55 & 7.2 & -- & 8.1 & 0.313 \\
\textsc{Memory} & 63.3 & 1.26 & 6.7 & -- & 27.0 & 0.586 \\
Reflexion & 61.6 & 0.38 & 9.1 & 2.6 & 29.2 & 0.300 \\
EvoTest & 62.4 & 0.14 & 24.6 & 11.5 & 24.1 & 0.658 \\
ReasoningBank & 59.9 & 0.34 & 12.3 & 3.0 & 29.8 & 0.340 \\
AWM & 56.1 & 0.26 & 7.3 & 1.8 & 28.4 & 0.224 \\
\bottomrule
\end{tabular}
}}
\end{table}

The savings differ by comparison. \textsc{Memory} supplies retained histories with each action request, giving it the largest input volume. EvoTest, Reflexion, ReasoningBank, and AWM generate additional updates between episodes. EvoTest's updates alone use 11.5K generated tokens per episode, exceeding OpenCode's total generation of 7.2K tokens. OpenCode uses fewer model calls: one \game{GemForge} trace batches nine game actions into a single call and returns only the last 60 output lines. Under the pricing snapshot, Kimi~K3 generation costs \$15 per million tokens versus \$3 for fresh input, yielding generation costs of \$0.108 for OpenCode, \$0.369 for EvoTest, and \$0.184 for ReasoningBank per episode. Reflexion and AWM have lower total estimated costs than OpenCode, alongside lower Max scores.

\begin{table}[!htbp]
\centering
\centering
\caption{\revised{\astrarevision{Cost comparison using average cost and token usage per episode; Astra uses the confirmed bill and other configurations use standard-price estimates.} Ratios divide Max by the displayed average cost or token usage per episode. Bold efficiency values are selected among configurations with Max $\geq70$. \astrarevision{$\geq$ marks incomplete token usage. Astra costs \$0.25 per episode (\$150/600); its efficiency ratios are omitted because token usage is incomplete and its billing basis differs.}}}
\label{tab:cost}
\appendixtable{{\begin{tabular}{@{}llrrrrr@{}}
\toprule
\textbf{System} & \textbf{Model} & \textbf{Max} & \textbf{Mtok/ep.} &
\textbf{USD/ep.} & \textbf{Max/(Mtok/ep.)} & \textbf{Max/(USD/ep.)} \\
\midrule
Claude Code & Claude Opus~5 & \textbf{81.8} & 0.71 & 0.647 & 115.7 & 126.4 \\
\opusrevision{OpenCode} & \opusrevision{Claude Opus~5.5} & \opusrevision{80.1} & \opusrevision{0.27} & \opusrevision{1.06} & \opusrevision{298.2} & \opusrevision{75.5} \\
\astrarevision{OpenCode} & \astrarevision{GPT-6 Astra} & \astrarevision{76.2} & \astrarevision{$\geq 0.279$} & \astrarevision{0.25} & \astrarevision{--} & \astrarevision{--} \\
OpenCode & Claude Opus~5 & 74.5 & 0.54 & 2.57 & 138.3 & 29.0 \\
Codex & GPT-5.6-SOL & 74.3 & 0.49 & 0.144 & 151.1 & \textbf{516.6} \\
OpenCode & Qwen 3.8 Max & 72.5 & 0.51 & 0.225 & 140.9 & 322.4 \\
OpenCode & Gemini~3.5 Flash & 72.0 & 1.63 & 0.598 & 44.3 & 120.5 \\
OpenCode & Kimi~K3 & 68.5 & 0.56 & 0.313 & 122.6 & 218.5 \\
EvoTest & Claude Opus~5 & 66.1 & 0.21 & 1.17 & 318.5 & 56.6 \\
Reflexion & Claude Opus~5 & 64.8 & 0.52 & 0.793 & 125.3 & 81.8 \\
\textsc{Memory} & Kimi~K3 & 63.3 & 1.27 & 0.586 & 50.0 & 107.9 \\
EvoTest & Kimi~K3 & 62.4 & 0.16 & 0.658 & \textbf{378.7} & 94.8 \\
Reflexion & Kimi~K3 & 61.6 & 0.39 & 0.300 & 158.6 & 205.5 \\
\textsc{Memory} & Claude Opus~5 & 60.9 & 3.52 & 3.09 & 17.3 & 19.7 \\
ReasoningBank & Claude Opus~5 & 60.2 & 0.56 & 0.858 & 108.3 & 70.2 \\
ReasoningBank & Kimi~K3 & 59.9 & 0.35 & 0.340 & 170.7 & 176.2 \\
OpenCode & Gemini~3.7 Flash & 58.3 & 0.33 & 0.052 & 175.5 & 1124.1 \\
OpenCode & GPT-5.6-SOL & 57.0 & 0.32 & 0.119 & 180.1 & 480.1 \\
AWM & Kimi~K3 & 56.1 & 0.26 & 0.224 & 213.3 & 250.8 \\
OpenCode & Claude Sonnet~5 & 55.9 & 0.73 & 1.33 & 77.1 & 41.9 \\
\textsc{Naive} & Kimi~K3 & 54.7 & 0.34 & 0.252 & 162.2 & 217.3 \\
OpenCode & DeepSeek V4 Pro & 54.2 & 0.23 & 0.021 & 234.3 & 2526.9 \\
Reflexion & GPT-5 Mini & 53.5 & 0.23 & 0.041 & 235.4 & 1319.6 \\
\textsc{Naive} & Claude Opus~5 & 53.4 & 0.71 & 0.800 & 75.6 & 66.8 \\
ReasoningBank & GPT-5 Mini & 51.9 & 0.52 & 0.090 & 100.2 & 575.7 \\
AWM & Claude Opus~5 & 50.7 & 0.33 & 0.423 & 153.2 & 119.8 \\
\textsc{Naive} & GPT-5 Mini & 48.7 & 0.17 & 0.037 & 294.6 & 1313.3 \\
\textsc{Memory} & GPT-5 Mini & 48.7 & 0.89 & 0.051 & 54.6 & 946.6 \\
EvoTest & GPT-5 Mini & 47.5 & 0.17 & 0.063 & 282.9 & 748.0 \\
AWM & GPT-5 Mini & 47.0 & 0.19 & 0.036 & 252.0 & 1323.4 \\
OpenCode & GPT-5 Mini & 34.0 & 0.10 & 0.0063 & 345.8 & 5437.8 \\
\bottomrule
\end{tabular}
}}
\end{table}

\clearpage
\section{Aggregate Results and Trial Variability}
\label{app:trial-variability}
% Generated by tools/gen_trial_variability.py.
\newcommand{\TrialVariabilityCells}{\astrarevision{620}}
\newcommand{\TrialVariabilityComplete}{\astrarevision{620}}
\newcommand{\TrialVariabilityTwo}{0}
\newcommand{\TrialVariabilityPartial}{0}

\textbf{We report trial variability for all \astrarevision{31} agent configurations and all four metrics.} Tables~\ref{tab:trial-variability-all}--\ref{tab:trial-variability-reshuffled} give the weighted results over all games and separately over fixed and re-shuffled games. Per-game results appear alongside each game's curves and rule evidence in Appendix~\ref{app:game-details}. Point estimates retain the normalization, scoring windows, and aggregation used elsewhere in the paper. The standard deviations (SDs) describe variation across independent trials, not variation across episodes or game difficulty.

\textbf{Within each game, we compute each metric separately for each complete trial.} Max takes that trial's highest score, Mean averages its scores, LG compares its first and last $k_g$ episodes, and LS fits its score trend against the original episode indices. For trial-level metric values $M_{g,r}$, the sample variance is $s_g^2=(n_g-1)^{-1}\sum_{r=1}^{n_g}(M_{g,r}-\bar M_g)^2$, and the tables report $\pm s_g$. With complete trajectories, averaging these trial-level metrics reproduces the paper's point estimates, apart from rounding of archived episode means.

\textbf{Across games, we propagate within-game variation using the paper's weights.} Writing $a_g=w_g/\sum_h w_h$, the reported aggregate SD is $s_{\mathrm{agg}}=\sqrt{\sum_g a_g^2s_g^2}$. Under independent runs across games, this estimates the variability of a benchmark score formed from one trial per game. It does not pair arbitrary trial indices across games or treat differences between games as trial noise. This is an SD, not the standard error or a confidence interval of the reported multi-trial mean. The instance panel and reference scores are held fixed.

\textbf{Each configuration--game cell contains three complete scored trajectories.} This applies to all \TrialVariabilityCells{} cells in these tables, with no missing episode scores imputed. The small number of trials limits the precision of the SD estimates.

\textbf{Human Top-$k$ groups are not independent repeat trials.} They select participants by achieved performance, so their descriptive results remain in Appendix~\ref{app:human-model-full}. We do not attach agent-trial SDs to those selected groups or interpret them as population-level uncertainty.

% Generated by tools/gen_trial_variability.py; do not edit numbers by hand.
\clearpage
\subsection{All 20 games}
\begin{table}[!htbp]
\centering
\caption{\captionrevision{Performance and learning over 20 games. Values are point estimates $\pm$ standard deviations estimated from within-game trial variability.}}
\label{tab:trial-variability-all}
\appendixtable{
\begin{tabular}{@{}llrrrr@{}}
\toprule
Backbone & Method & Max $\uparrow$ & Mean $\uparrow$ & LG $\uparrow$ & LS $\uparrow$ \\
\midrule
\opusrevision{Claude Opus~5.5} & \opusrevision{OpenCode} & \opusrevision{$80.1 \pm 2.7$} & \opusrevision{$61.0 \pm 2.4$} & \opusrevision{$+23.0 \pm 3.9$} & \opusrevision{$+5.22 \pm 0.83$} \\
\astrarevision{GPT-6 Astra} & \astrarevision{OpenCode} & \astrarevision{$76.2 \pm 5.5$} & \astrarevision{$56.1 \pm 5.2$} & \astrarevision{$+23.8 \pm 4.0$} & \astrarevision{$+5.88 \pm 1.13$} \\
Claude Opus~5 & OpenCode & $74.5 \pm 2.6$ & $53.8 \pm 2.2$ & $+21.3 \pm 3.1$ & $+5.13 \pm 0.71$ \\
Gemini~3.5 Flash & OpenCode & $72.0 \pm 3.9$ & $52.2 \pm 3.3$ & $+19.8 \pm 5.7$ & $+5.37 \pm 1.26$ \\
GPT-5.6-SOL & OpenCode & $57.0 \pm 3.6$ & $41.1 \pm 2.4$ & $+14.5 \pm 3.9$ & $+3.71 \pm 1.06$ \\
Claude Sonnet~5 & OpenCode & $55.9 \pm 3.7$ & $35.6 \pm 2.6$ & $+15.7 \pm 4.1$ & $+3.72 \pm 0.82$ \\
DeepSeek V4 Pro & OpenCode & $54.2 \pm 3.7$ & $38.2 \pm 4.0$ & $+14.7 \pm 3.6$ & $+3.66 \pm 1.09$ \\
Kimi~K3 & OpenCode & $68.5 \pm 3.8$ & $50.2 \pm 3.6$ & $+17.1 \pm 3.2$ & $+4.24 \pm 0.75$ \\
Qwen 3.8 Max & OpenCode & $72.5 \pm 4.3$ & $51.1 \pm 3.1$ & $+25.2 \pm 4.5$ & $+6.30 \pm 1.18$ \\
GPT-5 Mini & OpenCode & $34.0 \pm 3.2$ & $22.3 \pm 2.6$ & $+9.2 \pm 3.9$ & $+1.83 \pm 0.91$ \\
Gemini~3.7 Flash & OpenCode & $58.3 \pm 2.5$ & $39.9 \pm 2.9$ & $+10.7 \pm 3.4$ & $+2.47 \pm 0.84$ \\
\midrule
Claude Opus~5 & Claude Code & $81.8 \pm 4.4$ & $62.3 \pm 3.0$ & $+27.1 \pm 5.2$ & $+6.59 \pm 1.05$ \\
GPT-5.6-SOL & Codex & $74.3 \pm 4.1$ & $54.8 \pm 2.9$ & $+22.2 \pm 4.8$ & $+5.66 \pm 1.07$ \\
\midrule
Kimi~K3 & \textsc{Naive} & $54.7 \pm 3.1$ & $31.9 \pm 1.4$ & $+0.3 \pm 4.3$ & $+0.42 \pm 1.05$ \\
Kimi~K3 & Reflexion & $61.6 \pm 3.2$ & $40.4 \pm 2.8$ & $+14.8 \pm 4.7$ & $+3.35 \pm 1.19$ \\
Kimi~K3 & EvoTest & $62.4 \pm 3.5$ & $48.0 \pm 2.6$ & $+1.8 \pm 4.3$ & $+1.01 \pm 1.05$ \\
Kimi~K3 & \textsc{Memory} & $63.3 \pm 4.0$ & $48.0 \pm 3.0$ & $+16.9 \pm 4.0$ & $+4.69 \pm 0.96$ \\
Kimi~K3 & ReasoningBank & $59.9 \pm 3.6$ & $40.1 \pm 2.0$ & $+8.3 \pm 3.4$ & $+2.59 \pm 0.93$ \\
Kimi~K3 & AWM & $56.1 \pm 4.2$ & $37.0 \pm 2.4$ & $+4.8 \pm 4.4$ & $+0.62 \pm 1.23$ \\
\midrule
Opus~5 & \textsc{Naive} & $53.4 \pm 2.5$ & $36.6 \pm 0.8$ & $-5.1 \pm 2.8$ & $-1.08 \pm 0.65$ \\
Opus~5 & Reflexion & $64.8 \pm 2.3$ & $48.8 \pm 2.8$ & $+11.6 \pm 2.5$ & $+2.93 \pm 0.64$ \\
Opus~5 & EvoTest & $66.1 \pm 2.8$ & $51.3 \pm 2.3$ & $+9.4 \pm 3.5$ & $+2.52 \pm 0.99$ \\
Opus~5 & \textsc{Memory} & $60.9 \pm 3.9$ & $48.2 \pm 2.9$ & $+11.4 \pm 3.7$ & $+2.87 \pm 0.84$ \\
Opus~5 & ReasoningBank & $60.2 \pm 4.8$ & $45.3 \pm 3.1$ & $+12.3 \pm 3.9$ & $+2.85 \pm 0.63$ \\
Opus~5 & AWM & $50.7 \pm 2.6$ & $37.8 \pm 2.5$ & $+0.7 \pm 2.3$ & $+0.43 \pm 0.55$ \\
\midrule
GPT-5 Mini & \textsc{Naive} & $48.7 \pm 2.5$ & $30.5 \pm 1.4$ & $-0.1 \pm 2.8$ & $+0.36 \pm 0.68$ \\
GPT-5 Mini & Reflexion & $53.5 \pm 4.3$ & $34.4 \pm 2.6$ & $+3.6 \pm 3.5$ & $+0.97 \pm 0.93$ \\
GPT-5 Mini & EvoTest & $47.5 \pm 3.6$ & $31.8 \pm 1.9$ & $+7.4 \pm 2.6$ & $+1.61 \pm 0.63$ \\
GPT-5 Mini & \textsc{Memory} & $48.7 \pm 3.7$ & $37.3 \pm 2.6$ & $+5.9 \pm 3.4$ & $+1.56 \pm 0.82$ \\
GPT-5 Mini & ReasoningBank & $51.9 \pm 3.4$ & $32.5 \pm 1.9$ & $-0.4 \pm 3.6$ & $+0.00 \pm 1.03$ \\
GPT-5 Mini & AWM & $47.0 \pm 4.3$ & $28.9 \pm 2.2$ & $-2.9 \pm 2.5$ & $-0.98 \pm 0.52$ \\
\bottomrule
\end{tabular}}
\end{table}
\FloatBarrier
\clearpage
\subsection{Fixed games}
\begin{table}[!htbp]
\centering
\caption{\captionrevision{Performance and learning over 10 fixed games. Values are point estimates $\pm$ standard deviations estimated from within-game trial variability.}}
\label{tab:trial-variability-fixed}
\appendixtable{
\begin{tabular}{@{}llrrrr@{}}
\toprule
Backbone & Method & Max $\uparrow$ & Mean $\uparrow$ & LG $\uparrow$ & LS $\uparrow$ \\
\midrule
\opusrevision{Claude Opus~5.5} & \opusrevision{OpenCode} & \opusrevision{$76.1 \pm 4.5$} & \opusrevision{$57.2 \pm 2.7$} & \opusrevision{$+25.6 \pm 6.6$} & \opusrevision{$+4.98 \pm 1.30$} \\
\astrarevision{GPT-6 Astra} & \astrarevision{OpenCode} & \astrarevision{$77.0 \pm 7.0$} & \astrarevision{$53.2 \pm 5.1$} & \astrarevision{$+28.7 \pm 4.3$} & \astrarevision{$+5.77 \pm 0.84$} \\
Claude Opus~5 & OpenCode & $75.8 \pm 2.4$ & $55.2 \pm 2.0$ & $+19.9 \pm 4.4$ & $+3.97 \pm 0.92$ \\
Gemini~3.5 Flash & OpenCode & $71.3 \pm 5.7$ & $51.2 \pm 3.9$ & $+19.4 \pm 7.7$ & $+5.02 \pm 1.50$ \\
GPT-5.6-SOL & OpenCode & $57.1 \pm 5.3$ & $41.9 \pm 3.5$ & $+12.8 \pm 4.4$ & $+2.09 \pm 1.14$ \\
Claude Sonnet~5 & OpenCode & $60.8 \pm 4.5$ & $43.3 \pm 3.6$ & $+17.1 \pm 5.6$ & $+3.84 \pm 1.22$ \\
DeepSeek V4 Pro & OpenCode & $54.0 \pm 4.9$ & $44.2 \pm 5.4$ & $+10.1 \pm 3.4$ & $+2.32 \pm 1.04$ \\
Kimi~K3 & OpenCode & $64.3 \pm 3.2$ & $48.2 \pm 3.0$ & $+15.8 \pm 4.9$ & $+3.78 \pm 0.87$ \\
Qwen 3.8 Max & OpenCode & $75.1 \pm 5.6$ & $54.0 \pm 2.8$ & $+26.7 \pm 5.7$ & $+5.69 \pm 1.45$ \\
GPT-5 Mini & OpenCode & $33.1 \pm 3.7$ & $24.3 \pm 3.3$ & $+7.8 \pm 4.4$ & $+1.68 \pm 0.94$ \\
Gemini~3.7 Flash & OpenCode & $58.7 \pm 3.6$ & $37.8 \pm 4.4$ & $+15.5 \pm 5.1$ & $+2.68 \pm 1.00$ \\
\midrule
Claude Opus~5 & Claude Code & $83.4 \pm 6.7$ & $63.4 \pm 3.7$ & $+28.3 \pm 8.1$ & $+6.05 \pm 1.52$ \\
GPT-5.6-SOL & Codex & $73.2 \pm 6.4$ & $55.7 \pm 3.8$ & $+22.7 \pm 7.2$ & $+4.84 \pm 1.49$ \\
\midrule
Kimi~K3 & \textsc{Naive} & $49.7 \pm 3.5$ & $30.8 \pm 1.2$ & $-0.2 \pm 3.8$ & $+0.33 \pm 0.78$ \\
Kimi~K3 & Reflexion & $57.0 \pm 4.0$ & $37.6 \pm 2.1$ & $+13.8 \pm 5.3$ & $+2.70 \pm 1.21$ \\
Kimi~K3 & EvoTest & $55.1 \pm 5.3$ & $44.8 \pm 3.9$ & $+5.1 \pm 5.3$ & $+1.77 \pm 0.98$ \\
Kimi~K3 & \textsc{Memory} & $56.7 \pm 4.3$ & $42.8 \pm 2.7$ & $+15.8 \pm 5.9$ & $+4.04 \pm 1.32$ \\
Kimi~K3 & ReasoningBank & $55.1 \pm 3.8$ & $41.4 \pm 2.5$ & $+9.0 \pm 4.0$ & $+2.00 \pm 0.93$ \\
Kimi~K3 & AWM & $49.0 \pm 5.4$ & $34.6 \pm 3.0$ & $+8.4 \pm 6.2$ & $+1.72 \pm 1.64$ \\
\midrule
Opus~5 & \textsc{Naive} & $49.1 \pm 3.4$ & $34.3 \pm 1.1$ & $-4.0 \pm 4.8$ & $-0.41 \pm 1.09$ \\
Opus~5 & Reflexion & $55.5 \pm 2.9$ & $42.0 \pm 4.3$ & $+4.8 \pm 2.7$ & $+1.52 \pm 0.77$ \\
Opus~5 & EvoTest & $61.3 \pm 4.8$ & $48.9 \pm 3.9$ & $+11.3 \pm 4.0$ & $+2.19 \pm 0.99$ \\
Opus~5 & \textsc{Memory} & $56.9 \pm 3.6$ & $47.3 \pm 2.4$ & $+8.8 \pm 5.2$ & $+2.14 \pm 1.18$ \\
Opus~5 & ReasoningBank & $57.1 \pm 6.8$ & $45.3 \pm 3.8$ & $+13.4 \pm 4.9$ & $+2.87 \pm 0.71$ \\
Opus~5 & AWM & $47.6 \pm 4.4$ & $35.5 \pm 4.4$ & $+3.2 \pm 3.4$ & $+1.06 \pm 0.77$ \\
\midrule
GPT-5 Mini & \textsc{Naive} & $43.9 \pm 2.5$ & $32.1 \pm 1.1$ & $+0.7 \pm 2.0$ & $+0.40 \pm 0.48$ \\
GPT-5 Mini & Reflexion & $44.3 \pm 6.6$ & $31.3 \pm 3.5$ & $+1.4 \pm 4.8$ & $+0.26 \pm 1.03$ \\
GPT-5 Mini & EvoTest & $41.1 \pm 5.3$ & $33.1 \pm 2.2$ & $+3.5 \pm 2.2$ & $+1.08 \pm 0.58$ \\
GPT-5 Mini & \textsc{Memory} & $44.8 \pm 4.9$ & $38.3 \pm 3.4$ & $+6.9 \pm 5.1$ & $+1.36 \pm 0.86$ \\
GPT-5 Mini & ReasoningBank & $45.8 \pm 2.8$ & $31.0 \pm 2.2$ & $+1.3 \pm 3.8$ & $+0.54 \pm 0.91$ \\
GPT-5 Mini & AWM & $42.3 \pm 6.7$ & $29.7 \pm 3.4$ & $+0.6 \pm 3.5$ & $-0.30 \pm 0.74$ \\
\bottomrule
\end{tabular}}
\end{table}
\FloatBarrier
\clearpage
\subsection{Re-shuffled games}
\begin{table}[!htbp]
\centering
\caption{\captionrevision{Performance and learning over 10 re-shuffled games. Values are point estimates $\pm$ standard deviations estimated from within-game trial variability.}}
\label{tab:trial-variability-reshuffled}
\appendixtable{
\begin{tabular}{@{}llrrrr@{}}
\toprule
Backbone & Method & Max $\uparrow$ & Mean $\uparrow$ & LG $\uparrow$ & LS $\uparrow$ \\
\midrule
\opusrevision{Claude Opus~5.5} & \opusrevision{OpenCode} & \opusrevision{$84.7 \pm 2.6$} & \opusrevision{$65.3 \pm 4.2$} & \opusrevision{$+19.9 \pm 3.8$} & \opusrevision{$+5.49 \pm 0.99$} \\
\astrarevision{GPT-6 Astra} & \astrarevision{OpenCode} & \astrarevision{$75.4 \pm 8.8$} & \astrarevision{$59.3 \pm 9.5$} & \astrarevision{$+18.2 \pm 7.1$} & \astrarevision{$+6.01 \pm 2.22$} \\
Claude Opus~5 & OpenCode & $73.1 \pm 4.8$ & $52.2 \pm 4.1$ & $+22.9 \pm 4.3$ & $+6.45 \pm 1.10$ \\
Gemini~3.5 Flash & OpenCode & $72.8 \pm 5.3$ & $53.4 \pm 5.5$ & $+20.3 \pm 8.4$ & $+5.77 \pm 2.08$ \\
GPT-5.6-SOL & OpenCode & $56.9 \pm 5.0$ & $40.3 \pm 3.4$ & $+16.4 \pm 6.6$ & $+5.56 \pm 1.85$ \\
Claude Sonnet~5 & OpenCode & $50.4 \pm 6.0$ & $26.8 \pm 3.8$ & $+14.1 \pm 6.1$ & $+3.58 \pm 1.08$ \\
DeepSeek V4 Pro & OpenCode & $54.3 \pm 5.6$ & $31.3 \pm 5.8$ & $+19.9 \pm 6.8$ & $+5.20 \pm 2.01$ \\
Kimi~K3 & OpenCode & $73.3 \pm 7.3$ & $52.4 \pm 6.9$ & $+18.6 \pm 3.8$ & $+4.77 \pm 1.27$ \\
Qwen 3.8 Max & OpenCode & $69.6 \pm 6.6$ & $47.9 \pm 6.0$ & $+23.4 \pm 7.0$ & $+7.00 \pm 1.90$ \\
GPT-5 Mini & OpenCode & $35.1 \pm 5.4$ & $20.0 \pm 4.2$ & $+10.8 \pm 6.7$ & $+2.00 \pm 1.63$ \\
Gemini~3.7 Flash & OpenCode & $57.8 \pm 3.4$ & $42.3 \pm 3.6$ & $+5.3 \pm 4.3$ & $+2.22 \pm 1.38$ \\
\midrule
Claude Opus~5 & Claude Code & $79.9 \pm 5.6$ & $61.0 \pm 4.8$ & $+25.7 \pm 6.3$ & $+7.20 \pm 1.45$ \\
GPT-5.6-SOL & Codex & $75.5 \pm 4.9$ & $53.8 \pm 4.5$ & $+21.6 \pm 6.3$ & $+6.60 \pm 1.53$ \\
\midrule
Kimi~K3 & \textsc{Naive} & $60.5 \pm 5.3$ & $33.1 \pm 2.7$ & $+0.9 \pm 8.2$ & $+0.52 \pm 2.07$ \\
Kimi~K3 & Reflexion & $66.8 \pm 5.0$ & $43.7 \pm 5.4$ & $+16.0 \pm 8.1$ & $+4.10 \pm 2.15$ \\
Kimi~K3 & EvoTest & $70.7 \pm 4.2$ & $51.6 \pm 3.3$ & $-1.9 \pm 7.0$ & $+0.15 \pm 1.94$ \\
Kimi~K3 & \textsc{Memory} & $70.7 \pm 7.1$ & $54.0 \pm 5.6$ & $+18.2 \pm 5.1$ & $+5.44 \pm 1.39$ \\
Kimi~K3 & ReasoningBank & $65.3 \pm 6.2$ & $38.7 \pm 3.3$ & $+7.5 \pm 5.8$ & $+3.26 \pm 1.68$ \\
Kimi~K3 & AWM & $64.3 \pm 6.6$ & $39.8 \pm 4.0$ & $+0.6 \pm 6.3$ & $-0.63 \pm 1.84$ \\
\midrule
Opus~5 & \textsc{Naive} & $58.4 \pm 3.7$ & $39.3 \pm 1.3$ & $-6.3 \pm 2.5$ & $-1.84 \pm 0.60$ \\
Opus~5 & Reflexion & $75.5 \pm 3.7$ & $56.6 \pm 3.3$ & $+19.4 \pm 4.4$ & $+4.53 \pm 1.06$ \\
Opus~5 & EvoTest & $71.7 \pm 2.3$ & $54.0 \pm 2.4$ & $+7.3 \pm 5.9$ & $+2.89 \pm 1.80$ \\
Opus~5 & \textsc{Memory} & $65.5 \pm 7.3$ & $49.4 \pm 5.6$ & $+14.3 \pm 5.1$ & $+3.70 \pm 1.20$ \\
Opus~5 & ReasoningBank & $63.8 \pm 6.9$ & $45.4 \pm 5.1$ & $+11.0 \pm 6.1$ & $+2.83 \pm 1.09$ \\
Opus~5 & AWM & $54.2 \pm 2.4$ & $40.4 \pm 1.7$ & $-2.2 \pm 2.8$ & $-0.30 \pm 0.79$ \\
\midrule
GPT-5 Mini & \textsc{Naive} & $54.3 \pm 4.4$ & $28.6 \pm 2.7$ & $-1.0 \pm 5.5$ & $+0.31 \pm 1.35$ \\
GPT-5 Mini & Reflexion & $63.9 \pm 5.4$ & $38.0 \pm 3.8$ & $+6.2 \pm 5.2$ & $+1.77 \pm 1.61$ \\
GPT-5 Mini & EvoTest & $54.7 \pm 4.8$ & $30.3 \pm 3.1$ & $+11.9 \pm 5.0$ & $+2.20 \pm 1.18$ \\
GPT-5 Mini & \textsc{Memory} & $53.2 \pm 5.5$ & $36.1 \pm 4.1$ & $+4.8 \pm 4.4$ & $+1.78 \pm 1.46$ \\
GPT-5 Mini & ReasoningBank & $58.8 \pm 6.6$ & $34.3 \pm 3.3$ & $-2.3 \pm 6.4$ & $-0.61 \pm 1.94$ \\
GPT-5 Mini & AWM & $52.4 \pm 5.3$ & $27.9 \pm 2.6$ & $-6.8 \pm 3.5$ & $-1.76 \pm 0.72$ \\
\bottomrule
\end{tabular}}
\end{table}
\FloatBarrier

\clearpage

\FloatBarrier
% Generated by tools/gen_appendix_game_details.py.
\section{\revised{Per-Game Results and Rule-Discovery Evidence}}
\label{app:game-details}
\label{app:game-learning-curves}
\label{app:game-leaderboards}

\revised{Each entry brings together a game's learning task, raw-score curves, a single performance table with trial variability, and archived rule-discovery annotations. Scores and supporting evidence can therefore be checked in one place.}
\label{app:rq1-target-audit}

\revised{\textbf{Reading the curves.} Curves show episode-wise mean raw scores across trials. Their left, center, and right panels compare backbones, self-evolving methods, and harnesses. Each method curve averages its Kimi~K3, Opus~5, and GPT-5 Mini curves.}

\astrarevision{No rule-discovery labels are assigned to Opus~5.5 or GPT-6 Astra; score validation alone does not establish rule discovery.}

\textbf{Reading the performance tables.} All four metrics use normalized scores, with agent results reported as point estimates $\pm$ trial SD (Appendix~\ref{app:trial-variability}). Max averages individual trial peaks. Mean averages the episode-wise trial mean. First is the mean episode at which each trial first reaches its peak. The selected Human Top-1 record retains Max, Mean, and First. The symbol -- marks quantities not reported for this selected record, not zero. Curves remain in raw units, and each entry gives the reference score needed to convert between scales.

\revised{\textbf{Reading the rule-discovery labels.} Each game defines local targets T1--Tk. S means at least one available trial provides evidence for the whole listed criterion, P means partial evidence without a complete trial, and -- means insufficient evidence. S does not mean that all trials satisfy the target. We report these labels only within each game. The different archived criteria are not aggregated into overall percentages.}

\revised{\textbf{Scope of the rule annotations.} The labels provide supplementary qualitative evidence from interaction records. They do not measure rule discovery within the scoring window. The first five games use complete-rule criteria. Criteria for the other games also include successful execution supported by observed performance. Terse logs, especially for GPT-5.6-SOL, can leave a target unverified even when a successful policy uses it.}

\revised{These annotations describe evidence in interaction records, not additional score measurements. A high score alone does not establish every hidden rule. For Haunted Inn and Poisoner, the entries identify which seed-dependent rule and policy labels use the same seed as the scored trials.}

\subsection{\revised{PrimordialSoup}}
\label{app:game-primordialsoup}

\revised{\textbf{Learning task.} Discover reaction recipes and plan synthesis within a resource budget. Fixed game, scored seed 1, first 5 episodes, reference score $C=110$.}
\begin{center}
\includegraphics[width=0.95\linewidth]{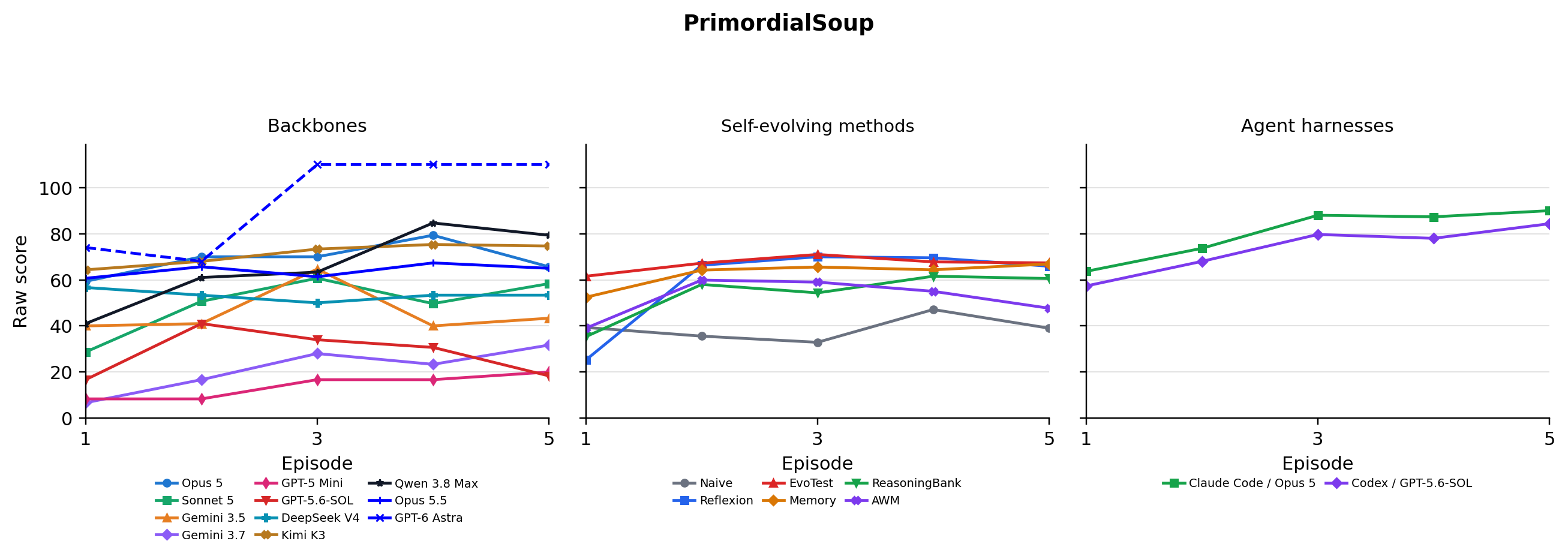}
\captionof{figure}{\revised{PrimordialSoup: raw-score learning curves within the scoring window.}}
\label{fig:game-curve-primordialsoup}
\label{fig:appendix-learning-curves}
\end{center}
\Needspace{10\baselineskip}
% Combined score table: normalized points and trial SD; human/First retained.
\begingroup
\setlength{\tabcolsep}{3pt}
\begin{longtable}{@{}lrrrrr@{}}
\caption{\captionrevision{PrimordialSoup: normalized performance and learning, with trial SD. Rows follow descending Max. First denotes the average episode at which a trial first reaches its highest score.}}
\label{tab:game-scores-primordialsoup}\label{tab:trial-variability-primordialsoup}\\
\toprule
Configuration & First & Max $\uparrow$ & Mean $\uparrow$ & LG $\uparrow$ & LS $\uparrow$ \\
\midrule
\endfirsthead
\toprule
Configuration & First & Max $\uparrow$ & Mean $\uparrow$ & LG $\uparrow$ & LS $\uparrow$ \\
\midrule
\endhead
Human Top-1 & 1 & 100.0 & 100.0 & -- & -- \\
\astrarevision{OpenCode + GPT-6 Astra} & \astrarevision{3} & \astrarevision{$100.0 \pm 0.0$} & \astrarevision{$85.8 \pm 2.2$} & \astrarevision{$+35.5 \pm 5.5$} & \astrarevision{$+10.36 \pm 1.14$} \\
Codex + GPT-5.6-SOL & 4 & $89.1 \pm 18.9$ & $66.8 \pm 15.2$ & $+16.8 \pm 33.3$ & $+5.82 \pm 10.89$ \\
Claude Code + Opus 5 & 3.7 & $81.8 \pm 15.7$ & $73.2 \pm 10.0$ & $+18.2 \pm 17.3$ & $+6.03 \pm 5.26$ \\
OpenCode + Kimi K3 & 4.7 & $78.5 \pm 19.1$ & $64.7 \pm 12.0$ & $+8.0 \pm 16.2$ & $+2.55 \pm 6.27$ \\
OpenCode + Qwen 3.8 Max & 4 & $77.0 \pm 16.0$ & $59.9 \pm 10.8$ & $+28.2 \pm 13.9$ & $+9.12 \pm 4.15$ \\
Reflexion + Opus 5 & 4.3 & $74.5 \pm 15.0$ & $53.1 \pm 5.1$ & $+23.2 \pm 14.6$ & $+8.64 \pm 5.21$ \\
OpenCode + Opus 5 & 4 & $72.1 \pm 1.0$ & $62.6 \pm 2.7$ & $+7.1 \pm 8.5$ & $+2.00 \pm 3.06$ \\
EvoTest + Opus 5 & 3 & $69.7 \pm 2.6$ & $65.5 \pm 1.2$ & $+4.8 \pm 2.2$ & $+2.06 \pm 1.00$ \\
Memory + Kimi K3 & 3.3 & $69.7 \pm 2.6$ & $64.4 \pm 1.6$ & $+8.8 \pm 3.5$ & $+3.09 \pm 1.18$ \\
EvoTest + Kimi K3 & 2.3 & $68.2 \pm 0.0$ & $65.8 \pm 1.1$ & $+6.1 \pm 2.7$ & $+2.30 \pm 0.93$ \\
ReasoningBank + Kimi K3 & 4.7 & $68.2 \pm 0.0$ & $54.6 \pm 3.7$ & $+25.6 \pm 14.7$ & $+9.33 \pm 4.78$ \\
Reflexion + Kimi K3 & 2.7 & $67.6 \pm 1.0$ & $56.4 \pm 6.7$ & $+24.4 \pm 15.1$ & $+9.67 \pm 5.74$ \\
ReasoningBank + Opus 5 & 4.3 & $67.0 \pm 2.1$ & $53.5 \pm 7.6$ & $+19.2 \pm 12.7$ & $+6.85 \pm 4.02$ \\
AWM + Opus 5 & 4.3 & $66.1 \pm 3.7$ & $51.3 \pm 8.7$ & $+8.3 \pm 14.0$ & $+2.61 \pm 4.30$ \\
AWM + Kimi K3 & 3 & $64.8 \pm 3.7$ & $45.9 \pm 17.4$ & $+2.4 \pm 12.9$ & $+2.15 \pm 3.64$ \\
Memory + Opus 5 & 3 & $63.6 \pm 15.7$ & $49.7 \pm 18.2$ & $+2.4 \pm 24.9$ & $+1.48 \pm 9.05$ \\
Reflexion + GPT-5 Mini & 3 & $62.7 \pm 8.7$ & $52.5 \pm 11.2$ & $+12.3 \pm 8.4$ & $+4.76 \pm 3.09$ \\
EvoTest + GPT-5 Mini & 3.3 & $62.1 \pm 5.6$ & $51.4 \pm 5.4$ & $-2.3 \pm 8.9$ & $-1.06 \pm 2.63$ \\
OpenCode + Sonnet 5 & 3.3 & $62.1 \pm 6.9$ & $45.1 \pm 1.9$ & $+13.0 \pm 22.5$ & $+5.30 \pm 6.32$ \\
\opusrevision{OpenCode + Opus 5.5} & \opusrevision{3.3} & \opusrevision{$61.8 \pm 3.3$} & \opusrevision{$58.2 \pm 3.7$} & \opusrevision{$+2.7 \pm 2.1$} & \opusrevision{$+0.94 \pm 0.64$} \\
Memory + GPT-5 Mini & 3.3 & $61.5 \pm 6.4$ & $56.9 \pm 3.0$ & $+8.6 \pm 15.4$ & $+3.33 \pm 5.94$ \\
OpenCode + Gemini 3.5 Flash & 3.7 & $60.3 \pm 10.8$ & $41.6 \pm 10.8$ & $+1.1 \pm 22.1$ & $+0.51 \pm 7.54$ \\
Naive + Kimi K3 & 3 & $57.9 \pm 4.3$ & $40.5 \pm 2.7$ & $+14.9 \pm 3.8$ & $+2.42 \pm 1.55$ \\
AWM + GPT-5 Mini & 2.3 & $51.8 \pm 15.9$ & $44.9 \pm 23.7$ & $-5.6 \pm 15.1$ & $-1.36 \pm 4.48$ \\
OpenCode + DeepSeek V4 Pro & 1 & $51.5 \pm 28.9$ & $48.5 \pm 34.1$ & $-1.5 \pm 2.6$ & $-0.61 \pm 1.05$ \\
Naive + Opus 5 & 2.3 & $50.9 \pm 11.4$ & $32.4 \pm 12.3$ & $-10.2 \pm 24.5$ & $-3.06 \pm 8.19$ \\
Naive + GPT-5 Mini & 2.7 & $50.3 \pm 20.7$ & $32.8 \pm 8.2$ & $+10.6 \pm 10.1$ & $+3.61 \pm 3.24$ \\
ReasoningBank + GPT-5 Mini & 1.7 & $46.1 \pm 18.6$ & $39.1 \pm 12.9$ & $-5.6 \pm 7.5$ & $-1.45 \pm 2.44$ \\
OpenCode + Gemini 3.7 Flash & 4 & $39.1 \pm 11.0$ & $19.3 \pm 5.8$ & $+14.4 \pm 3.5$ & $+5.15 \pm 0.95$ \\
OpenCode + GPT-5.6-SOL & 2 & $37.3 \pm 17.6$ & $25.6 \pm 15.7$ & $-3.9 \pm 1.5$ & $-0.64 \pm 1.43$ \\
OpenCode + GPT-5 Mini & 4 & $22.7 \pm 19.8$ & $12.7 \pm 16.6$ & $+9.1 \pm 16.4$ & $+2.88 \pm 5.31$ \\
\bottomrule
\end{longtable}
\endgroup
\begin{longtable}{@{}p{0.07\linewidth}p{0.87\linewidth}@{}}
\caption{\captionrevision{\revised{PrimordialSoup: criteria for assessing rule discovery.}}}\label{tab:targets-definition-primordialsoup}\\
\toprule
\textbf{Target} & \textbf{Criterion} \\
\midrule
\endfirsthead
\toprule
\textbf{Target} & \textbf{Criterion (continued)} \\
\midrule
\endhead
T1 & \revised{Both base reactions: $C+E\to\mathrm{L1a}$ and $D+E\to\mathrm{L1b}$.} \\
T2 & \revised{A and F are inert, and one react action fires each feasible base recipe at most once.} \\
T3 & \revised{React consumes B and sets the catalyst level to $\min(B,3)$.} \\
T4 & \revised{Catalyst persists through synthesis but is overwritten by the next react action.} \\
T5 & \revised{Catalyst level at least one enables $\mathrm{L1a}+\mathrm{L1b}\to\mathrm{L2b}$.} \\
T6 & \revised{Free $E+\mathrm{L1a}\to\mathrm{L2a}$ does not consume catalyst and takes priority over L2b.} \\
T7 & \revised{Catalyst level at least two enables $\mathrm{L2a}+\mathrm{L2b}\to\mathrm{L3}$.} \\
T8 & \revised{Only held molecules score, and the step penalty begins after step 30.} \\
\bottomrule
\end{longtable}
\begin{center}
\begin{minipage}{\linewidth}
\centering
\captionof{table}{\captionrevision{\revised{PrimordialSoup: rule-discovery evidence. S: complete evidence in at least one trial; P: partial evidence only; --: insufficient evidence. Evidence may include episodes beyond the scoring window.}}}
\label{tab:rq1-targets-game-primordialsoup}
{
\appendixtable{%
\begin{tabular}{@{}l*{8}{c}@{}}
\toprule
\textbf{Configuration} & T1 & T2 & T3 & T4 & T5 & T6 & T7 & T8 \\
\midrule
\multicolumn{9}{@{}l}{\textbf{backbone comparison}} \\
Claude Opus~5 & S & P & S & S & S & P & S & S \\
Qwen 3.8 Max & S & P & S & S & S & S & S & S \\
Kimi~K3 & S & S & S & S & S & S & S & S \\
Gemini~3.5 Flash & S & S & S & S & S & -- & -- & P \\
Gemini~3.7 Flash & S & -- & P & -- & S & -- & -- & -- \\
GPT-5.6-SOL & S & -- & P & -- & S & -- & S & -- \\
DeepSeek V4 Pro & S & P & P & P & S & P & P & P \\
Claude Sonnet~5 & S & P & P & S & S & P & S & P \\
GPT-5 Mini & P & S & P & P & P & P & P & P \\
\midrule
\multicolumn{9}{@{}l}{\textbf{agent harness comparison}} \\
Claude Code + Opus~5 & S & P & S & S & S & S & S & S \\
Codex + GPT-5.6-SOL & S & P & S & -- & S & P & S & S \\
\bottomrule
\end{tabular}
}
}

\end{minipage}
\end{center}
\noindent \textbf{The main separation is whether a system moves beyond the common L2b plateau to recover the alternative L2a/L3 branch within at least one trial.} Kimi~K3 recovers every target in at least one trial, and Qwen 3.8 Max and several Claude-based systems nearly do so. Weaker configurations typically retain gaps in T6--T7 or the catalyst-state account.

\subsection{\revised{GemForge}}
\label{app:game-gemforge}

\revised{\textbf{Learning task.} Discover multi-tier recipes and manage limiting ingredients. Fixed game, scored seed 1, first 5 episodes, reference score $C=791$.}
\begin{center}
\includegraphics[width=0.95\linewidth]{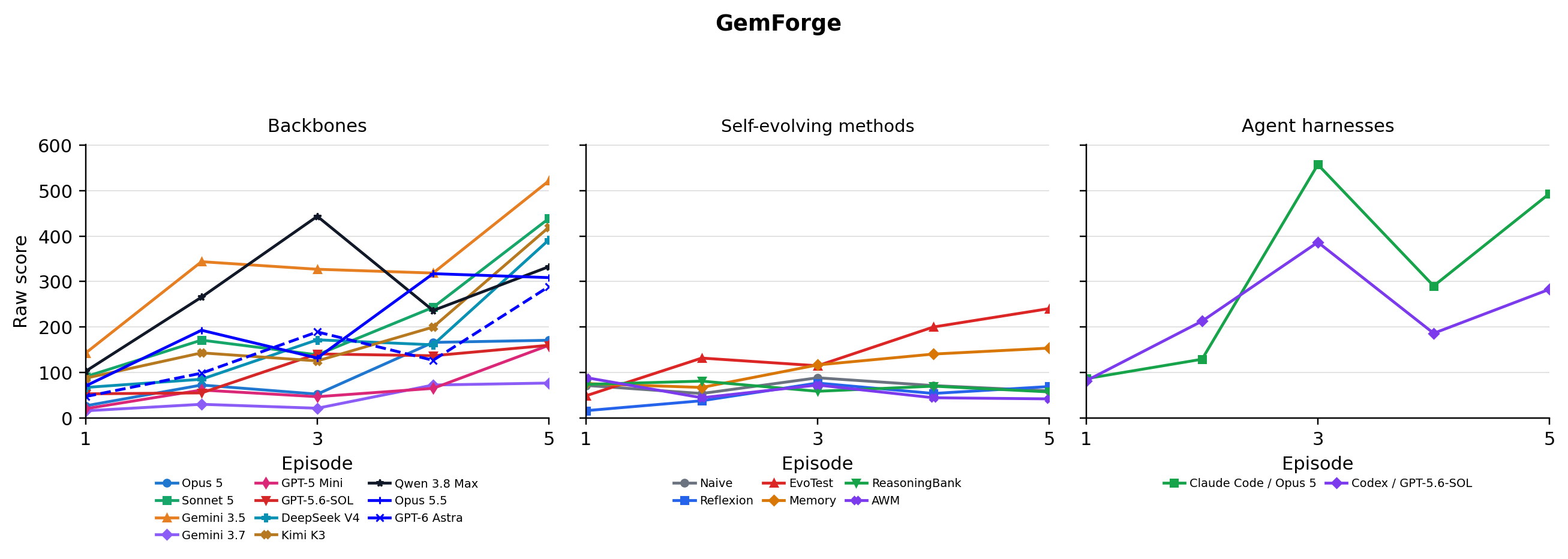}
\captionof{figure}{\revised{GemForge: raw-score learning curves within the scoring window.}}
\label{fig:game-curve-gemforge}

\end{center}
\Needspace{10\baselineskip}
% Combined score table: normalized points and trial SD; human/First retained.
\begingroup
\setlength{\tabcolsep}{3pt}
\begin{longtable}{@{}lrrrrr@{}}
\caption{\captionrevision{GemForge: normalized performance and learning, with trial SD. Rows follow descending Max. First denotes the average episode at which a trial first reaches its highest score.}}
\label{tab:game-scores-gemforge}\label{tab:trial-variability-gemforge}\\
\toprule
Configuration & First & Max $\uparrow$ & Mean $\uparrow$ & LG $\uparrow$ & LS $\uparrow$ \\
\midrule
\endfirsthead
\toprule
Configuration & First & Max $\uparrow$ & Mean $\uparrow$ & LG $\uparrow$ & LS $\uparrow$ \\
\midrule
\endhead
Human Top-1 & 4 & 99.9 & 59.4 & -- & -- \\
Claude Code + Opus 5 & 3.7 & $93.9 \pm 9.3$ & $39.4 \pm 12.7$ & $+35.9 \pm 28.1$ & $+12.32 \pm 8.33$ \\
EvoTest + Kimi K3 & 3 & $81.8 \pm 30.8$ & $46.9 \pm 11.8$ & $+46.0 \pm 37.0$ & $+16.09 \pm 8.99$ \\
OpenCode + Gemini 3.5 Flash & 3 & $72.6 \pm 45.4$ & $41.8 \pm 32.8$ & $+22.3 \pm 32.2$ & $+9.25 \pm 11.39$ \\
OpenCode + Qwen 3.8 Max & 4 & $63.8 \pm 31.7$ & $34.9 \pm 18.7$ & $+12.7 \pm 21.7$ & $+5.44 \pm 3.85$ \\
Codex + GPT-5.6-SOL & 4.3 & $61.8 \pm 36.4$ & $29.1 \pm 18.7$ & $+11.0 \pm 21.1$ & $+4.75 \pm 7.40$ \\
OpenCode + Sonnet 5 & 3.7 & $58.4 \pm 39.3$ & $27.4 \pm 15.6$ & $+26.5 \pm 32.0$ & $+9.70 \pm 11.20$ \\
OpenCode + DeepSeek V4 Pro & 4 & $53.6 \pm 43.0$ & $22.2 \pm 13.7$ & $+25.3 \pm 22.7$ & $+9.16 \pm 9.31$ \\
OpenCode + Kimi K3 & 5 & $53.0 \pm 0.7$ & $24.7 \pm 1.5$ & $+24.5 \pm 3.7$ & $+9.10 \pm 1.22$ \\
\opusrevision{OpenCode + Opus 5.5} & \opusrevision{4.3} & \opusrevision{$50.1 \pm 8.5$} & \opusrevision{$25.8 \pm 7.8$} & \opusrevision{$+22.9 \pm 7.9$} & \opusrevision{$+7.59 \pm 2.14$} \\
\astrarevision{OpenCode + GPT-6 Astra} & \astrarevision{4.3} & \astrarevision{$38.0 \pm 10.3$} & \astrarevision{$19.0 \pm 8.1$} & \astrarevision{$+17.1 \pm 8.4$} & \astrarevision{$+6.47 \pm 2.75$} \\
Memory + GPT-5 Mini & 2.7 & $32.1 \pm 13.8$ & $25.6 \pm 8.0$ & $+9.8 \pm 9.6$ & $+3.12 \pm 3.11$ \\
OpenCode + Opus 5 & 4 & $32.1 \pm 21.1$ & $12.4 \pm 4.2$ & $+15.0 \pm 16.6$ & $+4.83 \pm 5.93$ \\
Memory + Kimi K3 & 3.7 & $28.6 \pm 38.3$ & $8.2 \pm 10.3$ & $+14.6 \pm 24.2$ & $+5.50 \pm 8.78$ \\
Naive + GPT-5 Mini & 3 & $28.5 \pm 9.4$ & $13.2 \pm 0.5$ & $+3.3 \pm 9.5$ & $+1.33 \pm 1.79$ \\
OpenCode + GPT-5.6-SOL & 4.3 & $24.7 \pm 18.2$ & $13.8 \pm 8.8$ & $+11.9 \pm 12.6$ & $+3.73 \pm 4.62$ \\
AWM + GPT-5 Mini & 1.3 & $24.5 \pm 18.4$ & $13.3 \pm 12.8$ & $-6.7 \pm 8.2$ & $-1.93 \pm 3.95$ \\
Naive + Opus 5 & 3.7 & $24.4 \pm 24.8$ & $7.6 \pm 4.8$ & $-4.2 \pm 17.5$ & $-1.91 \pm 6.64$ \\
OpenCode + GPT-5 Mini & 3.3 & $20.2 \pm 22.0$ & $9.0 \pm 5.2$ & $+9.0 \pm 10.8$ & $+3.57 \pm 4.26$ \\
Naive + Kimi K3 & 3 & $20.1 \pm 12.0$ & $5.4 \pm 3.8$ & $+2.0 \pm 4.9$ & $+0.34 \pm 1.28$ \\
OpenCode + Gemini 3.7 Flash & 3.7 & $19.3 \pm 11.6$ & $5.5 \pm 2.5$ & $+6.5 \pm 6.5$ & $+2.08 \pm 1.85$ \\
ReasoningBank + GPT-5 Mini & 2.3 & $18.4 \pm 2.1$ & $8.0 \pm 3.0$ & $-5.5 \pm 7.4$ & $-1.32 \pm 3.29$ \\
AWM + Kimi K3 & 3.7 & $17.8 \pm 3.9$ & $6.3 \pm 0.6$ & $+4.5 \pm 3.2$ & $+1.03 \pm 0.74$ \\
ReasoningBank + Opus 5 & 3 & $17.8 \pm 7.8$ & $11.0 \pm 4.2$ & $+5.4 \pm 11.9$ & $+1.67 \pm 3.78$ \\
Memory + Opus 5 & 3.3 & $17.4 \pm 24.9$ & $8.2 \pm 11.5$ & $+4.2 \pm 6.1$ & $+0.06 \pm 0.28$ \\
Reflexion + Kimi K3 & 3.7 & $15.8 \pm 8.9$ & $4.9 \pm 2.1$ & $+1.3 \pm 1.7$ & $+0.58 \pm 0.60$ \\
Reflexion + GPT-5 Mini & 3.3 & $15.5 \pm 20.7$ & $10.3 \pm 14.9$ & $+8.7 \pm 16.8$ & $+2.90 \pm 5.61$ \\
AWM + Opus 5 & 1 & $13.3 \pm 11.3$ & $2.7 \pm 2.3$ & $-6.6 \pm 5.7$ & $-2.65 \pm 2.27$ \\
ReasoningBank + Kimi K3 & 1.7 & $11.0 \pm 3.2$ & $7.0 \pm 3.6$ & $-5.1 \pm 1.4$ & $-2.02 \pm 0.19$ \\
EvoTest + Opus 5 & 3.7 & $10.7 \pm 0.4$ & $8.8 \pm 0.6$ & $+3.3 \pm 2.6$ & $+1.03 \pm 0.58$ \\
Reflexion + Opus 5 & 3.7 & $6.5 \pm 1.8$ & $4.1 \pm 2.0$ & $+3.1 \pm 1.9$ & $+1.15 \pm 0.63$ \\
EvoTest + GPT-5 Mini & 1.7 & $0.9 \pm 1.5$ & $0.2 \pm 0.3$ & $+0.0 \pm 0.0$ & $+0.00 \pm 0.00$ \\
\bottomrule
\end{longtable}
\endgroup
\begin{longtable}{@{}p{0.07\linewidth}p{0.87\linewidth}@{}}
\caption{\captionrevision{\revised{GemForge: criteria for assessing rule discovery.}}}\label{tab:targets-definition-gemforge}\\
\toprule
\textbf{Target} & \textbf{Criterion} \\
\midrule
\endfirsthead
\toprule
\textbf{Target} & \textbf{Criterion (continued)} \\
\midrule
\endhead
T1 & \revised{Complete five-recipe map from raw materials to tier-1 gems.} \\
T2 & \revised{Complete four-recipe map from tier-1 to tier-2 gems.} \\
T3 & \revised{Complete two-recipe map from tier-2 to tier-3 gems.} \\
T4 & \revised{Every nonrecipe pair consumes its inputs and yields slag. No three-input recipe exists.} \\
T5 & \revised{Final inventory determines score. Solo firing yields one-point slag. Ash and crystal do not advance to higher tiers.} \\
\bottomrule
\end{longtable}
\begin{center}
\begin{minipage}{\linewidth}
\centering
\captionof{table}{\captionrevision{\revised{GemForge: rule-discovery evidence. S: complete evidence in at least one trial; P: partial evidence only; --: insufficient evidence. Evidence may include episodes beyond the scoring window.}}}
\label{tab:rq1-targets-game-gemforge}
{
\appendixtable{%
\begin{tabular}{@{}l*{5}{c}@{}}
\toprule
\textbf{Configuration} & T1 & T2 & T3 & T4 & T5 \\
\midrule
\multicolumn{6}{@{}l}{\textbf{backbone comparison}} \\
Claude Opus~5 & S & S & S & S & S \\
Qwen 3.8 Max & S & S & S & P & P \\
Kimi~K3 & S & S & P & P & P \\
Gemini~3.5 Flash & S & S & P & S & P \\
Gemini~3.7 Flash & S & P & P & P & P \\
GPT-5.6-SOL & P & P & P & P & P \\
DeepSeek V4 Pro & S & P & P & P & P \\
Claude Sonnet~5 & S & S & P & P & P \\
GPT-5 Mini & P & P & P & P & P \\
\midrule
\multicolumn{6}{@{}l}{\textbf{agent harness comparison}} \\
Claude Code + Opus~5 & S & S & S & S & S \\
Codex + GPT-5.6-SOL & S & P & P & P & P \\
\bottomrule
\end{tabular}
}
}

\end{minipage}
\end{center}
\noindent \textbf{High scores can be reached with only a partial recipe graph, so complete-rule recovery separates the systems more sharply than execution alone.} Claude Opus~5 and Claude Code + Opus~5 each recover every target in at least one trial, while most other configurations discover useful chains without establishing every tier-2 and tier-3 alternative.

\subsection{\revised{Catnip}}
\label{app:game-catnip}

\revised{\textbf{Learning task.} Learn action--target matches and when to exit for the best reward. Fixed game, scored seed 1, first 10 episodes, reference score $C=320$.}
\begin{center}
\includegraphics[width=0.95\linewidth]{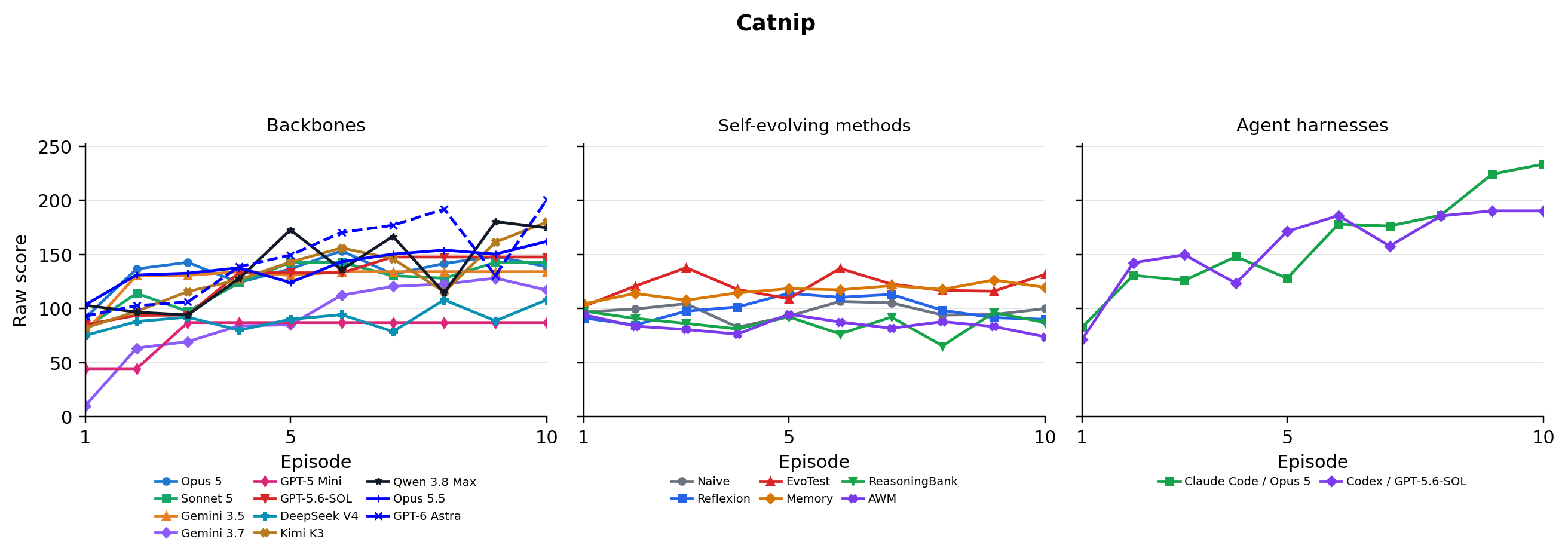}
\captionof{figure}{\revised{Catnip: raw-score learning curves within the scoring window.}}
\label{fig:game-curve-catnip}

\end{center}
\Needspace{10\baselineskip}
% Combined score table: normalized points and trial SD; human/First retained.
\begingroup
\setlength{\tabcolsep}{3pt}
\begin{longtable}{@{}lrrrrr@{}}
\caption{\captionrevision{Catnip: normalized performance and learning, with trial SD. Rows follow descending Max. First denotes the average episode at which a trial first reaches its highest score.}}
\label{tab:game-scores-catnip}\label{tab:trial-variability-catnip}\\
\toprule
Configuration & First & Max $\uparrow$ & Mean $\uparrow$ & LG $\uparrow$ & LS $\uparrow$ \\
\midrule
\endfirsthead
\toprule
Configuration & First & Max $\uparrow$ & Mean $\uparrow$ & LG $\uparrow$ & LS $\uparrow$ \\
\midrule
\endhead
Claude Code + Opus 5 & 9 & $73.0 \pm 1.2$ & $50.4 \pm 2.3$ & $+31.8 \pm 13.2$ & $+4.64 \pm 1.41$ \\
Human Top-1 & 8 & 70.3 & 53.5 & -- & -- \\
\astrarevision{OpenCode + GPT-6 Astra} & \astrarevision{6.3} & \astrarevision{$62.7 \pm 33.4$} & \astrarevision{$45.6 \pm 22.0$} & \astrarevision{$+23.1 \pm 18.8$} & \astrarevision{$+3.28 \pm 2.64$} \\
EvoTest + Kimi K3 & 3 & $59.8 \pm 5.5$ & $44.1 \pm 11.6$ & $-5.3 \pm 5.1$ & $-0.09 \pm 0.71$ \\
Codex + GPT-5.6-SOL & 4.3 & $59.5 \pm 5.9$ & $49.0 \pm 6.7$ & $+21.1 \pm 10.4$ & $+3.23 \pm 1.51$ \\
Reflexion + Kimi K3 & 7.3 & $56.6 \pm 5.4$ & $37.8 \pm 1.7$ & $+10.8 \pm 16.7$ & $+1.58 \pm 2.15$ \\
OpenCode + Qwen 3.8 Max & 7.3 & $56.4 \pm 10.3$ & $42.7 \pm 5.2$ & $+18.3 \pm 5.7$ & $+2.68 \pm 0.78$ \\
OpenCode + Kimi K3 & 8.7 & $56.2 \pm 13.0$ & $41.4 \pm 7.4$ & $+16.8 \pm 9.0$ & $+2.65 \pm 1.43$ \\
EvoTest + Opus 5 & 8 & $54.7 \pm 10.0$ & $39.3 \pm 0.7$ & $+3.5 \pm 7.2$ & $+0.53 \pm 1.18$ \\
\opusrevision{OpenCode + Opus 5.5} & \opusrevision{6.3} & \opusrevision{$50.6 \pm 1.6$} & \opusrevision{$43.4 \pm 2.1$} & \opusrevision{$+10.3 \pm 3.7$} & \opusrevision{$+1.57 \pm 0.52$} \\
OpenCode + Opus 5 & 6 & $50.3 \pm 3.0$ & $42.0 \pm 3.6$ & $+6.0 \pm 4.3$ & $+1.03 \pm 0.74$ \\
AWM + Kimi K3 & 2.3 & $47.7 \pm 8.5$ & $36.3 \pm 2.2$ & $-1.8 \pm 3.5$ & $-0.28 \pm 0.32$ \\
ReasoningBank + Kimi K3 & 5.7 & $47.0 \pm 5.6$ & $33.7 \pm 1.7$ & $-5.7 \pm 2.5$ & $-0.38 \pm 0.37$ \\
Memory + Opus 5 & 4.7 & $46.2 \pm 5.3$ & $39.4 \pm 1.7$ & $+6.4 \pm 1.7$ & $+0.80 \pm 0.30$ \\
OpenCode + GPT-5.6-SOL & 5 & $46.1 \pm 3.3$ & $39.4 \pm 6.2$ & $+17.7 \pm 2.7$ & $+2.38 \pm 0.66$ \\
Memory + Kimi K3 & 5.3 & $46.0 \pm 2.8$ & $39.9 \pm 3.6$ & $+4.5 \pm 3.1$ & $+0.75 \pm 0.44$ \\
OpenCode + Sonnet 5 & 3.7 & $45.2 \pm 9.0$ & $38.9 \pm 7.2$ & $+12.5 \pm 6.9$ & $+1.74 \pm 0.86$ \\
Naive + GPT-5 Mini & 5.3 & $43.3 \pm 6.3$ & $28.5 \pm 2.3$ & $-3.4 \pm 6.2$ & $+0.10 \pm 0.55$ \\
OpenCode + Gemini 3.5 Flash & 2.3 & $41.9 \pm 5.1$ & $39.8 \pm 5.4$ & $+6.5 \pm 5.7$ & $+1.03 \pm 0.90$ \\
Naive + Opus 5 & 3 & $40.6 \pm 11.3$ & $30.2 \pm 1.4$ & $-0.2 \pm 4.3$ & $-0.04 \pm 0.58$ \\
Memory + GPT-5 Mini & 3.7 & $40.6 \pm 0.5$ & $29.5 \pm 4.1$ & $+0.8 \pm 11.9$ & $+0.11 \pm 1.61$ \\
OpenCode + Gemini 3.7 Flash & 5.7 & $40.6 \pm 10.0$ & $28.5 \pm 13.8$ & $+23.4 \pm 4.1$ & $+3.44 \pm 0.80$ \\
ReasoningBank + Opus 5 & 4 & $39.9 \pm 6.1$ & $33.3 \pm 4.0$ & $+2.9 \pm 7.4$ & $+0.36 \pm 1.05$ \\
Reflexion + Opus 5 & 4.7 & $39.9 \pm 3.3$ & $34.2 \pm 3.3$ & $+0.8 \pm 3.6$ & $+0.01 \pm 0.34$ \\
Naive + Kimi K3 & 7.7 & $39.3 \pm 4.4$ & $32.8 \pm 2.6$ & $-0.2 \pm 2.9$ & $+0.07 \pm 0.50$ \\
EvoTest + GPT-5 Mini & 8.3 & $35.8 \pm 1.4$ & $30.1 \pm 1.7$ & $+3.0 \pm 2.2$ & $+0.53 \pm 0.30$ \\
OpenCode + DeepSeek V4 Pro & 3.7 & $33.8 \pm 6.0$ & $28.2 \pm 10.1$ & $+5.2 \pm 2.6$ & $+0.72 \pm 0.59$ \\
AWM + Opus 5 & 6 & $32.7 \pm 1.6$ & $28.2 \pm 0.9$ & $+0.4 \pm 4.2$ & $-0.00 \pm 0.53$ \\
Reflexion + GPT-5 Mini & 3.7 & $32.6 \pm 17.8$ & $21.0 \pm 11.5$ & $-9.7 \pm 15.5$ & $-1.20 \pm 1.71$ \\
ReasoningBank + GPT-5 Mini & 2.7 & $28.2 \pm 3.6$ & $14.2 \pm 6.5$ & $-5.3 \pm 8.1$ & $-0.80 \pm 1.30$ \\
OpenCode + GPT-5 Mini & 1.7 & $27.2 \pm 11.7$ & $24.5 \pm 7.4$ & $+8.9 \pm 15.4$ & $+1.29 \pm 2.24$ \\
AWM + GPT-5 Mini & 1.7 & $24.1 \pm 12.9$ & $14.6 \pm 12.9$ & $-2.9 \pm 4.9$ & $-0.52 \pm 0.53$ \\
\bottomrule
\end{longtable}
\endgroup
\begin{longtable}{@{}p{0.07\linewidth}p{0.87\linewidth}@{}}
\caption{\captionrevision{\revised{Catnip: criteria for assessing rule discovery.}}}\label{tab:targets-definition-catnip}\\
\toprule
\textbf{Target} & \textbf{Criterion} \\
\midrule
\endfirsthead
\toprule
\textbf{Target} & \textbf{Criterion (continued)} \\
\midrule
\endhead
T1 & \revised{The six targets occur in a fixed order.} \\
T2 & \revised{Complete mapping from all six targets to their sensory preferences.} \\
T3 & \revised{Complete mapping of soothing and chaotic style preferences.} \\
T4 & \revised{The five influence actions have distinct modalities and styles, and matching them changes influence gain.} \\
T5 & \revised{Conversion uses target-specific LOW, MED, and HIGH reward thresholds.} \\
T6 & \revised{Complete breed routing: Ragdoll to cuteness, Bobcat to menace, Black cat to stealth, and Orange to energy.} \\
T7 & \revised{Rewards trade immediate progress against attributes and items that affect later conversions.} \\
T8 & \revised{High stress or suspicion makes late mistakes costly.} \\
T9 & \revised{Items strengthen particular actions, control risk, or restore energy.} \\
T10 & \revised{Each unused turn adds 0.1 to the final multiplier, creating an early-exit tradeoff.} \\
\bottomrule
\end{longtable}
\begin{center}
\begin{minipage}{\linewidth}
\centering
\captionof{table}{\captionrevision{\revised{Catnip: rule-discovery evidence. S: complete evidence in at least one trial; P: partial evidence only; --: insufficient evidence. Evidence may include episodes beyond the scoring window.}}}
\label{tab:rq1-targets-game-catnip}
{
\appendixtable{%
\begin{tabular}{@{}l*{10}{c}@{}}
\toprule
\textbf{Configuration} & T1 & T2 & T3 & T4 & T5 & T6 & T7 & T8 & T9 & T10 \\
\midrule
\multicolumn{11}{@{}l}{\textbf{backbone comparison}} \\
Claude Opus~5 & S & P & P & P & S & P & S & P & P & -- \\
Qwen 3.8 Max & S & P & P & P & P & P & S & P & P & S \\
Kimi~K3 & S & P & P & P & S & P & S & S & S & S \\
Gemini~3.5 Flash & S & P & P & P & S & S & S & P & S & P \\
Gemini~3.7 Flash & P & P & -- & P & -- & P & P & P & P & P \\
GPT-5.6-SOL & P & P & -- & P & P & P & P & -- & P & P \\
DeepSeek V4 Pro & S & -- & P & P & P & P & S & -- & P & -- \\
Claude Sonnet~5 & S & P & -- & P & S & P & S & S & S & P \\
GPT-5 Mini & P & P & P & P & P & P & P & P & P & P \\
\midrule
\multicolumn{11}{@{}l}{\textbf{agent harness comparison}} \\
Claude Code + Opus~5 & S & P & P & P & S & P & S & S & S & S \\
Codex + GPT-5.6-SOL & P & P & P & P & P & P & P & P & P & P \\
\bottomrule
\end{tabular}
}
}

\end{minipage}
\end{center}
\noindent \textbf{The strongest runs combine local target preferences with the long-horizon reward and early-exit tradeoff.} Claude Code + Opus~5 and Kimi~K3 show the broadest complete-target coverage across their three opportunities, whereas other configurations often learn safe actions or reward thresholds without ever completing the preference, breed, item, or multiplier model.

\subsection{\revised{Dungeon}}
\label{app:game-dungeon}

\revised{\textbf{Learning task.} Remember rooms, keys, hazards, and treasures to refine an escape route. Fixed game, scored seed 2, first 5 episodes, reference score $C=54$.}
\begin{center}
\includegraphics[width=0.95\linewidth]{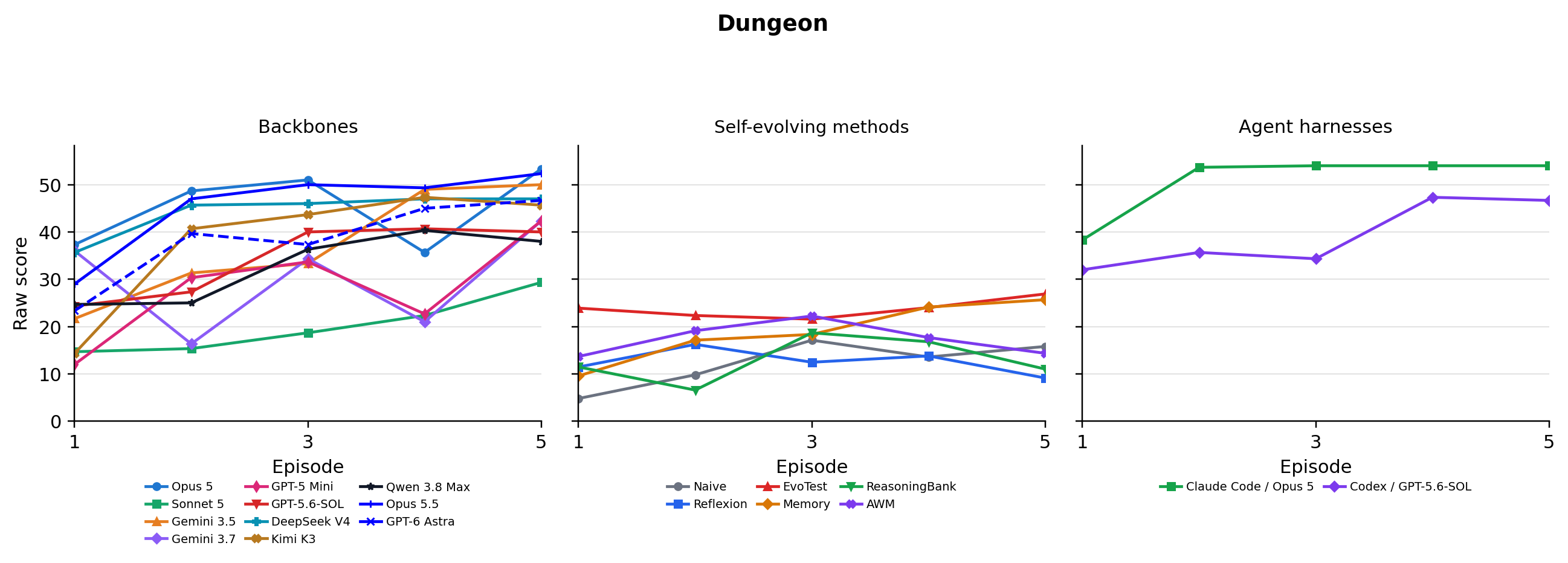}
\captionof{figure}{\revised{Dungeon: raw-score learning curves within the scoring window.}}
\label{fig:game-curve-dungeon}

\end{center}
\Needspace{10\baselineskip}
% Combined score table: normalized points and trial SD; human/First retained.
\begingroup
\setlength{\tabcolsep}{3pt}
\begin{longtable}{@{}lrrrrr@{}}
\caption{\captionrevision{Dungeon: normalized performance and learning, with trial SD. Rows follow descending Max. First denotes the average episode at which a trial first reaches its highest score.}}
\label{tab:game-scores-dungeon}\label{tab:trial-variability-dungeon}\\
\toprule
Configuration & First & Max $\uparrow$ & Mean $\uparrow$ & LG $\uparrow$ & LS $\uparrow$ \\
\midrule
\endfirsthead
\toprule
Configuration & First & Max $\uparrow$ & Mean $\uparrow$ & LG $\uparrow$ & LS $\uparrow$ \\
\midrule
\endhead
Claude Code + Opus 5 & 2.3 & $100.0 \pm 0.0$ & $94.1 \pm 6.9$ & $+14.8 \pm 17.3$ & $+5.86 \pm 6.84$ \\
Human Top-1 & 5 & 100.0 & 67.8 & -- & -- \\
OpenCode + Opus 5 & 4.7 & $98.8 \pm 2.1$ & $83.7 \pm 13.7$ & $+2.8 \pm 27.5$ & $+3.52 \pm 5.56$ \\
\opusrevision{OpenCode + Opus 5.5} & \opusrevision{3} & \opusrevision{$96.9 \pm 5.3$} & \opusrevision{$84.3 \pm 13.7$} & \opusrevision{$+23.8 \pm 16.8$} & \opusrevision{$+9.07 \pm 7.14$} \\
OpenCode + Gemini 3.5 Flash & 4.7 & $92.6 \pm 9.6$ & $68.6 \pm 4.1$ & $+42.6 \pm 19.4$ & $+13.76 \pm 5.40$ \\
EvoTest + Kimi K3 & 3 & $92.0 \pm 12.3$ & $56.4 \pm 2.8$ & $-2.5 \pm 19.0$ & $+0.19 \pm 5.47$ \\
OpenCode + Qwen 3.8 Max & 3.3 & $92.0 \pm 6.5$ & $60.9 \pm 18.2$ & $+26.5 \pm 63.3$ & $+7.78 \pm 20.37$ \\
Codex + GPT-5.6-SOL & 4 & $92.0 \pm 4.3$ & $72.6 \pm 11.9$ & $+24.4 \pm 31.5$ & $+7.59 \pm 10.08$ \\
OpenCode + Gemini 3.7 Flash & 3.7 & $89.5 \pm 5.3$ & $55.6 \pm 15.9$ & $+10.2 \pm 22.7$ & $+3.21 \pm 3.53$ \\
OpenCode + Kimi K3 & 3.3 & $87.7 \pm 10.7$ & $71.0 \pm 7.4$ & $+35.2 \pm 10.4$ & $+12.84 \pm 3.43$ \\
OpenCode + DeepSeek V4 Pro & 2 & $87.0 \pm 8.5$ & $82.0 \pm 16.2$ & $+11.7 \pm 20.3$ & $+4.44 \pm 7.70$ \\
\astrarevision{OpenCode + GPT-6 Astra} & \astrarevision{4} & \astrarevision{$87.0 \pm 9.6$} & \astrarevision{$71.1 \pm 4.0$} & \astrarevision{$+26.5 \pm 15.9$} & \astrarevision{$+9.63 \pm 5.34$} \\
OpenCode + GPT-5 Mini & 3.7 & $78.4 \pm 2.1$ & $52.2 \pm 8.7$ & $+21.0 \pm 24.6$ & $+9.81 \pm 5.85$ \\
OpenCode + GPT-5.6-SOL & 3.7 & $77.2 \pm 11.3$ & $63.8 \pm 6.3$ & $+26.9 \pm 29.1$ & $+8.27 \pm 8.52$ \\
AWM + Kimi K3 & 2.3 & $75.9 \pm 14.8$ & $48.0 \pm 10.6$ & $+9.6 \pm 36.9$ & $+0.80 \pm 14.60$ \\
Naive + GPT-5 Mini & 4.3 & $75.9 \pm 5.6$ & $31.0 \pm 9.5$ & $+18.8 \pm 14.7$ & $+5.56 \pm 5.48$ \\
Memory + Opus 5 & 4.3 & $71.0 \pm 14.0$ & $36.4 \pm 5.0$ & $+28.1 \pm 35.1$ & $+9.63 \pm 12.84$ \\
Naive + Kimi K3 & 3 & $70.4 \pm 28.7$ & $26.7 \pm 9.3$ & $+11.7 \pm 18.3$ & $+4.57 \pm 6.45$ \\
OpenCode + Sonnet 5 & 4.7 & $70.4 \pm 13.0$ & $37.2 \pm 20.7$ & $+20.1 \pm 23.2$ & $+6.73 \pm 7.71$ \\
Reflexion + Opus 5 & 4 & $66.7 \pm 9.6$ & $30.2 \pm 13.9$ & $+20.4 \pm 13.4$ & $+6.67 \pm 5.34$ \\
Memory + Kimi K3 & 3.3 & $62.3 \pm 46.6$ & $35.4 \pm 26.2$ & $+36.4 \pm 39.8$ & $+12.22 \pm 12.77$ \\
EvoTest + GPT-5 Mini & 2.7 & $58.6 \pm 39.6$ & $35.9 \pm 23.9$ & $+29.0 \pm 26.5$ & $+9.44 \pm 7.22$ \\
EvoTest + Opus 5 & 1 & $57.4 \pm 35.3$ & $39.5 \pm 21.9$ & $-13.6 \pm 37.1$ & $-5.37 \pm 10.29$ \\
Reflexion + GPT-5 Mini & 1.3 & $57.4 \pm 42.7$ & $25.2 \pm 16.5$ & $-17.6 \pm 24.1$ & $-7.04 \pm 9.63$ \\
ReasoningBank + Opus 5 & 3.7 & $51.9 \pm 13.0$ & $23.0 \pm 6.6$ & $+15.7 \pm 5.8$ & $+4.14 \pm 1.98$ \\
ReasoningBank + Kimi K3 & 1.7 & $50.0 \pm 42.1$ & $23.8 \pm 19.5$ & $-2.8 \pm 21.0$ & $-3.52 \pm 7.25$ \\
AWM + Opus 5 & 2.3 & $49.4 \pm 28.7$ & $28.9 \pm 19.8$ & $+2.5 \pm 23.2$ & $+2.90 \pm 6.58$ \\
Reflexion + Kimi K3 & 1.3 & $36.4 \pm 37.5$ & $14.6 \pm 7.8$ & $-16.0 \pm 20.1$ & $-3.58 \pm 3.74$ \\
Memory + GPT-5 Mini & 1.3 & $34.0 \pm 24.2$ & $33.5 \pm 24.8$ & $-0.3 \pm 0.5$ & $-0.06 \pm 0.11$ \\
ReasoningBank + GPT-5 Mini & 1.7 & $31.5 \pm 25.7$ & $24.8 \pm 14.8$ & $+14.2 \pm 26.2$ & $+4.57 \pm 8.23$ \\
AWM + GPT-5 Mini & 2.3 & $31.5 \pm 25.7$ & $19.8 \pm 14.2$ & $-14.2 \pm 20.4$ & $-3.76 \pm 5.76$ \\
Naive + Opus 5 & 3.3 & $29.0 \pm 27.8$ & $10.1 \pm 6.3$ & $+10.5 \pm 13.4$ & $+4.20 \pm 5.36$ \\
\bottomrule
\end{longtable}
\endgroup
\begin{longtable}{@{}p{0.07\linewidth}p{0.87\linewidth}@{}}
\caption{\captionrevision{\revised{Dungeon: criteria for assessing rule discovery.}}}\label{tab:targets-definition-dungeon}\\
\toprule
\textbf{Target} & \textbf{Criterion} \\
\midrule
\endfirsthead
\toprule
\textbf{Target} & \textbf{Criterion (continued)} \\
\midrule
\endhead
T1 & \revised{Complete 20-room graph, including every visible edge and every directed hidden passage.} \\
T2 & \revised{Room 14 is the exit and grants a ten-point escape bonus.} \\
T3 & \revised{Room 16 contains the 25-gold treasure.} \\
T4 & \revised{The key is in room 17 and is consumed when room 16 is opened.} \\
T5 & \revised{Complete location map for the charm, oil, boots, and potion.} \\
T6 & \revised{Complete effect map for those four items.} \\
T7 & \revised{Complete location and damage map for every trap and monster.} \\
T8 & \revised{Movement consumes torch while search does not. Death or torch exhaustion forfeits the escape bonus.} \\
T9 & \revised{Final score is held gold plus the escape bonus.} \\
\bottomrule
\end{longtable}
\begin{center}
\begin{minipage}{\linewidth}
\centering
\captionof{table}{\captionrevision{\revised{Dungeon: rule-discovery evidence. S: complete evidence in at least one trial; P: partial evidence only; --: insufficient evidence. Evidence may include episodes beyond the scoring window.}}}
\label{tab:rq1-targets-game-dungeon}
{
\appendixtable{%
\begin{tabular}{@{}l*{9}{c}@{}}
\toprule
\textbf{Configuration} & T1 & T2 & T3 & T4 & T5 & T6 & T7 & T8 & T9 \\
\midrule
\multicolumn{10}{@{}l}{\textbf{backbone comparison}} \\
Claude Opus~5 & P & S & S & S & S & S & S & S & S \\
Qwen 3.8 Max & P & S & S & S & P & S & P & S & S \\
Kimi~K3 & P & S & S & S & S & S & S & S & S \\
Gemini~3.5 Flash & P & S & S & S & S & S & P & S & S \\
Gemini~3.7 Flash & P & S & S & S & S & S & P & P & S \\
GPT-5.6-SOL & P & S & S & P & P & S & P & P & S \\
DeepSeek V4 Pro & P & S & S & S & S & S & P & P & S \\
Claude Sonnet~5 & P & S & S & S & S & S & P & P & S \\
GPT-5 Mini & P & S & S & P & P & P & P & P & S \\
\midrule
\multicolumn{10}{@{}l}{\textbf{agent harness comparison}} \\
Claude Code + Opus~5 & P & S & S & S & S & S & S & S & S \\
Codex + GPT-5.6-SOL & P & S & S & S & P & S & P & S & S \\
\bottomrule
\end{tabular}
}
}

\end{minipage}
\end{center}
\noindent \textbf{A successful treasure-and-exit route does not imply discovery of the complete dungeon.} Most systems recover the key landmarks and scoring rule in at least one trial, but T1 remains partial for every configuration because no trace enumerates the full room graph. Complete hazard coverage in T7 is also less common than route execution.

\subsection{\revised{SynergyCorp}}
\label{app:game-synergycorp}

\revised{\textbf{Learning task.} Learn which character traits predict preferred, weak, and taboo actions. Re-shuffled game, scored seed 2, first 5 episodes, reference score $C=137$.}
\begin{center}
\includegraphics[width=0.95\linewidth]{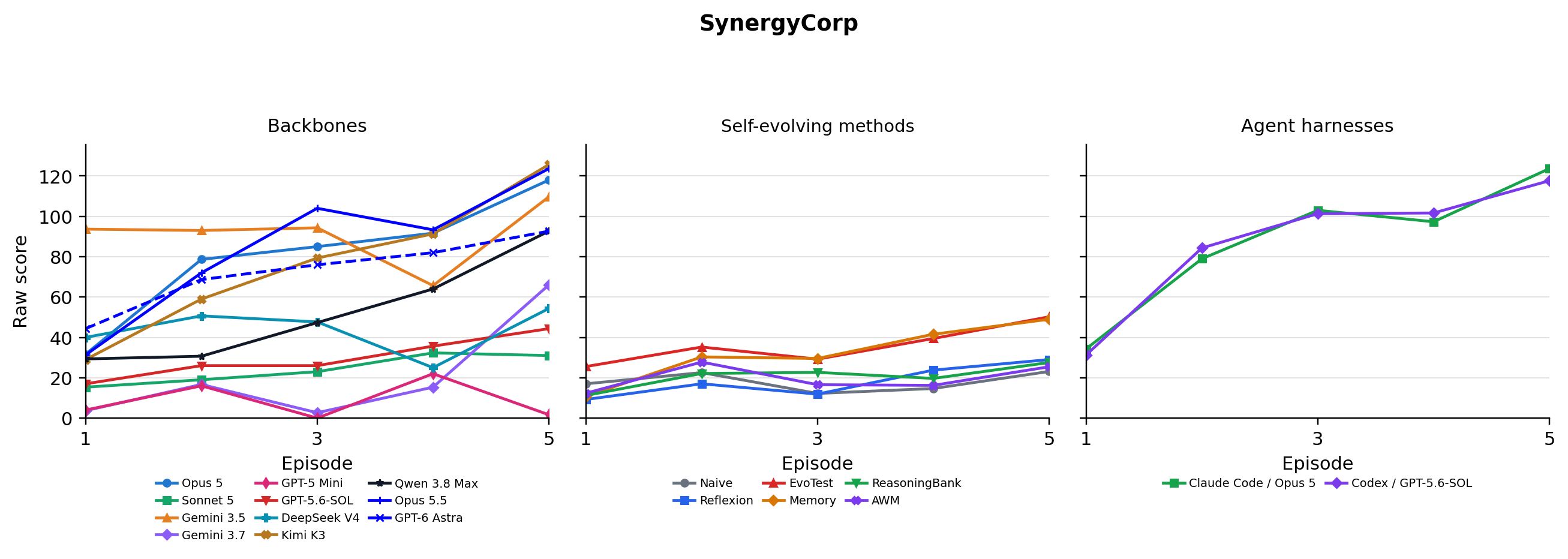}
\captionof{figure}{\revised{SynergyCorp: raw-score learning curves within the scoring window.}}
\label{fig:game-curve-synergycorp}

\end{center}
\Needspace{10\baselineskip}
% Combined score table: normalized points and trial SD; human/First retained.
\begingroup
\setlength{\tabcolsep}{3pt}
\begin{longtable}{@{}lrrrrr@{}}
\caption{\captionrevision{SynergyCorp: normalized performance and learning, with trial SD. Rows follow descending Max. First denotes the average episode at which a trial first reaches its highest score.}}
\label{tab:game-scores-synergycorp}\label{tab:trial-variability-synergycorp}\\
\toprule
Configuration & First & Max $\uparrow$ & Mean $\uparrow$ & LG $\uparrow$ & LS $\uparrow$ \\
\midrule
\endfirsthead
\toprule
Configuration & First & Max $\uparrow$ & Mean $\uparrow$ & LG $\uparrow$ & LS $\uparrow$ \\
\midrule
\endhead
OpenCode + Kimi K3 & 5 & $91.7 \pm 0.4$ & $56.1 \pm 8.4$ & $+47.1 \pm 7.1$ & $+16.47 \pm 1.46$ \\
Human Top-1 & 5 & 91.2 & 44.5 & -- & -- \\
Claude Code + Opus 5 & 5 & $90.3 \pm 2.9$ & $63.8 \pm 4.8$ & $+39.3 \pm 17.9$ & $+14.38 \pm 4.82$ \\
\opusrevision{OpenCode + Opus 5.5} & \opusrevision{5} & \opusrevision{$90.3 \pm 2.9$} & \opusrevision{$61.9 \pm 1.1$} & \opusrevision{$+41.5 \pm 5.9$} & \opusrevision{$+15.04 \pm 1.72$} \\
OpenCode + Opus 5 & 5 & $86.1 \pm 7.6$ & $59.1 \pm 1.0$ & $+36.3 \pm 9.4$ & $+13.55 \pm 2.96$ \\
Codex + GPT-5.6-SOL & 5 & $85.9 \pm 6.6$ & $63.7 \pm 4.2$ & $+37.8 \pm 5.9$ & $+13.87 \pm 0.91$ \\
OpenCode + Gemini 3.5 Flash & 2.3 & $81.8 \pm 17.7$ & $66.6 \pm 18.1$ & $-4.1 \pm 23.2$ & $+0.34 \pm 7.00$ \\
Memory + Kimi K3 & 4 & $70.3 \pm 5.9$ & $43.1 \pm 6.5$ & $+46.2 \pm 3.1$ & $+14.50 \pm 1.32$ \\
OpenCode + Qwen 3.8 Max & 5 & $67.6 \pm 26.6$ & $38.5 \pm 16.1$ & $+35.3 \pm 24.0$ & $+11.68 \pm 7.35$ \\
\astrarevision{OpenCode + GPT-6 Astra} & \astrarevision{5} & \astrarevision{$67.6 \pm 40.9$} & \astrarevision{$53.1 \pm 34.7$} & \astrarevision{$+22.5 \pm 15.7$} & \astrarevision{$+8.03 \pm 5.28$} \\
EvoTest + Kimi K3 & 4 & $61.6 \pm 10.0$ & $40.9 \pm 9.1$ & $+22.7 \pm 6.9$ & $+8.10 \pm 1.84$ \\
OpenCode + DeepSeek V4 Pro & 3.3 & $50.4 \pm 20.3$ & $31.8 \pm 21.3$ & $-4.1 \pm 24.3$ & $+0.22 \pm 6.84$ \\
OpenCode + Gemini 3.7 Flash & 5 & $48.2 \pm 33.6$ & $15.2 \pm 4.9$ & $+22.3 \pm 11.6$ & $+9.00 \pm 5.93$ \\
EvoTest + Opus 5 & 5 & $42.6 \pm 1.7$ & $28.9 \pm 3.3$ & $+6.8 \pm 1.4$ & $+3.14 \pm 1.42$ \\
ReasoningBank + Kimi K3 & 3 & $35.5 \pm 12.4$ & $23.4 \pm 7.0$ & $+7.7 \pm 7.4$ & $+2.97 \pm 0.91$ \\
AWM + Opus 5 & 5 & $34.5 \pm 2.8$ & $19.6 \pm 2.5$ & $+9.5 \pm 4.3$ & $+4.33 \pm 0.75$ \\
ReasoningBank + Opus 5 & 4 & $33.6 \pm 3.2$ & $13.9 \pm 2.0$ & $+11.7 \pm 10.7$ & $+5.04 \pm 1.80$ \\
Reflexion + Kimi K3 & 4.7 & $33.1 \pm 4.6$ & $14.6 \pm 3.3$ & $+7.2 \pm 9.1$ & $+2.26 \pm 2.89$ \\
OpenCode + GPT-5.6-SOL & 4.3 & $32.8 \pm 10.9$ & $21.8 \pm 4.2$ & $+13.5 \pm 9.5$ & $+4.70 \pm 2.95$ \\
AWM + Kimi K3 & 3.7 & $29.4 \pm 8.2$ & $16.0 \pm 4.3$ & $-0.6 \pm 13.4$ & $+0.73 \pm 4.97$ \\
Naive + Kimi K3 & 3.7 & $28.5 \pm 3.2$ & $19.4 \pm 3.1$ & $-0.2 \pm 7.1$ & $+0.78 \pm 2.17$ \\
Reflexion + Opus 5 & 4 & $28.5 \pm 1.9$ & $14.8 \pm 1.9$ & $+14.1 \pm 2.8$ & $+4.62 \pm 1.43$ \\
Memory + Opus 5 & 5 & $27.7 \pm 7.0$ & $13.9 \pm 5.0$ & $+1.2 \pm 7.5$ & $+2.21 \pm 1.43$ \\
OpenCode + Sonnet 5 & 4 & $27.3 \pm 4.4$ & $17.6 \pm 3.3$ & $+10.6 \pm 2.4$ & $+3.26 \pm 0.92$ \\
Memory + GPT-5 Mini & 3.7 & $26.5 \pm 9.3$ & $13.6 \pm 5.2$ & $+6.7 \pm 8.6$ & $+2.48 \pm 2.41$ \\
Naive + Opus 5 & 2.7 & $23.6 \pm 5.7$ & $15.2 \pm 3.0$ & $-1.3 \pm 0.9$ & $+0.17 \pm 1.72$ \\
Reflexion + GPT-5 Mini & 3.7 & $23.6 \pm 3.4$ & $10.4 \pm 2.6$ & $+7.8 \pm 10.2$ & $+3.26 \pm 3.03$ \\
AWM + GPT-5 Mini & 2.3 & $20.4 \pm 7.4$ & $7.4 \pm 2.6$ & $-7.2 \pm 4.6$ & $-1.85 \pm 1.29$ \\
ReasoningBank + GPT-5 Mini & 2.7 & $20.0 \pm 9.0$ & $7.9 \pm 3.6$ & $-4.1 \pm 4.1$ & $-1.39 \pm 1.74$ \\
EvoTest + GPT-5 Mini & 4.3 & $16.3 \pm 4.0$ & $8.9 \pm 0.3$ & $+2.1 \pm 1.6$ & $+0.49 \pm 0.84$ \\
OpenCode + GPT-5 Mini & 4 & $16.1 \pm 2.5$ & $6.4 \pm 0.4$ & $+1.3 \pm 3.8$ & $+0.10 \pm 1.33$ \\
Naive + GPT-5 Mini & 2.7 & $12.4 \pm 0.0$ & $4.6 \pm 1.1$ & $-0.4 \pm 4.6$ & $-0.00 \pm 0.96$ \\
\bottomrule
\end{longtable}
\endgroup
\begin{longtable}{@{}p{0.07\linewidth}p{0.87\linewidth}@{}}
\caption{\captionrevision{\revised{SynergyCorp: criteria for assessing rule discovery.}}}\label{tab:targets-definition-synergycorp}\\
\toprule
\textbf{Target} & \textbf{Criterion} \\
\midrule
\endfirsthead
\toprule
\textbf{Target} & \textbf{Criterion (continued)} \\
\midrule
\endhead
T1 & \revised{Complete partition of the eight actions into four behavior pairs. Designer-assigned modality names are not required.} \\
T2 & \revised{Complete mapping of eight informative desk traits and identification of four decoys.} \\
T3 & \revised{Complete preference ordering over the four behavior groups for each boss type.} \\
T4 & \revised{The overtime-selfie and buzzwords taboo clears approval and adds eight demerits.} \\
T5 & \revised{Repeating one action produces two-stage decay, motivating alternation within a pair.} \\
T6 & \revised{Steal-lunch followed by buzzwords applies the ordered combo bonus.} \\
T7 & \revised{A drifter changes preference at half its promotion threshold, with the second signal identifying the new preference.} \\
T8 & \revised{Promotion thresholds, escalating rewards, and the 30-step budget jointly determine total score.} \\
\bottomrule
\end{longtable}
\begin{center}
\begin{minipage}{\linewidth}
\centering
\captionof{table}{\captionrevision{\revised{SynergyCorp: rule-discovery evidence. S: complete evidence in at least one trial; P: partial evidence only; --: insufficient evidence. Evidence may include episodes beyond the scoring window.}}}
\label{tab:rq1-targets-game-synergycorp}
{
\appendixtable{%
\begin{tabular}{@{}l*{8}{c}@{}}
\toprule
\textbf{Configuration} & T1 & T2 & T3 & T4 & T5 & T6 & T7 & T8 \\
\midrule
\multicolumn{9}{@{}l}{\textbf{backbone comparison}} \\
Claude Opus~5 & S & P & S & S & S & -- & P & P \\
Qwen 3.8 Max & P & P & P & P & P & -- & S & P \\
Kimi~K3 & S & P & P & P & S & -- & P & S \\
Gemini~3.5 Flash & S & P & P & -- & P & -- & P & -- \\
Gemini~3.7 Flash & S & P & -- & P & P & -- & -- & P \\
GPT-5.6-SOL & P & P & P & P & P & P & P & P \\
DeepSeek V4 Pro & P & P & P & -- & -- & -- & P & -- \\
Claude Sonnet~5 & P & P & P & P & P & P & P & P \\
GPT-5 Mini & P & P & P & P & P & P & P & P \\
\midrule
\multicolumn{9}{@{}l}{\textbf{agent harness comparison}} \\
Claude Code + Opus~5 & S & S & P & P & S & P & P & P \\
Codex + GPT-5.6-SOL & P & P & P & -- & P & P & P & -- \\
\bottomrule
\end{tabular}
}
}

\end{minipage}
\end{center}
\noindent \textbf{Systems learn the action groups more readily than the temporal rules needed for a complete promotion policy.} Several configurations fully recover the action partition in at least one trial, but exact taboo, combo, repetition-decay, and drifter-switch conditions rarely become complete within any single trial.

\subsection{\revised{DreamGarden}}
\label{app:game-dreamgarden}

\revised{\textbf{Learning task.} Infer delayed ecological effects and plan actions across turns. Fixed game, scored seed 1, first 10 episodes, reference score $C=6300$.}
\begin{center}
\includegraphics[width=0.95\linewidth]{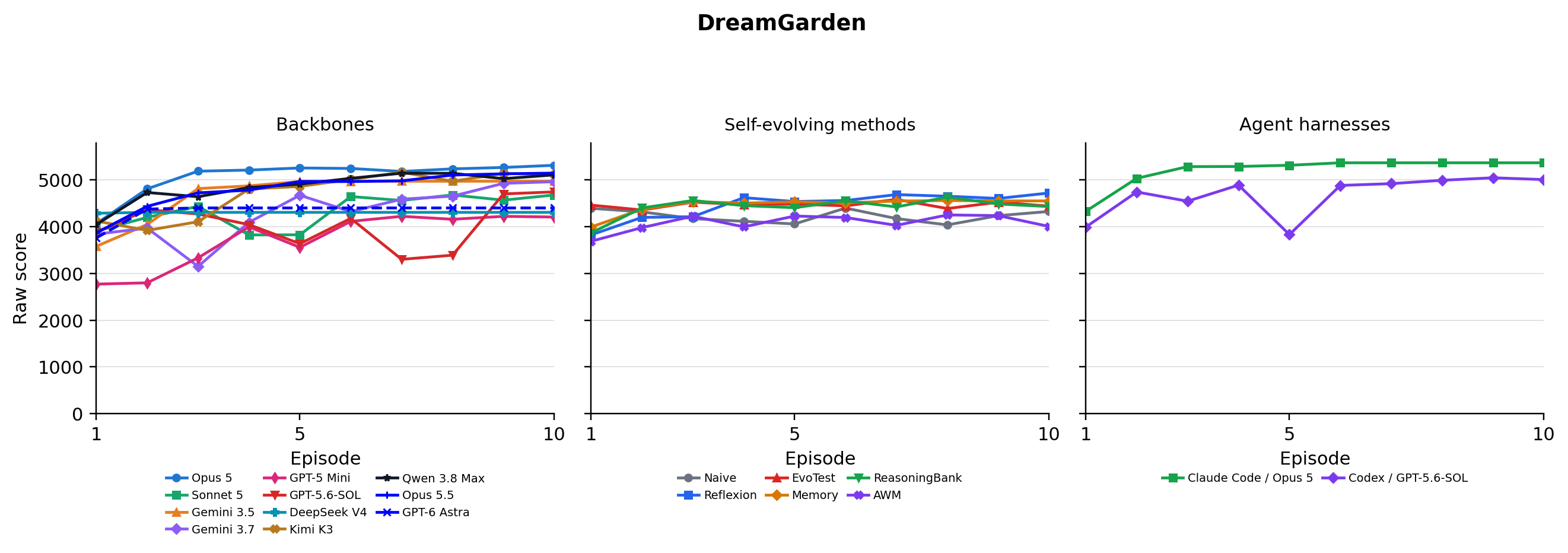}
\captionof{figure}{\revised{DreamGarden: raw-score learning curves within the scoring window.}}
\label{fig:game-curve-dreamgarden}

\end{center}
\Needspace{10\baselineskip}
% Combined score table: normalized points and trial SD; human/First retained.
\begingroup
\setlength{\tabcolsep}{3pt}
\begin{longtable}{@{}lrrrrr@{}}
\caption{\captionrevision{DreamGarden: normalized performance and learning, with trial SD. Rows follow descending Max. First denotes the average episode at which a trial first reaches its highest score.}}
\label{tab:game-scores-dreamgarden}\label{tab:trial-variability-dreamgarden}\\
\toprule
Configuration & First & Max $\uparrow$ & Mean $\uparrow$ & LG $\uparrow$ & LS $\uparrow$ \\
\midrule
\endfirsthead
\toprule
Configuration & First & Max $\uparrow$ & Mean $\uparrow$ & LG $\uparrow$ & LS $\uparrow$ \\
\midrule
\endhead
Claude Code + Opus 5 & 4.7 & $85.2 \pm 0.0$ & $82.7 \pm 1.5$ & $+7.7 \pm 4.0$ & $+1.19 \pm 0.55$ \\
OpenCode + Opus 5 & 7.7 & $84.3 \pm 0.8$ & $80.6 \pm 2.4$ & $+9.3 \pm 2.7$ & $+1.40 \pm 0.35$ \\
Human Top-1 & 8 & 84.2 & 78.7 & -- & -- \\
Reflexion + Opus 5 & 7.3 & $82.8 \pm 2.5$ & $74.4 \pm 1.0$ & $+6.4 \pm 1.7$ & $+0.97 \pm 0.31$ \\
OpenCode + Kimi K3 & 8.7 & $82.6 \pm 1.3$ & $75.0 \pm 3.6$ & $+16.3 \pm 6.8$ & $+2.22 \pm 0.92$ \\
OpenCode + Qwen 3.8 Max & 5.7 & $81.8 \pm 2.5$ & $77.2 \pm 1.4$ & $+9.8 \pm 4.6$ & $+1.46 \pm 0.51$ \\
\opusrevision{OpenCode + Opus 5.5} & \opusrevision{8.7} & \opusrevision{$81.7 \pm 2.1$} & \opusrevision{$76.4 \pm 2.9$} & \opusrevision{$+12.5 \pm 3.1$} & \opusrevision{$+1.82 \pm 0.29$} \\
ReasoningBank + GPT-5 Mini & 5.7 & $81.4 \pm 0.8$ & $72.6 \pm 2.4$ & $+1.9 \pm 7.2$ & $+0.32 \pm 0.98$ \\
Codex + GPT-5.6-SOL & 7 & $80.2 \pm 1.9$ & $74.4 \pm 4.4$ & $+9.3 \pm 3.5$ & $+1.41 \pm 0.52$ \\
Reflexion + Kimi K3 & 7.3 & $79.1 \pm 4.9$ & $70.5 \pm 4.3$ & $+10.3 \pm 10.6$ & $+1.31 \pm 1.12$ \\
OpenCode + Gemini 3.5 Flash & 4.3 & $78.9 \pm 5.8$ & $74.8 \pm 2.8$ & $+13.1 \pm 10.2$ & $+1.93 \pm 1.47$ \\
ReasoningBank + Opus 5 & 7.3 & $78.9 \pm 2.2$ & $71.5 \pm 2.2$ & $+6.3 \pm 2.0$ & $+1.02 \pm 0.30$ \\
OpenCode + Gemini 3.7 Flash & 7 & $78.8 \pm 0.2$ & $68.5 \pm 8.3$ & $+19.0 \pm 9.3$ & $+2.45 \pm 1.06$ \\
EvoTest + Opus 5 & 3 & $76.9 \pm 6.1$ & $72.6 \pm 4.0$ & $-0.5 \pm 0.7$ & $+0.03 \pm 0.22$ \\
ReasoningBank + Kimi K3 & 7 & $76.4 \pm 5.2$ & $66.4 \pm 1.7$ & $+3.2 \pm 3.3$ & $+0.43 \pm 0.60$ \\
Naive + GPT-5 Mini & 7.3 & $75.9 \pm 7.0$ & $68.4 \pm 1.7$ & $+1.4 \pm 3.1$ & $+0.34 \pm 0.38$ \\
Naive + Opus 5 & 4.3 & $75.6 \pm 0.4$ & $70.4 \pm 0.7$ & $-0.3 \pm 3.4$ & $+0.00 \pm 0.40$ \\
OpenCode + GPT-5.6-SOL & 6 & $75.3 \pm 7.4$ & $64.3 \pm 14.4$ & $+1.7 \pm 9.5$ & $+0.39 \pm 1.49$ \\
Memory + Kimi K3 & 4.7 & $75.2 \pm 2.0$ & $69.3 \pm 1.0$ & $+7.2 \pm 9.3$ & $+1.19 \pm 1.30$ \\
EvoTest + GPT-5 Mini & 1.7 & $74.9 \pm 7.2$ & $71.1 \pm 7.1$ & $+0.8 \pm 5.7$ & $+0.15 \pm 0.83$ \\
Memory + Opus 5 & 4 & $74.8 \pm 6.8$ & $72.4 \pm 5.1$ & $+3.3 \pm 1.6$ & $+0.44 \pm 0.36$ \\
OpenCode + Sonnet 5 & 5.7 & $74.3 \pm 4.7$ & $68.8 \pm 0.7$ & $+7.3 \pm 4.9$ & $+1.33 \pm 0.95$ \\
Naive + Kimi K3 & 3.7 & $73.4 \pm 0.8$ & $62.3 \pm 1.9$ & $-5.6 \pm 3.2$ & $-0.73 \pm 0.94$ \\
AWM + Opus 5 & 3.7 & $73.0 \pm 3.6$ & $67.2 \pm 4.4$ & $+1.9 \pm 5.3$ & $+0.39 \pm 0.86$ \\
AWM + GPT-5 Mini & 5.3 & $72.8 \pm 0.8$ & $67.5 \pm 1.5$ & $-0.7 \pm 3.1$ & $-0.12 \pm 0.46$ \\
EvoTest + Kimi K3 & 4 & $72.4 \pm 0.6$ & $69.1 \pm 0.2$ & $-0.2 \pm 1.4$ & $-0.01 \pm 0.17$ \\
Reflexion + GPT-5 Mini & 6.3 & $72.1 \pm 8.0$ & $67.6 \pm 7.2$ & $+10.9 \pm 11.4$ & $+1.57 \pm 1.63$ \\
Memory + GPT-5 Mini & 2 & $71.7 \pm 4.2$ & $71.1 \pm 4.5$ & $+1.5 \pm 1.3$ & $+0.25 \pm 0.22$ \\
\astrarevision{OpenCode + GPT-6 Astra} & \astrarevision{1.7} & \astrarevision{$69.9 \pm 5.9$} & \astrarevision{$68.8 \pm 4.7$} & \astrarevision{$+3.5 \pm 4.6$} & \astrarevision{$+0.57 \pm 0.76$} \\
OpenCode + DeepSeek V4 Pro & 1.7 & $68.4 \pm 0.5$ & $68.3 \pm 0.6$ & $+0.1 \pm 0.1$ & $+0.02 \pm 0.01$ \\
OpenCode + GPT-5 Mini & 5.7 & $68.0 \pm 10.3$ & $59.3 \pm 13.8$ & $+19.5 \pm 10.5$ & $+2.72 \pm 1.28$ \\
AWM + Kimi K3 & 6.7 & $66.8 \pm 2.8$ & $59.7 \pm 2.0$ & $+8.3 \pm 4.6$ & $+1.13 \pm 1.04$ \\
\bottomrule
\end{longtable}
\endgroup
\begin{longtable}{@{}p{0.07\linewidth}p{0.87\linewidth}@{}}
\caption{\captionrevision{\revised{DreamGarden: criteria for assessing rule discovery.}}}\label{tab:targets-definition-dreamgarden}\\
\toprule
\textbf{Target} & \textbf{Criterion} \\
\midrule
\endfirsthead
\toprule
\textbf{Target} & \textbf{Criterion (continued)} \\
\midrule
\endhead
T1 & \revised{Resonance per turn is \(\sum v_j\min(pop_j,35)\). Population is capped at 100 and decays by 20\%. For seed 1, Spores and Moths are worth \(+5\), Nightmares \(-3\), and Jelly and Weavers 0. Nightmares suppress the Weaver chain.} \\
T2 & \revised{Eclipse increases Moths but also introduces Nightmares. Throb increases Weavers, Whisper suppresses Nightmares, and Tide feeds only Nightmares.} \\
T3 & \revised{Throb starts the chain Weavers\(\to\)Jelly\(\to\)Spores. Each ecological link takes one turn, so the final benefit is delayed.} \\
T4 & \revised{Over 18 turns, combine direct Moth growth, delayed Spore growth, and Nightmare removal. Throb has little immediate value but a large payoff two turns later.} \\
T5 & \revised{The archived 5,367-point sequence starts with Throb, then interleaves Eclipse and Whisper to keep Moths and Spores near saturation and Nightmares near zero. It never uses Tide.} \\
\bottomrule
\end{longtable}
\begin{center}
\begin{minipage}{\linewidth}
\centering
\captionof{table}{\captionrevision{\revised{DreamGarden: rule-discovery evidence. S: complete evidence in at least one trial; P: partial evidence only; --: insufficient evidence. Evidence may include episodes beyond the scoring window.}}}
\label{tab:rq1-targets-game-dreamgarden}
{
\appendixtable{%
\begin{tabular}{@{}l*{5}{c}@{}}
\toprule
\textbf{Configuration} & T1 & T2 & T3 & T4 & T5 \\
\midrule
\multicolumn{6}{@{}l}{\textbf{backbone comparison}} \\
Claude Opus~5 & S & S & S & S & S \\
Qwen 3.8 Max & S & S & S & S & S \\
Kimi~K3 & S & S & S & S & S \\
Gemini~3.5 Flash & S & S & S & S & S \\
Gemini~3.7 Flash & S & S & S & S & -- \\
GPT-5.6-SOL & S & -- & -- & S & -- \\
DeepSeek V4 Pro & S & S & S & S & -- \\
Claude Sonnet~5 & S & S & S & S & -- \\
GPT-5 Mini & S & S & -- & S & -- \\
\midrule
\multicolumn{6}{@{}l}{\textbf{agent harness comparison}} \\
Claude Code + Opus~5 & S & S & S & S & S \\
Codex + GPT-5.6-SOL & S & S & S & S & -- \\
\bottomrule
\end{tabular}
}
}

\end{minipage}
\end{center}

\subsection{\revised{Ecosphere}}
\label{app:game-ecosphere}

\revised{\textbf{Learning task.} Infer food-web, competition, and waste dynamics to plan stocking. Fixed game, scored seed 1, first 5 episodes, reference score $C=339$.}
\begin{center}
\includegraphics[width=0.95\linewidth]{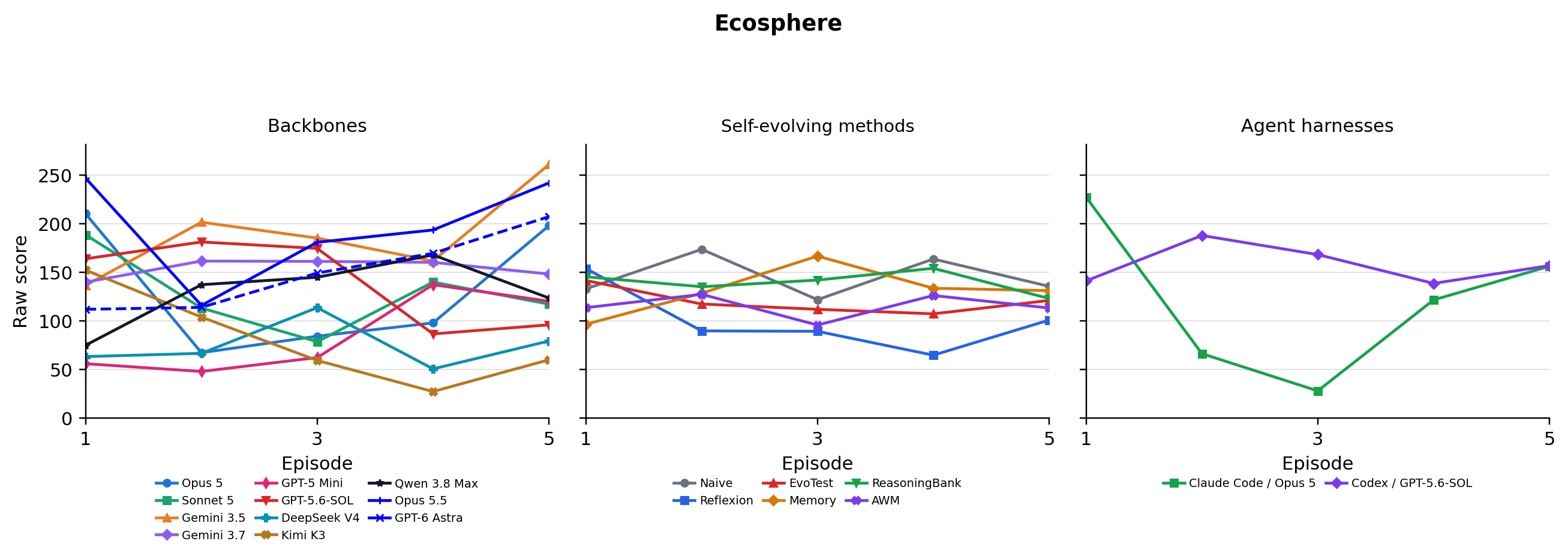}
\captionof{figure}{\revised{Ecosphere: raw-score learning curves within the scoring window.}}
\label{fig:game-curve-ecosphere}

\end{center}
\Needspace{10\baselineskip}
% Combined score table: normalized points and trial SD; human/First retained.
\begingroup
\setlength{\tabcolsep}{3pt}
\begin{longtable}{@{}lrrrrr@{}}
\caption{\captionrevision{Ecosphere: normalized performance and learning, with trial SD. Rows follow descending Max. First denotes the average episode at which a trial first reaches its highest score.}}
\label{tab:game-scores-ecosphere}\label{tab:trial-variability-ecosphere}\\
\toprule
Configuration & First & Max $\uparrow$ & Mean $\uparrow$ & LG $\uparrow$ & LS $\uparrow$ \\
\midrule
\endfirsthead
\toprule
Configuration & First & Max $\uparrow$ & Mean $\uparrow$ & LG $\uparrow$ & LS $\uparrow$ \\
\midrule
\endhead
Human Top-1 & 5 & 85.0 & 59.7 & -- & -- \\
Claude Code + Opus 5 & 2.3 & $80.1 \pm 19.6$ & $35.4 \pm 18.2$ & $-2.3 \pm 23.0$ & $-2.56 \pm 9.55$ \\
\opusrevision{OpenCode + Opus 5.5} & \opusrevision{2.7} & \opusrevision{$79.9 \pm 5.7$} & \opusrevision{$57.8 \pm 5.4$} & \opusrevision{$+10.7 \pm 3.7$} & \opusrevision{$+2.00 \pm 1.81$} \\
OpenCode + Gemini 3.5 Flash & 4 & $77.1 \pm 15.3$ & $55.8 \pm 12.2$ & $+12.4 \pm 21.4$ & $+6.16 \pm 4.96$ \\
Reflexion + Opus 5 & 4 & $76.3 \pm 21.7$ & $46.5 \pm 13.2$ & $-10.0 \pm 15.6$ & $-3.97 \pm 7.40$ \\
Naive + Opus 5 & 3 & $75.1 \pm 6.6$ & $56.7 \pm 8.1$ & $-4.2 \pm 21.8$ & $-2.35 \pm 8.88$ \\
Memory + Kimi K3 & 3.3 & $73.0 \pm 23.7$ & $31.4 \pm 16.0$ & $+11.5 \pm 22.4$ & $+5.01 \pm 5.27$ \\
Naive + Kimi K3 & 3.3 & $69.4 \pm 10.9$ & $42.2 \pm 8.8$ & $+1.3 \pm 30.8$ & $+2.16 \pm 6.68$ \\
\astrarevision{OpenCode + GPT-6 Astra} & \astrarevision{4} & \astrarevision{$69.4 \pm 18.8$} & \astrarevision{$44.4 \pm 11.2$} & \astrarevision{$+22.3 \pm 20.5$} & \astrarevision{$+7.27 \pm 6.33$} \\
OpenCode + Opus 5 & 3 & $68.5 \pm 16.1$ & $38.8 \pm 5.4$ & $+2.7 \pm 17.7$ & $+0.18 \pm 5.61$ \\
OpenCode + Gemini 3.7 Flash & 2.3 & $66.7 \pm 6.2$ & $45.5 \pm 13.3$ & $+1.0 \pm 15.7$ & $+0.45 \pm 3.91$ \\
AWM + Kimi K3 & 2.7 & $65.5 \pm 16.6$ & $27.7 \pm 12.2$ & $-18.8 \pm 54.2$ & $-5.75 \pm 18.52$ \\
EvoTest + Opus 5 & 1.7 & $64.1 \pm 19.8$ & $50.2 \pm 19.5$ & $-7.0 \pm 7.6$ & $-2.59 \pm 2.30$ \\
AWM + Opus 5 & 4 & $63.8 \pm 12.0$ & $48.2 \pm 5.5$ & $+24.6 \pm 6.8$ & $+8.53 \pm 1.76$ \\
Memory + Opus 5 & 1.3 & $63.3 \pm 7.6$ & $59.7 \pm 8.2$ & $+4.7 \pm 2.8$ & $+1.07 \pm 0.38$ \\
OpenCode + Sonnet 5 & 2 & $59.0 \pm 5.4$ & $37.6 \pm 8.9$ & $-6.6 \pm 14.0$ & $-3.42 \pm 2.74$ \\
Codex + GPT-5.6-SOL & 3 & $57.8 \pm 4.1$ & $46.8 \pm 6.3$ & $-5.0 \pm 10.3$ & $-0.53 \pm 3.76$ \\
ReasoningBank + Kimi K3 & 1.3 & $55.7 \pm 6.0$ & $51.3 \pm 12.2$ & $-7.3 \pm 12.7$ & $-2.90 \pm 5.02$ \\
OpenCode + GPT-5.6-SOL & 1.7 & $53.5 \pm 15.8$ & $41.5 \pm 11.5$ & $-24.0 \pm 17.3$ & $-6.80 \pm 5.46$ \\
OpenCode + Qwen 3.8 Max & 3.7 & $52.8 \pm 8.3$ & $38.3 \pm 1.0$ & $+11.7 \pm 4.5$ & $+3.78 \pm 0.92$ \\
OpenCode + GPT-5 Mini & 2.7 & $46.1 \pm 19.1$ & $25.0 \pm 12.6$ & $+22.7 \pm 23.1$ & $+6.43 \pm 6.64$ \\
OpenCode + Kimi K3 & 1.3 & $45.3 \pm 14.2$ & $23.8 \pm 10.6$ & $-24.9 \pm 25.7$ & $-7.71 \pm 6.14$ \\
Reflexion + Kimi K3 & 2.3 & $44.8 \pm 8.7$ & $12.8 \pm 4.8$ & $-15.5 \pm 28.3$ & $-5.27 \pm 10.49$ \\
Reflexion + GPT-5 Mini & 1.3 & $43.6 \pm 7.1$ & $28.9 \pm 6.1$ & $-9.0 \pm 9.9$ & $-2.37 \pm 2.56$ \\
ReasoningBank + GPT-5 Mini & 2 & $43.3 \pm 16.8$ & $31.4 \pm 8.8$ & $+4.0 \pm 10.9$ & $-0.10 \pm 2.10$ \\
ReasoningBank + Opus 5 & 2 & $41.9 \pm 14.4$ & $41.2 \pm 14.1$ & $+1.8 \pm 1.3$ & $+0.73 \pm 0.53$ \\
AWM + GPT-5 Mini & 1 & $36.6 \pm 7.4$ & $26.2 \pm 6.3$ & $-6.5 \pm 2.6$ & $-2.94 \pm 0.62$ \\
EvoTest + Kimi K3 & 1 & $36.3 \pm 12.6$ & $36.3 \pm 12.6$ & $+0.0 \pm 0.0$ & $+0.00 \pm 0.00$ \\
Naive + GPT-5 Mini & 2.7 & $35.9 \pm 7.0$ & $30.0 \pm 2.9$ & $-0.3 \pm 2.5$ & $-0.23 \pm 1.16$ \\
OpenCode + DeepSeek V4 Pro & 3.3 & $34.5 \pm 27.3$ & $22.0 \pm 18.6$ & $+0.0 \pm 19.2$ & $+0.47 \pm 5.68$ \\
EvoTest + GPT-5 Mini & 1.7 & $25.8 \pm 4.1$ & $19.6 \pm 2.7$ & $-6.6 \pm 8.8$ & $-1.95 \pm 2.10$ \\
Memory + GPT-5 Mini & 2 & $25.7 \pm 12.3$ & $25.1 \pm 11.6$ & $+1.4 \pm 1.9$ & $+0.50 \pm 0.67$ \\
\bottomrule
\end{longtable}
\endgroup
\begin{longtable}{@{}p{0.07\linewidth}p{0.87\linewidth}@{}}
\caption{\captionrevision{\revised{Ecosphere: criteria for assessing rule discovery.}}}\label{tab:targets-definition-ecosphere}\\
\toprule
\textbf{Target} & \textbf{Criterion} \\
\midrule
\endfirsthead
\toprule
\textbf{Target} & \textbf{Criterion (continued)} \\
\midrule
\endhead
T1 & \revised{Spend a one-time budget of 150 on stocking, using at most 40 setup actions. Sealing the habitat starts a deterministic 30-day simulation.} \\
T2 & \revised{Moss, algae, and microbes support the ecosystem but score zero. Herbivores score 4 per animal, subject to saturation. Minnows and newts score 20.} \\
T3 & \revised{Seed-1 producer layer: moss \(24\times\) and algae \(18\times\) biomass per unit, each capped at 10 stocked units.} \\
T4 & \revised{Seed-1 food web and competition: snail and shrimp eat moss, daphnia eats algae, rank order gives shrimp priority over snail, and overgrazed producers collapse.} \\
T5 & \revised{Predator and keystone dynamics: minnow hunts shrimp only, newt hunts shrimp/snail, prey below refuge 5 cannot be caught, and a single newt culls shrimp enough for snails to survive.} \\
T6 & \revised{Living animals and corpses create waste that produces toxic gas. Enough decomposers must be stocked to clear it.} \\
T7 & \revised{Unique 339 stocking vector: moss 10, algae 9, snail 2, shrimp 2, daphnia 2, newt 1, microbe 8, no minnow. This ends at snail 12, shrimp 18, daphnia 27, newt 7 with no extinction event.} \\
\bottomrule
\end{longtable}
\begin{center}
\begin{minipage}{\linewidth}
\centering
\captionof{table}{\captionrevision{\revised{Ecosphere: rule-discovery evidence. S: complete evidence in at least one trial; P: partial evidence only; --: insufficient evidence. Evidence may include episodes beyond the scoring window.}}}
\label{tab:rq1-targets-game-ecosphere}
{
\appendixtable{%
\begin{tabular}{@{}l*{7}{c}@{}}
\toprule
\textbf{Configuration} & T1 & T2 & T3 & T4 & T5 & T6 & T7 \\
\midrule
\multicolumn{8}{@{}l}{\textbf{backbone comparison}} \\
Claude Opus~5 & S & S & S & S & S & S & -- \\
Qwen 3.8 Max & S & S & S & S & S & S & -- \\
Kimi~K3 & S & S & S & S & S & S & -- \\
Gemini~3.5 Flash & S & S & S & S & S & S & S \\
Gemini~3.7 Flash & S & S & S & S & S & S & -- \\
GPT-5.6-SOL & -- & S & S & S & S & S & -- \\
DeepSeek V4 Pro & -- & S & S & S & S & S & -- \\
Claude Sonnet~5 & S & S & S & S & S & S & -- \\
GPT-5 Mini & S & S & S & S & S & S & -- \\
\midrule
\multicolumn{8}{@{}l}{\textbf{agent harness comparison}} \\
Claude Code + Opus~5 & S & S & S & S & S & S & S \\
Codex + GPT-5.6-SOL & S & S & S & S & S & S & -- \\
\bottomrule
\end{tabular}
}
}

\end{minipage}
\end{center}

\subsection{\revised{Grapevine}}
\label{app:game-grapevine}

\revised{\textbf{Learning task.} Infer a cascade network and choose effective seed combinations. Fixed game, scored seed 1, first 5 episodes, reference score $C=16$.}
\begin{center}
\includegraphics[width=0.95\linewidth]{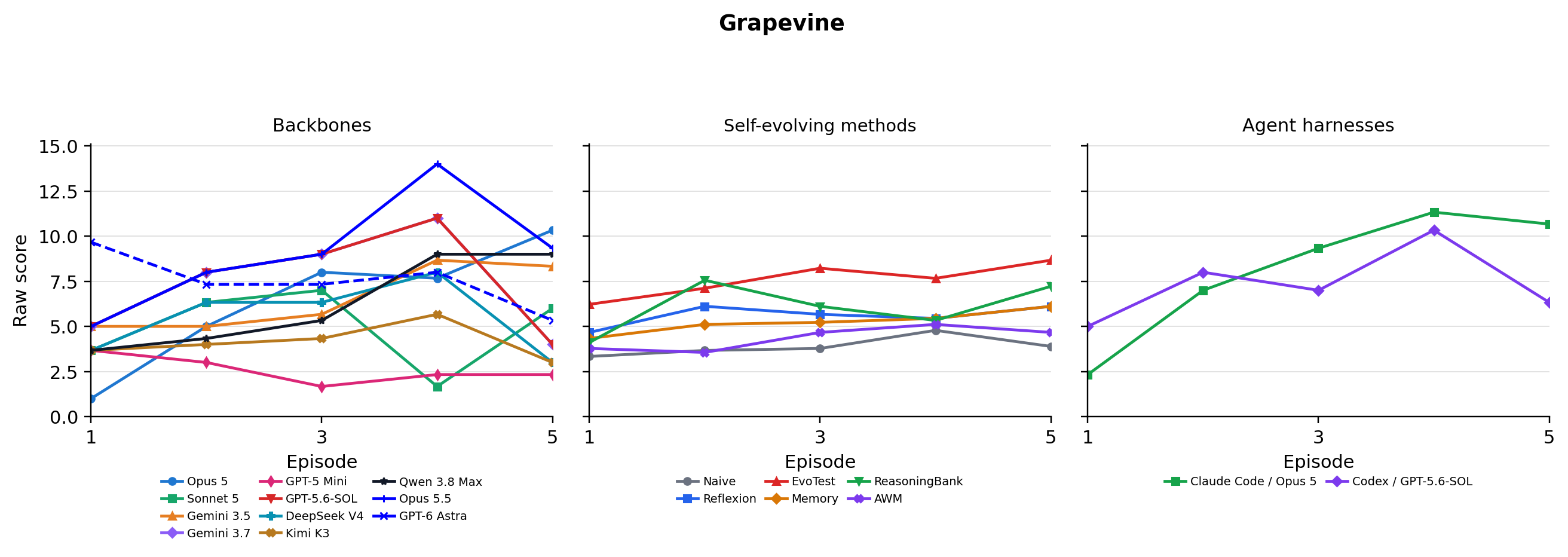}
\captionof{figure}{\revised{Grapevine: raw-score learning curves within the scoring window.}}
\label{fig:game-curve-grapevine}

\end{center}
\Needspace{10\baselineskip}
% Combined score table: normalized points and trial SD; human/First retained.
\begingroup
\setlength{\tabcolsep}{3pt}
\begin{longtable}{@{}lrrrrr@{}}
\caption{\captionrevision{Grapevine: normalized performance and learning, with trial SD. Rows follow descending Max. First denotes the average episode at which a trial first reaches its highest score.}}
\label{tab:game-scores-grapevine}\label{tab:trial-variability-grapevine}\\
\toprule
Configuration & First & Max $\uparrow$ & Mean $\uparrow$ & LG $\uparrow$ & LS $\uparrow$ \\
\midrule
\endfirsthead
\toprule
Configuration & First & Max $\uparrow$ & Mean $\uparrow$ & LG $\uparrow$ & LS $\uparrow$ \\
\midrule
\endhead
Human Top-1 & 5 & 93.8 & 52.5 & -- & -- \\
\opusrevision{OpenCode + Opus 5.5} & \opusrevision{4.3} & \opusrevision{$89.6 \pm 3.6$} & \opusrevision{$56.7 \pm 6.2$} & \opusrevision{$+32.3 \pm 15.4$} & \opusrevision{$+9.17 \pm 6.17$} \\
Claude Code + Opus 5 & 4.3 & $75.0 \pm 10.8$ & $50.8 \pm 8.8$ & $+39.6 \pm 22.2$ & $+13.13 \pm 7.98$ \\
\astrarevision{OpenCode + GPT-6 Astra} & \astrarevision{2} & \astrarevision{$72.9 \pm 3.6$} & \astrarevision{$47.1 \pm 1.4$} & \astrarevision{$-11.5 \pm 15.4$} & \astrarevision{$-5.00 \pm 4.96$} \\
Memory + Opus 5 & 3.3 & $70.8 \pm 14.4$ & $53.8 \pm 1.2$ & $+17.7 \pm 3.6$ & $+7.08 \pm 1.44$ \\
OpenCode + Gemini 3.7 Flash & 4 & $68.8 \pm 0.0$ & $46.2 \pm 0.0$ & $+6.2 \pm 0.0$ & $+0.62 \pm 0.00$ \\
OpenCode + GPT-5.6-SOL & 4 & $68.8 \pm 0.0$ & $46.2 \pm 0.0$ & $+6.2 \pm 0.0$ & $+0.62 \pm 0.00$ \\
Codex + GPT-5.6-SOL & 4.3 & $68.8 \pm 0.0$ & $45.8 \pm 0.7$ & $+11.4 \pm 9.0$ & $+3.12 \pm 4.33$ \\
AWM + GPT-5 Mini & 4.3 & $66.7 \pm 7.2$ & $45.4 \pm 5.6$ & $+19.8 \pm 19.1$ & $+5.00 \pm 5.96$ \\
EvoTest + Opus 5 & 3 & $64.6 \pm 15.7$ & $53.3 \pm 13.5$ & $+25.0 \pm 14.3$ & $+8.32 \pm 3.97$ \\
Reflexion + Opus 5 & 3 & $64.6 \pm 3.6$ & $51.7 \pm 3.1$ & $+18.7 \pm 8.3$ & $+6.66 \pm 2.53$ \\
Reflexion + Kimi K3 & 3 & $64.6 \pm 13.0$ & $31.7 \pm 12.5$ & $-7.3 \pm 17.2$ & $+0.21 \pm 6.32$ \\
OpenCode + Opus 5 & 5 & $64.6 \pm 7.2$ & $40.0 \pm 4.3$ & $+37.5 \pm 10.8$ & $+13.33 \pm 2.89$ \\
OpenCode + Qwen 3.8 Max & 3.3 & $62.5 \pm 28.6$ & $39.2 \pm 17.5$ & $+31.2 \pm 16.5$ & $+9.58 \pm 6.41$ \\
OpenCode + Gemini 3.5 Flash & 4.7 & $60.4 \pm 7.2$ & $40.8 \pm 6.3$ & $+21.9 \pm 9.4$ & $+6.46 \pm 3.77$ \\
ReasoningBank + GPT-5 Mini & 1.7 & $58.3 \pm 19.1$ & $47.1 \pm 10.8$ & $-3.1 \pm 14.3$ & $+0.00 \pm 3.75$ \\
Naive + GPT-5 Mini & 3.7 & $58.3 \pm 13.0$ & $43.8 \pm 8.8$ & $+3.2 \pm 11.3$ & $+1.05 \pm 2.01$ \\
ReasoningBank + Opus 5 & 3.7 & $58.3 \pm 9.5$ & $42.5 \pm 8.8$ & $+17.7 \pm 7.9$ & $+7.29 \pm 2.19$ \\
Naive + Kimi K3 & 4 & $54.2 \pm 26.0$ & $16.7 \pm 3.8$ & $+12.5 \pm 19.0$ & $+3.12 \pm 3.48$ \\
EvoTest + Kimi K3 & 2 & $52.1 \pm 29.5$ & $45.0 \pm 33.6$ & $-7.3 \pm 18.3$ & $-1.46 \pm 3.66$ \\
OpenCode + DeepSeek V4 Pro & 3.3 & $52.1 \pm 28.9$ & $34.2 \pm 20.9$ & $+3.1 \pm 5.4$ & $+0.21 \pm 0.72$ \\
EvoTest + GPT-5 Mini & 3.7 & $52.1 \pm 7.2$ & $43.8 \pm 6.6$ & $+10.4 \pm 7.2$ & $+3.33 \pm 2.60$ \\
OpenCode + Sonnet 5 & 3.3 & $47.9 \pm 26.0$ & $30.8 \pm 16.6$ & $-7.3 \pm 13.0$ & $+0.00 \pm 4.38$ \\
OpenCode + Kimi K3 & 3 & $43.8 \pm 21.7$ & $25.8 \pm 16.8$ & $+3.1 \pm 3.1$ & $+0.21 \pm 0.95$ \\
Memory + GPT-5 Mini & 1 & $39.6 \pm 3.6$ & $30.8 \pm 7.3$ & $-4.2 \pm 6.5$ & $-1.04 \pm 1.44$ \\
ReasoningBank + Kimi K3 & 2.3 & $39.6 \pm 29.5$ & $24.2 \pm 22.9$ & $-6.3 \pm 22.5$ & $+0.20 \pm 6.56$ \\
Reflexion + GPT-5 Mini & 1 & $37.5 \pm 18.8$ & $21.7 \pm 5.2$ & $-4.2 \pm 14.1$ & $-2.71 \pm 5.32$ \\
OpenCode + GPT-5 Mini & 2 & $31.2 \pm 0.0$ & $16.2 \pm 3.3$ & $-6.3 \pm 11.3$ & $-2.09 \pm 2.89$ \\
Memory + Kimi K3 & 4 & $31.2 \pm 0.0$ & $13.8 \pm 0.0$ & $+6.2 \pm 0.0$ & $+1.25 \pm 0.00$ \\
AWM + Kimi K3 & 1.3 & $25.0 \pm 32.5$ & $23.3 \pm 29.6$ & $+4.2 \pm 7.2$ & $+1.66 \pm 2.89$ \\
AWM + Opus 5 & 1 & $14.6 \pm 3.6$ & $12.9 \pm 0.7$ & $-1.0 \pm 1.8$ & $-0.41 \pm 0.72$ \\
Naive + Opus 5 & 1 & $12.5 \pm 0.0$ & $12.5 \pm 0.0$ & $+0.0 \pm 0.0$ & $+0.00 \pm 0.00$ \\
\bottomrule
\end{longtable}
\endgroup
\begin{longtable}{@{}p{0.07\linewidth}p{0.87\linewidth}@{}}
\caption{\captionrevision{\revised{Grapevine: criteria for assessing rule discovery.}}}\label{tab:targets-definition-grapevine}\\
\toprule
\textbf{Target} & \textbf{Criterion} \\
\midrule
\endfirsthead
\toprule
\textbf{Target} & \textbf{Criterion (continued)} \\
\midrule
\endhead
T1 & \revised{Choose at most three seeds in a deterministic 20-node linear-threshold cascade. Each episode allows one release. Feedback after each wave provides evidence about activation thresholds and incoming connections.} \\
T2 & \revised{Seed-1 synergy trap: best singles are T/D at 5, the greedy high-single trio reaches only about 8, while the exact optimum reaches 16.} \\
T3 & \revised{Before release, add or remove seeds to compare candidate seed sets across episodes and identify which nodes contribute additional activation.} \\
T4 & \revised{Deterministic exact-trio search over the discovered activation frontier, not choosing T/D/Q/L/M/B by single-node strength.} \\
T5 & \revised{16-reach triples: EFT activates KQS, then CMR, GH, JO, I, D, L and leaves A/B/N/P. EPT activates KQS, then CM, GHR, JO, I, D, L and leaves A/B/F/N.} \\
\bottomrule
\end{longtable}
\begin{center}
\begin{minipage}{\linewidth}
\centering
\captionof{table}{\captionrevision{\revised{Grapevine: rule-discovery evidence. S: complete evidence in at least one trial; P: partial evidence only; --: insufficient evidence. Evidence may include episodes beyond the scoring window.}}}
\label{tab:rq1-targets-game-grapevine}
{
\appendixtable{%
\begin{tabular}{@{}l*{5}{c}@{}}
\toprule
\textbf{Configuration} & T1 & T2 & T3 & T4 & T5 \\
\midrule
\multicolumn{6}{@{}l}{\textbf{backbone comparison}} \\
Claude Opus~5 & S & S & S & S & S \\
Qwen 3.8 Max & S & S & S & S & S \\
Kimi~K3 & S & -- & S & S & -- \\
Gemini~3.5 Flash & S & -- & -- & S & -- \\
Gemini~3.7 Flash & S & -- & S & S & -- \\
GPT-5.6-SOL & S & S & S & S & S \\
DeepSeek V4 Pro & S & -- & -- & -- & -- \\
Claude Sonnet~5 & S & S & S & S & -- \\
GPT-5 Mini & S & S & S & -- & -- \\
\midrule
\multicolumn{6}{@{}l}{\textbf{agent harness comparison}} \\
Claude Code + Opus~5 & S & S & S & S & S \\
Codex + GPT-5.6-SOL & S & S & S & S & S \\
\bottomrule
\end{tabular}
}
}

\end{minipage}
\end{center}

\subsection{\revised{Haunted Inn}}
\label{app:game-hauntedinn}

\revised{\textbf{Learning task.} Discover guest--room, adjacency, and count taboos through controlled placements. Re-shuffled game, scored seed 2, first 5 episodes, reference score $C=200$.}
\begin{center}
\includegraphics[width=0.95\linewidth]{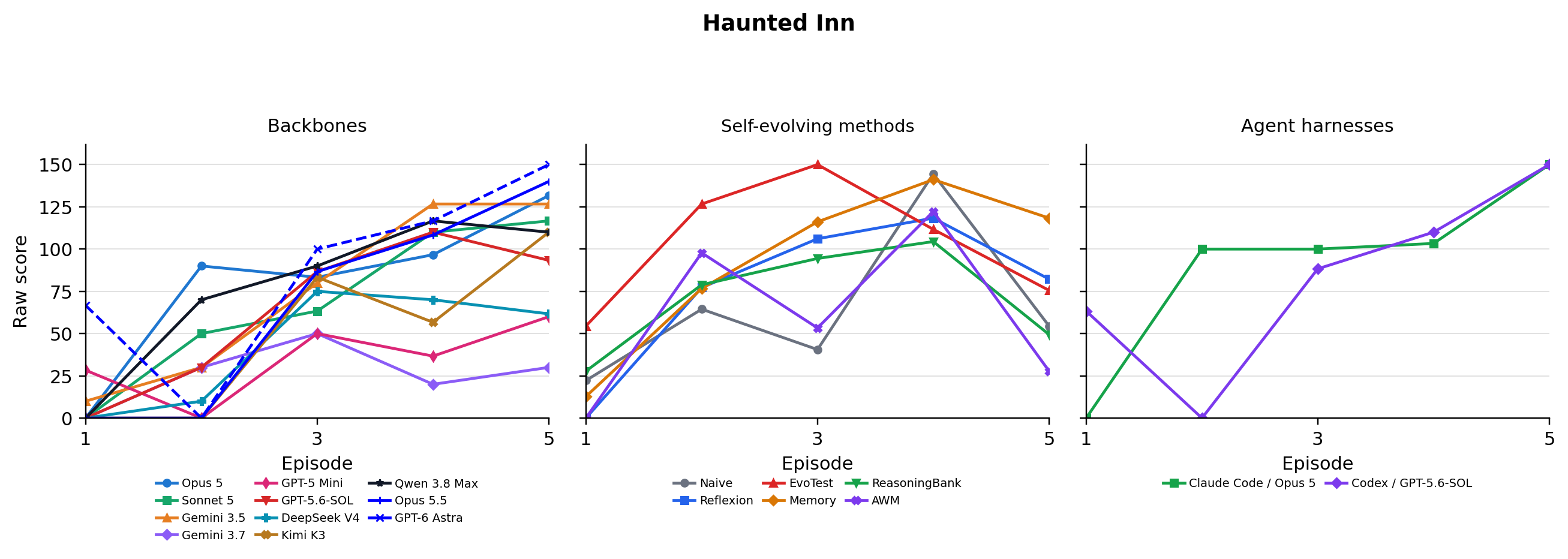}
\captionof{figure}{\revised{Haunted Inn: raw-score learning curves within the scoring window.}}
\label{fig:game-curve-hauntedinn}

\end{center}
\Needspace{10\baselineskip}
% Combined score table: normalized points and trial SD; human/First retained.
\begingroup
\setlength{\tabcolsep}{3pt}
\begin{longtable}{@{}lrrrrr@{}}
\caption{\captionrevision{Haunted Inn: normalized performance and learning, with trial SD. Rows follow descending Max. First denotes the average episode at which a trial first reaches its highest score.}}
\label{tab:game-scores-hauntedinn}\label{tab:trial-variability-hauntedinn}\\
\toprule
Configuration & First & Max $\uparrow$ & Mean $\uparrow$ & LG $\uparrow$ & LS $\uparrow$ \\
\midrule
\endfirsthead
\toprule
Configuration & First & Max $\uparrow$ & Mean $\uparrow$ & LG $\uparrow$ & LS $\uparrow$ \\
\midrule
\endhead
Human Top-1 & 1 & 100.0 & 50.0 & -- & -- \\
EvoTest + Kimi K3 & 1.3 & $100.0 \pm 0.0$ & $80.0 \pm 20.0$ & $-16.7 \pm 28.9$ & $-3.33 \pm 5.77$ \\
Naive + Kimi K3 & 3 & $100.0 \pm 0.0$ & $35.3 \pm 13.6$ & $+50.0 \pm 50.0$ & $+13.33 \pm 15.28$ \\
\astrarevision{OpenCode + GPT-6 Astra} & \astrarevision{3} & \astrarevision{$91.7 \pm 7.2$} & \astrarevision{$43.3 \pm 1.4$} & \astrarevision{$+50.0 \pm 54.1$} & \astrarevision{$+14.17 \pm 16.60$} \\
Naive + GPT-5 Mini & 3.7 & $83.3 \pm 14.4$ & $37.5 \pm 10.9$ & $+21.7 \pm 15.9$ & $+5.83 \pm 5.77$ \\
Codex + GPT-5.6-SOL & 4.7 & $83.3 \pm 14.4$ & $41.2 \pm 19.6$ & $+49.2 \pm 18.1$ & $+14.17 \pm 7.78$ \\
Reflexion + Opus 5 & 3.3 & $78.3 \pm 15.9$ & $41.3 \pm 15.3$ & $+31.7 \pm 15.8$ & $+9.67 \pm 6.83$ \\
OpenCode + Qwen 3.8 Max & 4.3 & $78.3 \pm 8.0$ & $38.7 \pm 8.6$ & $+39.2 \pm 18.3$ & $+13.33 \pm 2.53$ \\
Memory + GPT-5 Mini & 2.7 & $77.5 \pm 21.4$ & $46.0 \pm 10.9$ & $+32.9 \pm 26.3$ & $+9.92 \pm 8.43$ \\
Memory + Kimi K3 & 3 & $75.0 \pm 0.0$ & $49.2 \pm 0.6$ & $+49.6 \pm 1.4$ & $+15.92 \pm 0.29$ \\
Reflexion + Kimi K3 & 3 & $75.0 \pm 0.0$ & $49.0 \pm 0.9$ & $+47.5 \pm 2.2$ & $+15.25 \pm 0.87$ \\
AWM + Kimi K3 & 3 & $75.0 \pm 25.0$ & $29.2 \pm 14.2$ & $+6.2 \pm 10.8$ & $+1.25 \pm 2.17$ \\
AWM + Opus 5 & 4 & $75.0 \pm 0.0$ & $29.0 \pm 7.8$ & $+25.8 \pm 21.0$ & $+7.50 \pm 8.23$ \\
Naive + Opus 5 & 4 & $75.0 \pm 0.0$ & $25.0 \pm 0.0$ & $+12.5 \pm 0.0$ & $+2.50 \pm 0.00$ \\
Claude Code + Opus 5 & 5 & $75.0 \pm 0.0$ & $45.3 \pm 1.5$ & $+38.3 \pm 10.1$ & $+15.17 \pm 2.02$ \\
Memory + Opus 5 & 3.7 & $74.2 \pm 13.8$ & $44.3 \pm 12.1$ & $+45.0 \pm 6.5$ & $+15.50 \pm 1.80$ \\
EvoTest + GPT-5 Mini & 3 & $71.7 \pm 25.0$ & $39.0 \pm 12.8$ & $+3.3 \pm 21.3$ & $+0.83 \pm 9.00$ \\
AWM + GPT-5 Mini & 3.3 & $71.7 \pm 17.7$ & $32.2 \pm 6.3$ & $+7.1 \pm 8.3$ & $+3.25 \pm 1.52$ \\
OpenCode + Gemini 3.5 Flash & 4.3 & $71.7 \pm 17.7$ & $37.3 \pm 2.0$ & $+53.3 \pm 33.5$ & $+16.50 \pm 9.50$ \\
OpenCode + Sonnet 5 & 4.3 & $71.7 \pm 17.7$ & $34.0 \pm 18.7$ & $+44.2 \pm 6.4$ & $+14.67 \pm 3.92$ \\
\opusrevision{OpenCode + Opus 5.5} & \opusrevision{4.7} & \opusrevision{$70.8 \pm 9.5$} & \opusrevision{$33.5 \pm 5.8$} & \opusrevision{$+62.1 \pm 14.4$} & \opusrevision{$+19.42 \pm 3.66$} \\
ReasoningBank + Kimi K3 & 3.7 & $70.0 \pm 8.7$ & $35.5 \pm 7.1$ & $+33.8 \pm 1.2$ & $+10.58 \pm 0.14$ \\
EvoTest + Opus 5 & 3.3 & $68.3 \pm 6.3$ & $36.5 \pm 5.8$ & $+17.9 \pm 8.3$ & $+6.58 \pm 4.72$ \\
OpenCode + Opus 5 & 4.7 & $67.5 \pm 13.0$ & $40.2 \pm 2.3$ & $+34.6 \pm 6.9$ & $+13.50 \pm 2.88$ \\
ReasoningBank + Opus 5 & 4 & $66.7 \pm 10.4$ & $38.3 \pm 3.6$ & $+26.7 \pm 8.0$ & $+8.00 \pm 3.91$ \\
OpenCode + Kimi K3 & 4.3 & $66.7 \pm 14.4$ & $25.0 \pm 8.7$ & $+41.7 \pm 19.1$ & $+13.83 \pm 6.83$ \\
ReasoningBank + GPT-5 Mini & 2.3 & $63.3 \pm 10.1$ & $32.7 \pm 12.0$ & $-25.0 \pm 22.9$ & $-8.25 \pm 8.27$ \\
OpenCode + GPT-5.6-SOL & 3.3 & $61.7 \pm 22.7$ & $32.0 \pm 9.7$ & $+43.3 \pm 35.8$ & $+13.33 \pm 9.27$ \\
Reflexion + GPT-5 Mini & 3 & $49.2 \pm 3.8$ & $24.8 \pm 5.8$ & $+13.3 \pm 10.0$ & $+5.92 \pm 1.63$ \\
OpenCode + DeepSeek V4 Pro & 3.7 & $48.3 \pm 2.9$ & $21.7 \pm 4.5$ & $+30.4 \pm 9.0$ & $+9.17 \pm 3.21$ \\
OpenCode + GPT-5 Mini & 4.7 & $30.0 \pm 26.0$ & $17.5 \pm 18.2$ & $+17.1 \pm 10.8$ & $+5.00 \pm 2.29$ \\
OpenCode + Gemini 3.7 Flash & 3 & $25.0 \pm 13.0$ & $13.0 \pm 3.8$ & $+5.0 \pm 4.3$ & $+2.50 \pm 0.87$ \\
\bottomrule
\end{longtable}
\endgroup

\revised{\textbf{Seed-matched evidence.} Labels for T2 and T6 use the three seed-2 trials used for score reporting. This seed-matched assessment covers these two targets, not the other labels, and does not determine the episode in which a target was discovered.}
\begin{longtable}{@{}p{0.07\linewidth}p{0.87\linewidth}@{}}
\caption{\captionrevision{\revised{Haunted Inn: criteria for assessing rule discovery.}}}\label{tab:targets-definition-haunted-inn}\\
\toprule
\textbf{Target} & \textbf{Criterion} \\
\midrule
\endfirsthead
\toprule
\textbf{Target} & \textbf{Criterion (continued)} \\
\midrule
\endhead
T1 & \revised{Eight rooms with visible fixed features. For the evaluated seed 2, rooms 1, 3, 6, and 7 are plain. Room 2 has a shrine, room 4 a window, room 5 is dark, and room 8 has a mirror.} \\
T2 & \revised{All three hidden taboos for the evaluated seed 2: housing at least three actors haunts every actor room. Adjacent actors and scholars haunt both rooms. Any lantern-bearing guest in the mirror room is haunted.} \\
T3 & \revised{Clues identify rooms that trigger taboos. Panic then empties the entire connected block of occupied rooms. An empty room stops that spread.} \\
T4 & \revised{Place a few guests at a time to distinguish the room that triggered a taboo from rooms affected by spreading panic. Use these comparisons to separate adjacency rules from guest-count rules.} \\
T5 & \revised{Fee-preserving packing under constraints: keep high-fee merchants/actors when legal, turn away or separate conflicts, and use empty gaps only when they save more fee than they cost.} \\
T6 & \revised{200-point seating policy on favorable queues: avoid the three taboo triggers, insert firebreaks where needed, and do not sacrifice high-fee guests for untested superstition.} \\
\bottomrule
\end{longtable}
\begin{center}
\begin{minipage}{\linewidth}
\centering
\captionof{table}{\captionrevision{\revised{Haunted Inn: rule-discovery evidence. S: complete evidence in at least one trial; P: partial evidence only; --: insufficient evidence. Evidence may include episodes beyond the scoring window.}}}
\label{tab:rq1-targets-game-haunted-inn}
{
\appendixtable{%
\begin{tabular}{@{}l*{6}{c}@{}}
\toprule
\textbf{Configuration} & T1 & T2 & T3 & T4 & T5 & T6 \\
\midrule
\multicolumn{7}{@{}l}{\textbf{backbone comparison}} \\
Claude Opus~5 & S & P & S & S & S & P \\
Qwen 3.8 Max & S & P & S & S & S & P \\
Kimi~K3 & S & S & S & S & S & P \\
Gemini~3.5 Flash & S & P & S & S & S & P \\
Gemini~3.7 Flash & S & P & S & S & S & P \\
GPT-5.6-SOL & S & P & S & S & S & P \\
DeepSeek V4 Pro & S & P & S & S & S & P \\
Claude Sonnet~5 & S & P & S & S & S & P \\
GPT-5 Mini & S & P & S & S & S & P \\
\midrule
\multicolumn{7}{@{}l}{\textbf{agent harness comparison}} \\
Claude Code + Opus~5 & S & P & S & S & S & P \\
Codex + GPT-5.6-SOL & S & S & S & S & S & P \\
\bottomrule
\end{tabular}
}
}

\end{minipage}
\end{center}

\subsection{\revised{Castaway}}
\label{app:game-castaway}

\revised{\textbf{Learning task.} Remember the map, recipes, hazards, and steps needed to escape. Fixed game, scored seed 1, first 5 episodes, reference score $C=241$.}
\begin{center}
\includegraphics[width=0.95\linewidth]{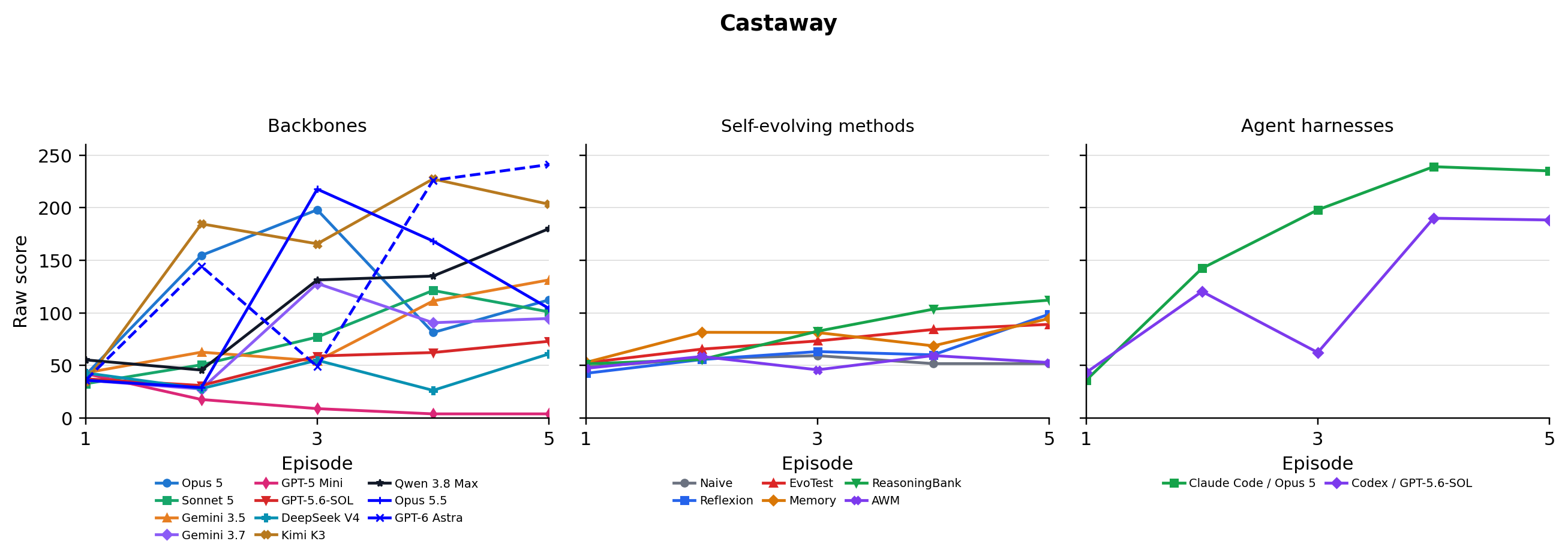}
\captionof{figure}{\revised{Castaway: raw-score learning curves within the scoring window.}}
\label{fig:game-curve-castaway}

\end{center}
\Needspace{10\baselineskip}
% Combined score table: normalized points and trial SD; human/First retained.
\begingroup
\setlength{\tabcolsep}{3pt}
\begin{longtable}{@{}lrrrrr@{}}
\caption{\captionrevision{Castaway: normalized performance and learning, with trial SD. Rows follow descending Max. First denotes the average episode at which a trial first reaches its highest score.}}
\label{tab:game-scores-castaway}\label{tab:trial-variability-castaway}\\
\toprule
Configuration & First & Max $\uparrow$ & Mean $\uparrow$ & LG $\uparrow$ & LS $\uparrow$ \\
\midrule
\endfirsthead
\toprule
Configuration & First & Max $\uparrow$ & Mean $\uparrow$ & LG $\uparrow$ & LS $\uparrow$ \\
\midrule
\endhead
Human Top-1 & 4 & 100.0 & 54.8 & -- & -- \\
\astrarevision{OpenCode + GPT-6 Astra} & \astrarevision{4.3} & \astrarevision{$100.0 \pm 0.0$} & \astrarevision{$57.8 \pm 6.9$} & \astrarevision{$+59.5 \pm 12.2$} & \astrarevision{$+20.40 \pm 2.33$} \\
Claude Code + Opus 5 & 2.7 & $99.4 \pm 1.0$ & $70.6 \pm 3.5$ & $+61.3 \pm 19.2$ & $+20.53 \pm 3.50$ \\
OpenCode + Kimi K3 & 4 & $94.3 \pm 1.9$ & $67.7 \pm 4.4$ & $+43.8 \pm 13.6$ & $+15.77 \pm 4.48$ \\
Codex + GPT-5.6-SOL & 3.7 & $94.2 \pm 3.5$ & $50.2 \pm 7.5$ & $+44.5 \pm 32.8$ & $+14.92 \pm 8.25$ \\
\opusrevision{OpenCode + Opus 5.5} & \opusrevision{3.7} & \opusrevision{$92.8 \pm 7.4$} & \opusrevision{$46.1 \pm 14.6$} & \opusrevision{$+43.1 \pm 39.6$} & \opusrevision{$+11.45 \pm 10.81$} \\
OpenCode + Qwen 3.8 Max & 4.7 & $89.2 \pm 9.7$ & $45.4 \pm 15.0$ & $+44.4 \pm 9.2$ & $+14.05 \pm 0.95$ \\
OpenCode + Opus 5 & 3.7 & $83.7 \pm 25.0$ & $48.7 \pm 23.3$ & $-0.4 \pm 24.2$ & $+2.88 \pm 8.44$ \\
ReasoningBank + Kimi K3 & 4.7 & $81.5 \pm 22.1$ & $47.0 \pm 10.9$ & $+47.6 \pm 20.7$ & $+15.02 \pm 6.27$ \\
Memory + Kimi K3 & 4 & $70.0 \pm 22.1$ & $38.0 \pm 9.6$ & $+16.3 \pm 13.2$ & $+6.28 \pm 5.23$ \\
EvoTest + Opus 5 & 3 & $69.4 \pm 41.0$ & $48.5 \pm 34.9$ & $+29.8 \pm 30.7$ & $+10.12 \pm 9.76$ \\
OpenCode + Sonnet 5 & 4.7 & $67.2 \pm 22.6$ & $31.8 \pm 12.9$ & $+28.8 \pm 9.2$ & $+8.60 \pm 3.51$ \\
OpenCode + Gemini 3.5 Flash & 4.3 & $66.4 \pm 19.8$ & $33.4 \pm 5.7$ & $+28.4 \pm 9.4$ & $+9.35 \pm 5.06$ \\
Reflexion + Kimi K3 & 4.3 & $55.5 \pm 26.9$ & $31.5 \pm 13.2$ & $+25.9 \pm 32.1$ & $+8.69 \pm 9.38$ \\
OpenCode + Gemini 3.7 Flash & 3.7 & $54.1 \pm 29.6$ & $31.3 \pm 14.7$ & $+25.2 \pm 21.5$ & $+7.48 \pm 5.87$ \\
Memory + Opus 5 & 2.7 & $51.5 \pm 36.8$ & $34.6 \pm 23.0$ & $+3.8 \pm 5.5$ & $+3.20 \pm 3.90$ \\
ReasoningBank + Opus 5 & 4 & $50.1 \pm 10.2$ & $34.1 \pm 4.2$ & $+14.6 \pm 15.0$ & $+4.58 \pm 3.56$ \\
Reflexion + Opus 5 & 5 & $42.2 \pm 2.0$ & $29.8 \pm 6.6$ & $+6.5 \pm 5.9$ & $+3.28 \pm 1.76$ \\
Naive + Opus 5 & 3 & $39.3 \pm 14.3$ & $22.5 \pm 3.3$ & $+2.9 \pm 9.1$ & $+1.13 \pm 2.14$ \\
OpenCode + GPT-5.6-SOL & 4 & $38.9 \pm 25.8$ & $21.8 \pm 12.0$ & $+13.9 \pm 26.0$ & $+4.26 \pm 6.48$ \\
AWM + Opus 5 & 3.3 & $38.7 \pm 7.5$ & $25.1 \pm 3.8$ & $+9.3 \pm 8.9$ & $+2.38 \pm 2.64$ \\
AWM + Kimi K3 & 3 & $33.5 \pm 21.4$ & $21.5 \pm 5.1$ & $-4.8 \pm 9.0$ & $-0.73 \pm 1.42$ \\
Naive + GPT-5 Mini & 2.7 & $32.4 \pm 1.4$ & $22.6 \pm 3.1$ & $-2.8 \pm 5.5$ & $-0.73 \pm 2.48$ \\
Reflexion + GPT-5 Mini & 4 & $32.1 \pm 12.6$ & $18.5 \pm 4.1$ & $+5.4 \pm 4.7$ & $+2.59 \pm 1.75$ \\
OpenCode + DeepSeek V4 Pro & 3 & $30.4 \pm 14.2$ & $17.7 \pm 6.4$ & $+3.4 \pm 15.4$ & $+1.42 \pm 5.21$ \\
Memory + GPT-5 Mini & 3 & $30.0 \pm 9.5$ & $21.8 \pm 12.5$ & $-2.1 \pm 8.9$ & $-0.68 \pm 2.72$ \\
Naive + Kimi K3 & 2 & $27.7 \pm 3.4$ & $21.6 \pm 4.4$ & $-1.0 \pm 9.4$ & $-0.17 \pm 2.99$ \\
ReasoningBank + GPT-5 Mini & 1.7 & $26.7 \pm 15.0$ & $19.8 \pm 6.0$ & $+5.3 \pm 10.7$ & $+1.44 \pm 3.01$ \\
EvoTest + Kimi K3 & 4 & $25.7 \pm 4.3$ & $23.0 \pm 3.2$ & $+3.6 \pm 1.0$ & $+1.33 \pm 0.50$ \\
AWM + GPT-5 Mini & 2.3 & $23.5 \pm 7.5$ & $19.1 \pm 4.5$ & $-0.6 \pm 3.1$ & $-0.18 \pm 0.74$ \\
EvoTest + GPT-5 Mini & 2.7 & $23.5 \pm 6.2$ & $19.3 \pm 3.3$ & $+1.0 \pm 5.0$ & $-0.03 \pm 1.92$ \\
OpenCode + GPT-5 Mini & 1 & $17.6 \pm 5.3$ & $6.4 \pm 3.5$ & $-10.8 \pm 6.9$ & $-3.75 \pm 1.90$ \\
\bottomrule
\end{longtable}
\endgroup
\begin{longtable}{@{}p{0.07\linewidth}p{0.87\linewidth}@{}}
\caption{\captionrevision{\revised{Castaway: criteria for assessing rule discovery.}}}\label{tab:targets-definition-castaway}\\
\toprule
\textbf{Target} & \textbf{Criterion} \\
\midrule
\endfirsthead
\toprule
\textbf{Target} & \textbf{Criterion (continued)} \\
\midrule
\endhead
T1 & \revised{Seven-place seed-1 island map from beach to forest, swamp, cave, ravine, bamboo, and ruins.} \\
T2 & \revised{Resource/recipe table: wood/flint/vine/water, seed-1 food effects, beach fire, torch, tool, and raft, including mandatory tool and two-vine raft lashing.} \\
T3 & \revised{Survival and hazard model: 56 segments, five segments per day, hunger drain, day survival points, cave current, swamp mosquitoes/collapse, bamboo falling rocks/night trap, ruins crab, ravine boar, and forest monkey.} \\
T4 & \revised{True-vs-decoy escape chain: frame at cave, sail at forest, rudder at ruins are real, while compass at bamboo is the decoy that makes raft crafting fail if used as a substitute.} \\
T5 & \revised{241-point script: build fire/tool/torch, route around the two lethal traps, neutralize guards, collect all real raft parts while optionally collecting the decoy for information/score, safely stall for day points, and craft the raft at the beach.} \\
\bottomrule
\end{longtable}
\begin{center}
\begin{minipage}{\linewidth}
\centering
\captionof{table}{\captionrevision{\revised{Castaway: rule-discovery evidence. S: complete evidence in at least one trial; P: partial evidence only; --: insufficient evidence. Evidence may include episodes beyond the scoring window.}}}
\label{tab:rq1-targets-game-castaway}
{
\appendixtable{%
\begin{tabular}{@{}l*{5}{c}@{}}
\toprule
\textbf{Configuration} & T1 & T2 & T3 & T4 & T5 \\
\midrule
\multicolumn{6}{@{}l}{\textbf{backbone comparison}} \\
Claude Opus~5 & S & S & S & S & S \\
Qwen 3.8 Max & S & S & S & S & S \\
Kimi~K3 & S & S & S & S & S \\
Gemini~3.5 Flash & S & S & S & -- & -- \\
Gemini~3.7 Flash & S & S & S & -- & -- \\
GPT-5.6-SOL & S & S & S & -- & -- \\
DeepSeek V4 Pro & S & S & S & -- & -- \\
Claude Sonnet~5 & S & S & S & -- & -- \\
GPT-5 Mini & S & S & S & S & -- \\
\midrule
\multicolumn{6}{@{}l}{\textbf{agent harness comparison}} \\
Claude Code + Opus~5 & S & S & S & S & S \\
Codex + GPT-5.6-SOL & S & S & S & S & S \\
\bottomrule
\end{tabular}
}
}

\end{minipage}
\end{center}

\subsection{\revised{Roadside Observatory}}
\label{app:game-roadside_observatory}

\revised{\textbf{Learning task.} Learn social protocols and resource requirements for new encounters. Re-shuffled game, scored seed 2, first 10 episodes, reference score $C=1000$.}
\begin{center}
\includegraphics[width=0.95\linewidth]{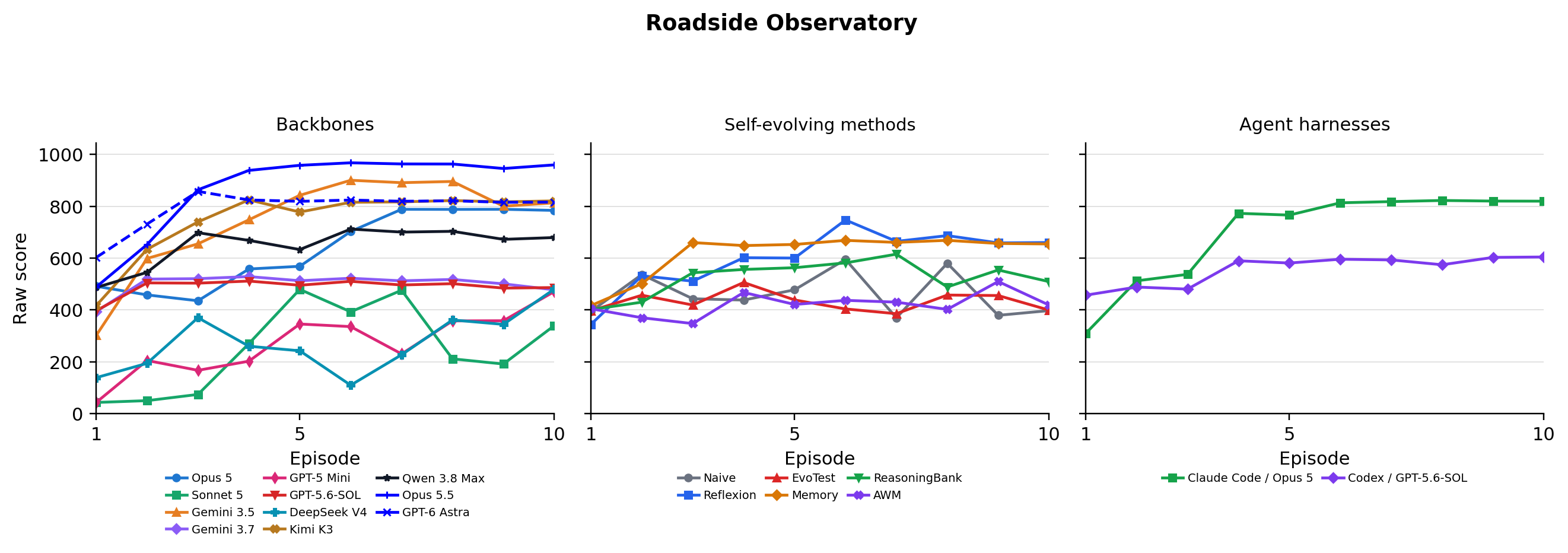}
\captionof{figure}{\revised{Roadside Observatory: raw-score learning curves within the scoring window.}}
\label{fig:game-curve-roadside_observatory}

\end{center}
\Needspace{10\baselineskip}
% Combined score table: normalized points and trial SD; human/First retained.
\begingroup
\setlength{\tabcolsep}{3pt}
\begin{longtable}{@{}lrrrrr@{}}
\caption{\captionrevision{Roadside Observatory: normalized performance and learning, with trial SD. Rows follow descending Max. First denotes the average episode at which a trial first reaches its highest score.}}
\label{tab:game-scores-roadside_observatory}\label{tab:trial-variability-roadside-observatory}\\
\toprule
Configuration & First & Max $\uparrow$ & Mean $\uparrow$ & LG $\uparrow$ & LS $\uparrow$ \\
\midrule
\endfirsthead
\toprule
Configuration & First & Max $\uparrow$ & Mean $\uparrow$ & LG $\uparrow$ & LS $\uparrow$ \\
\midrule
\endhead
\opusrevision{OpenCode + Opus 5.5} & \opusrevision{4.3} & \opusrevision{$96.8 \pm 5.6$} & \opusrevision{$87.0 \pm 8.3$} & \opusrevision{$+28.8 \pm 9.0$} & \opusrevision{$+4.17 \pm 1.05$} \\
Reflexion + Opus 5 & 6 & $94.5 \pm 5.4$ & $74.0 \pm 10.3$ & $+33.1 \pm 8.4$ & $+4.73 \pm 1.46$ \\
OpenCode + Gemini 3.5 Flash & 5.3 & $90.1 \pm 3.0$ & $74.5 \pm 15.8$ & $+31.8 \pm 28.5$ & $+4.67 \pm 4.19$ \\
Human Top-1 & 7 & 89.6 & 47.9 & -- & -- \\
\astrarevision{OpenCode + GPT-6 Astra} & \astrarevision{2.3} & \astrarevision{$85.7 \pm 24.7$} & \astrarevision{$79.3 \pm 33.8$} & \astrarevision{$+8.7 \pm 8.4$} & \astrarevision{$+1.41 \pm 1.34$} \\
Claude Code + Opus 5 & 8.3 & $82.7 \pm 25.7$ & $69.9 \pm 18.2$ & $+36.8 \pm 22.6$ & $+5.06 \pm 2.99$ \\
OpenCode + Kimi K3 & 5.3 & $82.5 \pm 27.6$ & $74.8 \pm 26.2$ & $+22.3 \pm 4.4$ & $+3.24 \pm 0.45$ \\
Naive + Opus 5 & 6.7 & $79.4 \pm 13.3$ & $57.5 \pm 4.8$ & $+5.4 \pm 8.2$ & $+0.64 \pm 0.60$ \\
OpenCode + Opus 5 & 7.7 & $79.2 \pm 11.2$ & $63.6 \pm 8.3$ & $+32.5 \pm 11.2$ & $+4.57 \pm 1.46$ \\
Reflexion + GPT-5 Mini & 4.7 & $77.4 \pm 17.9$ & $57.8 \pm 6.5$ & $+4.5 \pm 7.3$ & $+0.77 \pm 1.27$ \\
Memory + Kimi K3 & 5 & $77.4 \pm 23.8$ & $70.8 \pm 20.7$ & $+16.4 \pm 13.6$ & $+2.65 \pm 1.93$ \\
OpenCode + Qwen 3.8 Max & 8 & $74.1 \pm 21.2$ & $65.0 \pm 19.6$ & $+10.8 \pm 10.1$ & $+1.71 \pm 1.53$ \\
Memory + Opus 5 & 4.3 & $73.6 \pm 26.6$ & $66.4 \pm 21.1$ & $+22.0 \pm 16.8$ & $+3.15 \pm 2.43$ \\
ReasoningBank + Kimi K3 & 6 & $73.5 \pm 24.0$ & $47.6 \pm 9.9$ & $+0.6 \pm 11.2$ & $+0.73 \pm 1.60$ \\
ReasoningBank + GPT-5 Mini & 3.3 & $71.2 \pm 11.3$ & $52.1 \pm 6.7$ & $-1.7 \pm 7.3$ & $-0.24 \pm 0.74$ \\
Naive + Kimi K3 & 4 & $68.2 \pm 21.9$ & $37.1 \pm 4.6$ & $-3.3 \pm 16.4$ & $-0.67 \pm 1.31$ \\
ReasoningBank + Opus 5 & 8 & $67.5 \pm 28.1$ & $57.5 \pm 22.2$ & $+18.8 \pm 22.4$ & $+2.71 \pm 3.09$ \\
Reflexion + Kimi K3 & 7.3 & $67.5 \pm 15.3$ & $48.4 \pm 15.8$ & $+24.2 \pm 21.4$ & $+3.49 \pm 2.60$ \\
EvoTest + Kimi K3 & 2 & $63.8 \pm 18.0$ & $46.7 \pm 1.5$ & $-9.1 \pm 17.1$ & $-1.50 \pm 2.70$ \\
Codex + GPT-5.6-SOL & 6.7 & $62.1 \pm 15.8$ & $55.7 \pm 9.3$ & $+11.9 \pm 14.9$ & $+1.59 \pm 1.92$ \\
AWM + GPT-5 Mini & 5 & $60.0 \pm 7.2$ & $40.8 \pm 7.9$ & $+0.6 \pm 5.0$ & $-0.40 \pm 0.59$ \\
Naive + GPT-5 Mini & 6 & $59.3 \pm 3.8$ & $43.7 \pm 4.4$ & $-3.9 \pm 12.8$ & $-0.85 \pm 1.53$ \\
AWM + Kimi K3 & 8 & $59.1 \pm 23.0$ & $42.2 \pm 6.2$ & $+17.6 \pm 13.5$ & $+2.40 \pm 1.78$ \\
EvoTest + Opus 5 & 2 & $57.5 \pm 0.3$ & $53.2 \pm 2.2$ & $-2.7 \pm 2.8$ & $-0.35 \pm 0.32$ \\
OpenCode + Gemini 3.7 Flash & 4 & $52.9 \pm 1.8$ & $50.1 \pm 1.7$ & $+2.1 \pm 4.8$ & $+0.35 \pm 0.79$ \\
Memory + GPT-5 Mini & 4.7 & $51.3 \pm 8.9$ & $48.4 \pm 8.8$ & $+1.8 \pm 0.9$ & $+0.25 \pm 0.24$ \\
OpenCode + GPT-5.6-SOL & 4 & $51.1 \pm 0.3$ & $48.9 \pm 0.5$ & $+2.2 \pm 1.1$ & $+0.38 \pm 0.18$ \\
EvoTest + GPT-5 Mini & 3.7 & $49.5 \pm 11.1$ & $29.7 \pm 5.7$ & $+15.8 \pm 11.4$ & $+1.49 \pm 1.01$ \\
OpenCode + Sonnet 5 & 5.3 & $49.5 \pm 1.2$ & $25.3 \pm 5.6$ & $+19.2 \pm 21.5$ & $+2.96 \pm 3.01$ \\
OpenCode + DeepSeek V4 Pro & 5.7 & $49.4 \pm 1.7$ & $27.3 \pm 14.6$ & $+16.1 \pm 9.5$ & $+2.33 \pm 1.47$ \\
OpenCode + GPT-5 Mini & 5.7 & $48.9 \pm 0.7$ & $27.2 \pm 13.7$ & $+25.7 \pm 16.7$ & $+3.61 \pm 1.90$ \\
AWM + Opus 5 & 6.3 & $47.0 \pm 5.0$ & $43.2 \pm 3.6$ & $+2.7 \pm 5.0$ & $+0.33 \pm 0.63$ \\
\bottomrule
\end{longtable}
\endgroup
\begin{longtable}{@{}p{0.07\linewidth}p{0.87\linewidth}@{}}
\caption{\captionrevision{\revised{Roadside Observatory: criteria for assessing rule discovery.}}}\label{tab:targets-definition-roadside-observatory}\\
\toprule
\textbf{Target} & \textbf{Criterion} \\
\midrule
\endfirsthead
\toprule
\textbf{Target} & \textbf{Criterion (continued)} \\
\midrule
\endhead
T1 & \revised{Travel-state accounting for distance, time, energy, supplies, cash, rest, bus, safe hitching, and main-road walking.} \\
T2 & \revised{Offer to help the mechanic before asking for a ride. Asking without first helping wastes the encounter.} \\
T3 & \revised{Keep energy at least 50. Ask the ranger about weather and the trail before asking for the shortcut.} \\
T4 & \revised{Hiker protocol: pay supply/cash or forage, then listen for the route clue.} \\
T5 & \revised{Collector protocol: visit landmark or keep token, then tell story/show token to receive the coded map clue.} \\
T6 & \revised{Driver protocol and trap avoidance: refuse offer, question detail, then accept only the verified safe ride.} \\
T7 & \revised{Observatory gate: obtain the map-clue item and show it at the final encounter with enough energy.} \\
T8 & \revised{Transport/score optimization: rest when energy is low, use bus only when cash permits, avoid slow-route tags, and preserve time/resource bonuses.} \\
T9 & \revised{1,000-point arrival: complete all five social protocols, visit landmark, show the map clue, arrive during the viewing window, and avoid risky-route/false-help penalties.} \\
\bottomrule
\end{longtable}
\begin{center}
\begin{minipage}{\linewidth}
\centering
\captionof{table}{\captionrevision{\revised{Roadside Observatory: rule-discovery evidence. S: complete evidence in at least one trial; P: partial evidence only; --: insufficient evidence. Evidence may include episodes beyond the scoring window.}}}
\label{tab:rq1-targets-game-roadside-observatory}
{
\appendixtable{%
\begin{tabular}{@{}l*{9}{c}@{}}
\toprule
\textbf{Configuration} & T1 & T2 & T3 & T4 & T5 & T6 & T7 & T8 & T9 \\
\midrule
\multicolumn{10}{@{}l}{\textbf{backbone comparison}} \\
Claude Opus~5 & S & S & S & S & S & S & S & S & -- \\
Qwen 3.8 Max & S & S & S & S & S & S & S & S & S \\
Kimi~K3 & S & S & S & S & S & S & S & S & S \\
Gemini~3.5 Flash & S & S & S & S & S & S & S & S & -- \\
Gemini~3.7 Flash & S & S & S & S & S & S & S & S & -- \\
GPT-5.6-SOL & S & S & S & S & S & S & -- & S & -- \\
DeepSeek V4 Pro & S & S & S & S & S & S & -- & S & -- \\
Claude Sonnet~5 & S & S & S & S & S & S & -- & S & -- \\
GPT-5 Mini & S & -- & -- & -- & -- & -- & -- & S & -- \\
\midrule
\multicolumn{10}{@{}l}{\textbf{agent harness comparison}} \\
Claude Code + Opus~5 & S & S & S & S & S & S & S & S & S \\
Codex + GPT-5.6-SOL & S & S & S & S & S & S & S & S & -- \\
\bottomrule
\end{tabular}
}
}

\end{minipage}
\end{center}

\subsection{\revised{Lost and Found}}
\label{app:game-lost_and_found}

\revised{\textbf{Learning task.} Learn how to verify ownership without revealing private information. Re-shuffled game, scored seed 1, first 5 episodes, reference score $C=1000$.}
\begin{center}
\includegraphics[width=0.95\linewidth]{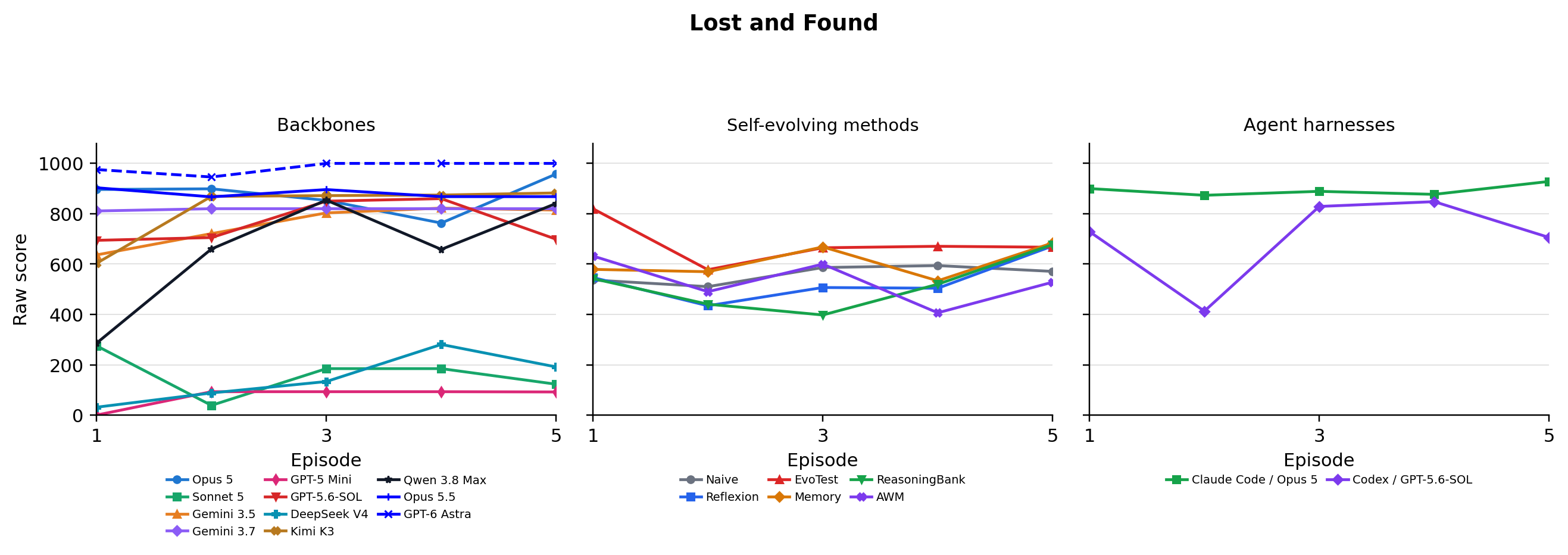}
\captionof{figure}{\revised{Lost and Found: raw-score learning curves within the scoring window.}}
\label{fig:game-curve-lost_and_found}

\end{center}
\Needspace{10\baselineskip}
% Combined score table: normalized points and trial SD; human/First retained.
\begingroup
\setlength{\tabcolsep}{3pt}
\begin{longtable}{@{}lrrrrr@{}}
\caption{\captionrevision{Lost and Found: normalized performance and learning, with trial SD. Rows follow descending Max. First denotes the average episode at which a trial first reaches its highest score.}}
\label{tab:game-scores-lost_and_found}\label{tab:trial-variability-lost-and-found}\\
\toprule
Configuration & First & Max $\uparrow$ & Mean $\uparrow$ & LG $\uparrow$ & LS $\uparrow$ \\
\midrule
\endfirsthead
\toprule
Configuration & First & Max $\uparrow$ & Mean $\uparrow$ & LG $\uparrow$ & LS $\uparrow$ \\
\midrule
\endhead
AWM + Opus 5 & 3.3 & $100.0 \pm 0.0$ & $90.9 \pm 5.7$ & $-2.7 \pm 20.6$ & $+0.13 \pm 4.48$ \\
\astrarevision{OpenCode + GPT-6 Astra} & \astrarevision{1.7} & \astrarevision{$100.0 \pm 0.0$} & \astrarevision{$98.4 \pm 2.0$} & \astrarevision{$+3.9 \pm 5.1$} & \astrarevision{$+1.03 \pm 1.43$} \\
Memory + Opus 5 & 4 & $96.8 \pm 5.6$ & $87.7 \pm 10.8$ & $+6.0 \pm 5.5$ & $+1.92 \pm 1.34$ \\
OpenCode + Opus 5 & 5 & $95.8 \pm 4.2$ & $87.4 \pm 6.7$ & $-3.7 \pm 14.7$ & $-0.13 \pm 3.30$ \\
Memory + Kimi K3 & 3.7 & $95.1 \pm 0.2$ & $81.0 \pm 15.0$ & $+17.8 \pm 13.6$ & $+8.70 \pm 6.94$ \\
Naive + Opus 5 & 2 & $94.5 \pm 2.3$ & $90.3 \pm 2.1$ & $-0.6 \pm 2.9$ & $-0.22 \pm 0.84$ \\
Claude Code + Opus 5 & 4 & $94.1 \pm 1.6$ & $89.4 \pm 3.3$ & $+1.6 \pm 5.6$ & $+0.60 \pm 1.61$ \\
ReasoningBank + Opus 5 & 2.3 & $93.7 \pm 1.8$ & $88.8 \pm 2.1$ & $-1.1 \pm 4.5$ & $-0.50 \pm 1.49$ \\
Reflexion + Opus 5 & 2 & $93.0 \pm 4.5$ & $91.3 \pm 3.9$ & $+3.0 \pm 3.4$ & $+0.92 \pm 0.99$ \\
Human Top-1 & 4 & 91.3 & 78.9 & -- & -- \\
\opusrevision{OpenCode + Opus 5.5} & \opusrevision{2.3} & \opusrevision{$91.1 \pm 4.2$} & \opusrevision{$88.0 \pm 1.7$} & \opusrevision{$-1.7 \pm 2.1$} & \opusrevision{$-0.70 \pm 0.80$} \\
OpenCode + Kimi K3 & 3.3 & $90.1 \pm 2.6$ & $82.0 \pm 10.3$ & $+14.2 \pm 27.0$ & $+5.65 \pm 10.66$ \\
Naive + Kimi K3 & 3.3 & $89.0 \pm 4.9$ & $52.5 \pm 20.3$ & $+21.0 \pm 53.5$ & $+8.03 \pm 14.27$ \\
ReasoningBank + Kimi K3 & 3.3 & $88.7 \pm 2.3$ & $50.3 \pm 16.2$ & $+34.2 \pm 37.1$ & $+12.36 \pm 12.56$ \\
EvoTest + Opus 5 & 2 & $88.5 \pm 1.2$ & $87.1 \pm 0.6$ & $-0.3 \pm 0.5$ & $-0.21 \pm 0.07$ \\
Reflexion + Kimi K3 & 3.7 & $88.4 \pm 3.0$ & $49.9 \pm 29.1$ & $+24.8 \pm 55.8$ & $+8.21 \pm 20.41$ \\
EvoTest + Kimi K3 & 2 & $87.4 \pm 0.9$ & $86.7 \pm 0.3$ & $-0.0 \pm 0.7$ & $-0.01 \pm 0.28$ \\
OpenCode + GPT-5.6-SOL & 2.3 & $86.9 \pm 2.3$ & $76.2 \pm 10.8$ & $+8.0 \pm 34.6$ & $+1.63 \pm 11.32$ \\
AWM + Kimi K3 & 2.3 & $86.5 \pm 0.2$ & $60.4 \pm 17.1$ & $-5.1 \pm 60.1$ & $-0.74 \pm 19.92$ \\
Codex + GPT-5.6-SOL & 3 & $85.6 \pm 8.6$ & $70.5 \pm 22.1$ & $+20.6 \pm 15.3$ & $+3.88 \pm 2.55$ \\
OpenCode + Qwen 3.8 Max & 3 & $85.4 \pm 2.4$ & $66.0 \pm 16.8$ & $+27.7 \pm 35.1$ & $+11.09 \pm 10.90$ \\
OpenCode + Gemini 3.5 Flash & 3.3 & $82.4 \pm 5.0$ & $76.0 \pm 5.6$ & $+14.0 \pm 13.4$ & $+4.58 \pm 4.10$ \\
OpenCode + Gemini 3.7 Flash & 1.3 & $82.0 \pm 4.2$ & $81.8 \pm 3.8$ & $+0.5 \pm 0.8$ & $+0.18 \pm 0.31$ \\
EvoTest + GPT-5 Mini & 1 & $70.9 \pm 26.8$ & $30.1 \pm 5.4$ & $-8.7 \pm 12.5$ & $-6.18 \pm 6.64$ \\
Reflexion + GPT-5 Mini & 2.3 & $53.3 \pm 25.6$ & $18.6 \pm 8.4$ & $+1.4 \pm 9.4$ & $+0.48 \pm 3.81$ \\
ReasoningBank + GPT-5 Mini & 2.3 & $45.3 \pm 9.5$ & $15.6 \pm 6.4$ & $-0.9 \pm 15.9$ & $-1.36 \pm 5.73$ \\
Naive + GPT-5 Mini & 2.3 & $42.4 \pm 1.8$ & $25.0 \pm 8.6$ & $-2.7 \pm 16.7$ & $-3.26 \pm 5.35$ \\
AWM + GPT-5 Mini & 1 & $40.7 \pm 0.2$ & $8.1 \pm 0.0$ & $-20.4 \pm 0.1$ & $-8.14 \pm 0.03$ \\
Memory + GPT-5 Mini & 1 & $40.3 \pm 0.3$ & $13.4 \pm 9.2$ & $-13.5 \pm 11.6$ & $-5.40 \pm 4.62$ \\
OpenCode + DeepSeek V4 Pro & 4.3 & $28.2 \pm 28.4$ & $14.5 \pm 20.5$ & $+17.6 \pm 19.8$ & $+5.13 \pm 6.58$ \\
OpenCode + Sonnet 5 & 1 & $27.5 \pm 47.7$ & $16.1 \pm 27.9$ & $-0.3 \pm 0.6$ & $-1.59 \pm 2.76$ \\
OpenCode + GPT-5 Mini & 1.3 & $9.3 \pm 16.1$ & $7.4 \pm 12.9$ & $+4.6 \pm 8.0$ & $+1.84 \pm 3.19$ \\
\bottomrule
\end{longtable}
\endgroup
\begin{longtable}{@{}p{0.07\linewidth}p{0.87\linewidth}@{}}
\caption{\captionrevision{\revised{Lost and Found: criteria for assessing rule discovery.}}}\label{tab:targets-definition-lost-and-found}\\
\toprule
\textbf{Target} & \textbf{Criterion} \\
\midrule
\endfirsthead
\toprule
\textbf{Target} & \textbf{Criterion (continued)} \\
\midrule
\endhead
T1 & \revised{Three reshuffled cases per episode with four claimant roles: owner, finder, associate, and mimic.} \\
T2 & \revised{Inspect the item privately to learn its concealed marker. Opening it publicly or asking questions too early reveals details that later claimants can repeat.} \\
T3 & \revised{Verify the concealed detail, the claimant's relationship to the item and its purpose, the loss timeline, and the relevant external record. A matching decoy record is not sufficient.} \\
T4 & \revised{Resource budgeting: 24 turns, eight privacy tokens, three isolation tokens, and three record checks across all cases.} \\
T5 & \revised{1,000-point office protocol: inspect each item privately, verify owner on all independent channels, check the matching record, and return or store without leaking decisive evidence.} \\
\bottomrule
\end{longtable}
\begin{center}
\begin{minipage}{\linewidth}
\centering
\captionof{table}{\captionrevision{\revised{Lost and Found: rule-discovery evidence. S: complete evidence in at least one trial; P: partial evidence only; --: insufficient evidence. Evidence may include episodes beyond the scoring window.}}}
\label{tab:rq1-targets-game-lost-and-found}
{
\appendixtable{%
\begin{tabular}{@{}l*{5}{c}@{}}
\toprule
\textbf{Configuration} & T1 & T2 & T3 & T4 & T5 \\
\midrule
\multicolumn{6}{@{}l}{\textbf{backbone comparison}} \\
Claude Opus~5 & S & S & S & S & S \\
Qwen 3.8 Max & S & S & S & S & -- \\
Kimi~K3 & S & S & S & S & -- \\
Gemini~3.5 Flash & S & S & S & S & -- \\
Gemini~3.7 Flash & S & S & S & S & -- \\
GPT-5.6-SOL & S & S & S & S & -- \\
DeepSeek V4 Pro & S & S & S & -- & -- \\
Claude Sonnet~5 & S & S & S & S & -- \\
GPT-5 Mini & S & S & S & S & -- \\
\midrule
\multicolumn{6}{@{}l}{\textbf{agent harness comparison}} \\
Claude Code + Opus~5 & S & S & S & S & S \\
Codex + GPT-5.6-SOL & S & S & S & S & S \\
\bottomrule
\end{tabular}
}
}

\end{minipage}
\end{center}

\subsection{\revised{Mola Tea}}
\label{app:game-mola_tea}

\revised{\textbf{Learning task.} Infer the adoption chain and when each promotion becomes effective. Re-shuffled game, scored seed 1, first 10 episodes, reference score $C=1000$.}
\begin{center}
\includegraphics[width=0.95\linewidth]{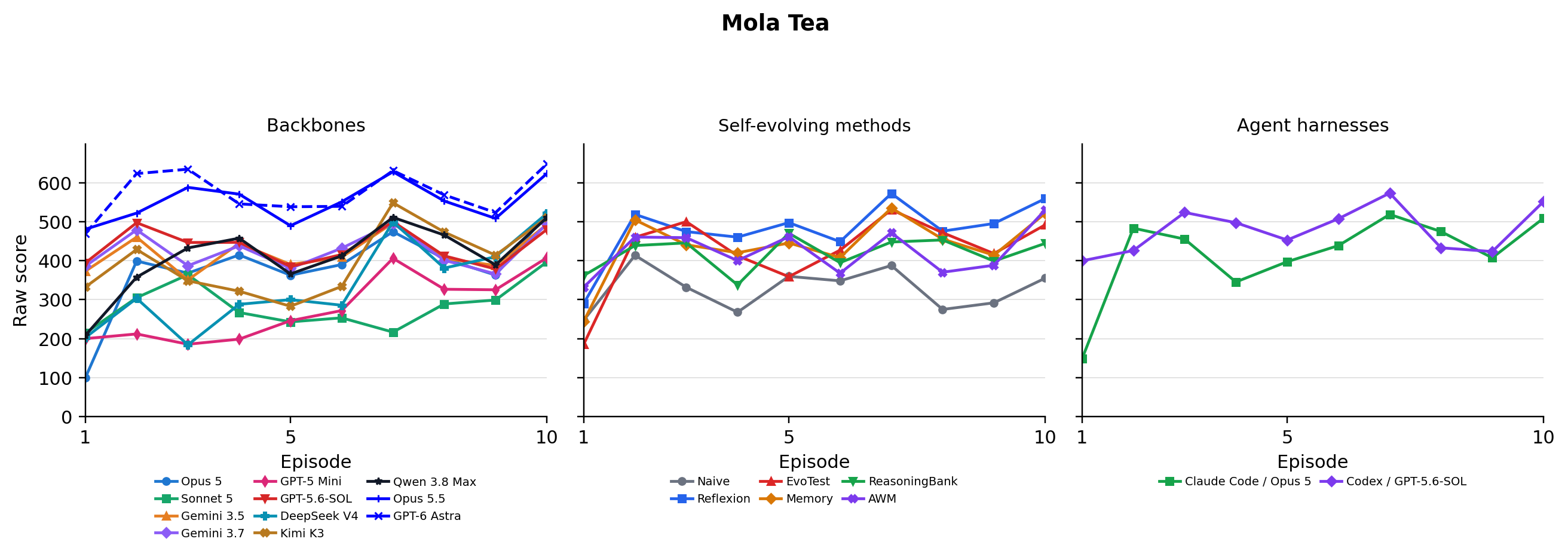}
\captionof{figure}{\revised{Mola Tea: raw-score learning curves within the scoring window.}}
\label{fig:game-curve-mola_tea}

\end{center}
\Needspace{10\baselineskip}
% Combined score table: normalized points and trial SD; human/First retained.
\begingroup
\setlength{\tabcolsep}{3pt}
\begin{longtable}{@{}lrrrrr@{}}
\caption{\captionrevision{Mola Tea: normalized performance and learning, with trial SD. Rows follow descending Max. First denotes the average episode at which a trial first reaches its highest score.}}
\label{tab:game-scores-mola_tea}\label{tab:trial-variability-mola-tea}\\
\toprule
Configuration & First & Max $\uparrow$ & Mean $\uparrow$ & LG $\uparrow$ & LS $\uparrow$ \\
\midrule
\endfirsthead
\toprule
Configuration & First & Max $\uparrow$ & Mean $\uparrow$ & LG $\uparrow$ & LS $\uparrow$ \\
\midrule
\endhead
Reflexion + Opus 5 & 10 & $79.0 \pm 13.6$ & $63.8 \pm 5.8$ & $+18.0 \pm 13.9$ & $+2.96 \pm 1.80$ \\
AWM + Kimi K3 & 8.7 & $74.8 \pm 18.1$ & $46.1 \pm 14.5$ & $+5.6 \pm 3.3$ & $+0.98 \pm 0.83$ \\
Human Top-1 & 7 & 68.4 & 40.7 & -- & -- \\
ReasoningBank + Opus 5 & 8.7 & $66.1 \pm 4.8$ & $51.7 \pm 4.1$ & $+5.6 \pm 8.1$ & $+1.36 \pm 1.01$ \\
\astrarevision{OpenCode + GPT-6 Astra} & \astrarevision{6.3} & \astrarevision{$66.1 \pm 2.9$} & \astrarevision{$57.2 \pm 2.7$} & \astrarevision{$+0.4 \pm 2.4$} & \astrarevision{$+0.51 \pm 0.26$} \\
EvoTest + Opus 5 & 3 & $65.8 \pm 3.6$ & $52.9 \pm 4.5$ & $+1.6 \pm 6.3$ & $+0.92 \pm 0.84$ \\
\opusrevision{OpenCode + Opus 5.5} & \opusrevision{6.7} & \opusrevision{$65.6 \pm 3.8$} & \opusrevision{$55.1 \pm 4.8$} & \opusrevision{$+3.1 \pm 6.1$} & \opusrevision{$+0.76 \pm 0.72$} \\
AWM + Opus 5 & 5.3 & $64.3 \pm 4.3$ & $52.4 \pm 1.8$ & $+2.6 \pm 3.8$ & $+0.88 \pm 0.11$ \\
Memory + Opus 5 & 7 & $62.6 \pm 9.6$ & $52.6 \pm 6.4$ & $+9.4 \pm 2.8$ & $+1.75 \pm 0.58$ \\
Reflexion + Kimi K3 & 3.7 & $60.7 \pm 8.5$ & $44.1 \pm 7.2$ & $+1.8 \pm 11.9$ & $+0.63 \pm 1.56$ \\
ReasoningBank + GPT-5 Mini & 3.7 & $60.5 \pm 16.5$ & $38.9 \pm 3.0$ & $+11.2 \pm 15.2$ & $+1.64 \pm 2.07$ \\
ReasoningBank + Kimi K3 & 4.7 & $59.6 \pm 12.2$ & $35.2 \pm 6.3$ & $-11.8 \pm 10.6$ & $-1.62 \pm 1.01$ \\
Codex + GPT-5.6-SOL & 8 & $57.4 \pm 11.7$ & $47.9 \pm 6.3$ & $+2.0 \pm 3.7$ & $+0.71 \pm 0.47$ \\
Memory + GPT-5 Mini & 5 & $56.5 \pm 9.1$ & $35.2 \pm 5.7$ & $+1.6 \pm 7.1$ & $+0.76 \pm 0.87$ \\
OpenCode + Kimi K3 & 8 & $55.1 \pm 8.0$ & $39.9 \pm 6.3$ & $+9.6 \pm 6.3$ & $+1.73 \pm 0.92$ \\
Memory + Kimi K3 & 3.7 & $54.2 \pm 7.4$ & $43.8 \pm 3.7$ & $+9.0 \pm 10.2$ & $+1.56 \pm 1.50$ \\
Naive + Opus 5 & 2 & $54.1 \pm 6.5$ & $37.6 \pm 2.5$ & $-3.1 \pm 2.1$ & $-0.22 \pm 0.24$ \\
EvoTest + Kimi K3 & 4 & $54.0 \pm 3.6$ & $39.1 \pm 2.6$ & $+6.6 \pm 6.4$ & $+1.57 \pm 0.69$ \\
Naive + GPT-5 Mini & 3 & $52.7 \pm 12.6$ & $29.0 \pm 4.3$ & $-6.7 \pm 13.2$ & $-0.15 \pm 1.98$ \\
OpenCode + DeepSeek V4 Pro & 8 & $52.7 \pm 4.2$ & $33.8 \pm 6.9$ & $+20.7 \pm 1.4$ & $+3.16 \pm 0.47$ \\
Claude Code + Opus 5 & 8 & $52.0 \pm 0.0$ & $41.8 \pm 0.4$ & $+10.1 \pm 5.1$ & $+2.04 \pm 0.76$ \\
OpenCode + Opus 5 & 9 & $52.0 \pm 0.0$ & $38.0 \pm 2.0$ & $+14.0 \pm 7.5$ & $+2.38 \pm 1.06$ \\
OpenCode + Qwen 3.8 Max & 8 & $51.8 \pm 0.3$ & $41.1 \pm 2.5$ & $+12.2 \pm 3.6$ & $+2.01 \pm 0.47$ \\
OpenCode + GPT-5.6-SOL & 5.3 & $50.1 \pm 1.6$ & $43.6 \pm 1.6$ & $-2.2 \pm 3.7$ & $-0.03 \pm 0.57$ \\
OpenCode + Gemini 3.5 Flash & 8 & $50.0 \pm 1.2$ & $42.1 \pm 0.5$ & $+3.4 \pm 5.2$ & $+0.62 \pm 0.50$ \\
OpenCode + Gemini 3.7 Flash & 7 & $49.6 \pm 0.0$ & $42.6 \pm 0.4$ & $+0.3 \pm 1.2$ & $+0.29 \pm 0.11$ \\
Reflexion + GPT-5 Mini & 7 & $49.1 \pm 2.0$ & $35.9 \pm 1.3$ & $+4.9 \pm 7.1$ & $+1.04 \pm 0.88$ \\
OpenCode + Sonnet 5 & 7.3 & $48.7 \pm 0.1$ & $28.5 \pm 0.9$ & $+3.4 \pm 14.3$ & $+0.66 \pm 1.78$ \\
EvoTest + GPT-5 Mini & 6.7 & $48.2 \pm 2.6$ & $35.7 \pm 4.1$ & $+15.4 \pm 10.8$ & $+2.50 \pm 1.21$ \\
AWM + GPT-5 Mini & 6 & $47.1 \pm 7.3$ & $28.8 \pm 4.7$ & $-4.5 \pm 11.5$ & $-0.13 \pm 1.83$ \\
Naive + Kimi K3 & 6.7 & $44.5 \pm 7.3$ & $31.8 \pm 2.9$ & $+2.7 \pm 9.7$ & $+0.70 \pm 1.80$ \\
OpenCode + GPT-5 Mini & 8 & $41.0 \pm 14.2$ & $27.8 \pm 6.3$ & $+15.4 \pm 10.3$ & $+2.43 \pm 1.67$ \\
\bottomrule
\end{longtable}
\endgroup
\begin{longtable}{@{}p{0.07\linewidth}p{0.87\linewidth}@{}}
\caption{\captionrevision{\revised{Mola Tea: criteria for assessing rule discovery.}}}\label{tab:targets-definition-mola-tea}\\
\toprule
\textbf{Target} & \textbf{Criterion} \\
\midrule
\endfirsthead
\toprule
\textbf{Target} & \textbf{Criterion (continued)} \\
\midrule
\endhead
T1 & \revised{21-turn, seven-day shop economy with cash, unit cost, launch stock, stockouts, expired inventory, and regular-customer churn.} \\
T2 & \revised{Recipe rule: only products at distance 1 from the classic drink are viable, while far variants can sell briefly but fail the adoption chain.} \\
T3 & \revised{Customer-role inference from clues: Explorer, Echo, Verifier, Broadcaster, and Anchor/regulars have distinct responses to sample, feature, discount, gift, and bundle.} \\
T4 & \revised{Stage transition chain LAUNCHED\(\to\)SEEDED\(\to\)REPEATED\(\to\)VERIFIED\(\to\)TRENDING, with clean intervals and short verification/diffusion deadlines.} \\
T5 & \revised{Sampling Explorers starts adoption. Feature promotion works after a natural repeat purchase. A Broadcaster gift works only after verification. Bundles pay off only after the product is trending. Repeated discounts teach customers to expect low prices.} \\
T6 & \revised{Inventory/novelty control: restock before return and bundle windows, use hold/restore-classic to reduce novelty pressure, and avoid leftover-expiry costs.} \\
T7 & \revised{708-point schedule: develop a distance-1 card, launch, sample Explorers, large-restock through the natural repeat, feature within deadline, gift cheapest merch to Broadcasters, then bundle and manage stock/novelty to close the week.} \\
\bottomrule
\end{longtable}
\begin{center}
\begin{minipage}{\linewidth}
\centering
\captionof{table}{\captionrevision{\revised{Mola Tea: rule-discovery evidence. S: complete evidence in at least one trial; P: partial evidence only; --: insufficient evidence. Evidence may include episodes beyond the scoring window.}}}
\label{tab:rq1-targets-game-mola-tea}
{
\appendixtable{%
\begin{tabular}{@{}l*{7}{c}@{}}
\toprule
\textbf{Configuration} & T1 & T2 & T3 & T4 & T5 & T6 & T7 \\
\midrule
\multicolumn{8}{@{}l}{\textbf{backbone comparison}} \\
Claude Opus~5 & S & S & S & S & S & S & -- \\
Qwen 3.8 Max & S & S & S & S & S & S & -- \\
Kimi~K3 & S & S & S & S & S & S & S \\
Gemini~3.5 Flash & S & S & S & S & S & S & -- \\
Gemini~3.7 Flash & S & S & -- & S & S & S & -- \\
GPT-5.6-SOL & S & S & S & S & S & S & -- \\
DeepSeek V4 Pro & S & S & -- & S & S & S & -- \\
Claude Sonnet~5 & S & S & S & S & S & S & -- \\
GPT-5 Mini & S & -- & -- & -- & S & S & -- \\
\midrule
\multicolumn{8}{@{}l}{\textbf{agent harness comparison}} \\
Claude Code + Opus~5 & S & S & S & S & S & S & -- \\
Codex + GPT-5.6-SOL & S & S & S & S & S & S & S \\
\bottomrule
\end{tabular}
}
}

\end{minipage}
\end{center}

\subsection{\revised{Plague}}
\label{app:game-plague}

\revised{\textbf{Learning task.} Locate asymptomatic carriers from spread patterns and isolate them early. Re-shuffled game, scored seed 2, first 5 episodes, reference score $C=100$.}
\begin{center}
\includegraphics[width=0.95\linewidth]{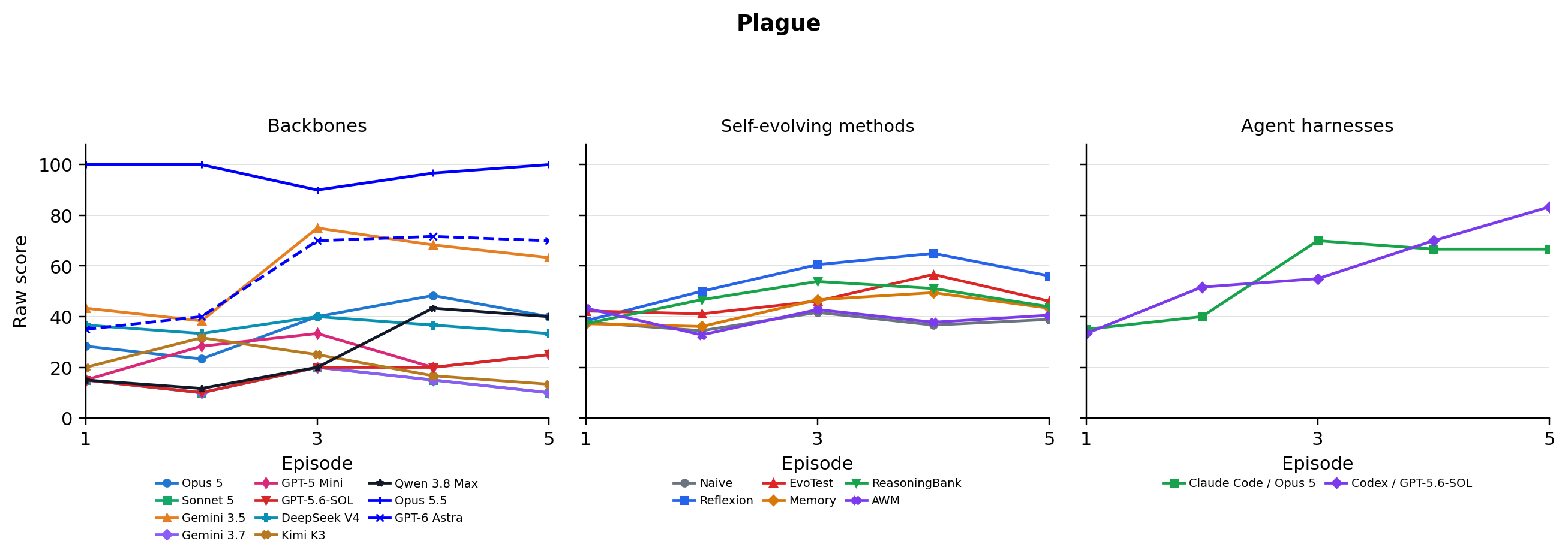}
\captionof{figure}{\revised{Plague: raw-score learning curves within the scoring window.}}
\label{fig:game-curve-plague}

\end{center}
\Needspace{10\baselineskip}
% Combined score table: normalized points and trial SD; human/First retained.
\begingroup
\setlength{\tabcolsep}{3pt}
\begin{longtable}{@{}lrrrrr@{}}
\caption{\captionrevision{Plague: normalized performance and learning, with trial SD. Rows follow descending Max. First denotes the average episode at which a trial first reaches its highest score.}}
\label{tab:game-scores-plague}\label{tab:trial-variability-plague}\\
\toprule
Configuration & First & Max $\uparrow$ & Mean $\uparrow$ & LG $\uparrow$ & LS $\uparrow$ \\
\midrule
\endfirsthead
\toprule
Configuration & First & Max $\uparrow$ & Mean $\uparrow$ & LG $\uparrow$ & LS $\uparrow$ \\
\midrule
\endhead
Human Top-1 & 2 & 100.0 & 58.0 & -- & -- \\
Reflexion + GPT-5 Mini & 3 & $100.0 \pm 0.0$ & $65.0 \pm 12.0$ & $+20.0 \pm 43.4$ & $+6.50 \pm 11.76$ \\
\opusrevision{OpenCode + Opus 5.5} & \opusrevision{1} & \opusrevision{$100.0 \pm 0.0$} & \opusrevision{$97.3 \pm 3.1$} & \opusrevision{$-1.7 \pm 2.9$} & \opusrevision{$-0.33 \pm 0.58$} \\
Codex + GPT-5.6-SOL & 4.7 & $90.0 \pm 17.3$ & $58.7 \pm 22.2$ & $+34.2 \pm 11.8$ & $+11.83 \pm 2.47$ \\
Claude Code + Opus 5 & 4 & $76.7 \pm 10.4$ & $55.7 \pm 15.7$ & $+29.2 \pm 16.6$ & $+9.00 \pm 5.29$ \\
OpenCode + Gemini 3.5 Flash & 2.3 & $75.0 \pm 35.0$ & $57.7 \pm 40.5$ & $+25.0 \pm 39.1$ & $+7.00 \pm 11.72$ \\
EvoTest + Opus 5 & 4 & $75.0 \pm 21.8$ & $50.0 \pm 5.3$ & $+19.2 \pm 22.4$ & $+6.33 \pm 7.94$ \\
ReasoningBank + GPT-5 Mini & 2.3 & $73.3 \pm 30.6$ & $59.7 \pm 26.1$ & $+13.3 \pm 17.7$ & $+4.83 \pm 7.42$ \\
\astrarevision{OpenCode + GPT-6 Astra} & \astrarevision{3} & \astrarevision{$73.3 \pm 46.2$} & \astrarevision{$57.3 \pm 40.8$} & \astrarevision{$+33.3 \pm 47.3$} & \astrarevision{$+10.17 \pm 13.99$} \\
Reflexion + Opus 5 & 4.7 & $71.7 \pm 2.9$ & $48.7 \pm 1.2$ & $+17.5 \pm 10.9$ & $+5.00 \pm 5.20$ \\
AWM + GPT-5 Mini & 4.3 & $63.3 \pm 32.1$ & $47.0 \pm 18.5$ & $+5.0 \pm 2.5$ & $+1.67 \pm 1.44$ \\
Memory + Opus 5 & 4 & $61.7 \pm 12.6$ & $47.3 \pm 7.6$ & $+15.8 \pm 7.6$ & $+4.17 \pm 1.76$ \\
EvoTest + GPT-5 Mini & 2 & $60.0 \pm 0.0$ & $45.0 \pm 7.5$ & $-11.7 \pm 5.2$ & $-5.33 \pm 0.76$ \\
Reflexion + Kimi K3 & 3 & $58.3 \pm 2.9$ & $48.3 \pm 5.0$ & $+11.7 \pm 9.5$ & $+3.67 \pm 3.06$ \\
Memory + Kimi K3 & 3 & $58.3 \pm 2.9$ & $46.3 \pm 2.1$ & $+11.7 \pm 12.3$ & $+3.50 \pm 2.65$ \\
OpenCode + Opus 5 & 4 & $58.3 \pm 40.1$ & $36.0 \pm 31.4$ & $+18.3 \pm 17.6$ & $+4.83 \pm 6.11$ \\
ReasoningBank + Opus 5 & 2.3 & $56.7 \pm 7.6$ & $42.7 \pm 4.2$ & $-2.5 \pm 10.0$ & $-0.84 \pm 2.52$ \\
EvoTest + Kimi K3 & 3.7 & $56.7 \pm 2.9$ & $44.3 \pm 1.5$ & $+21.7 \pm 3.8$ & $+6.00 \pm 1.32$ \\
ReasoningBank + Kimi K3 & 4 & $56.7 \pm 7.6$ & $37.3 \pm 2.3$ & $+5.8 \pm 13.8$ & $+1.33 \pm 3.06$ \\
Naive + GPT-5 Mini & 3.7 & $50.0 \pm 10.0$ & $37.0 \pm 3.0$ & $+5.8 \pm 8.0$ & $+1.67 \pm 3.33$ \\
OpenCode + Qwen 3.8 Max & 3.3 & $46.7 \pm 46.2$ & $26.0 \pm 20.8$ & $+28.3 \pm 49.1$ & $+8.17 \pm 15.01$ \\
Naive + Kimi K3 & 4 & $46.7 \pm 2.9$ & $38.0 \pm 1.7$ & $+0.0 \pm 10.9$ & $+0.50 \pm 3.12$ \\
AWM + Kimi K3 & 2.3 & $45.0 \pm 0.0$ & $35.3 \pm 0.6$ & $-0.0 \pm 5.0$ & $-0.84 \pm 1.53$ \\
Naive + Opus 5 & 3 & $45.0 \pm 0.0$ & $38.7 \pm 1.2$ & $-0.8 \pm 5.2$ & $-0.83 \pm 1.15$ \\
AWM + Opus 5 & 1 & $40.0 \pm 0.0$ & $36.0 \pm 1.0$ & $-1.7 \pm 1.4$ & $-1.00 \pm 0.50$ \\
OpenCode + DeepSeek V4 Pro & 2.3 & $40.0 \pm 34.6$ & $36.0 \pm 38.1$ & $+0.0 \pm 0.0$ & $-0.33 \pm 0.29$ \\
Memory + GPT-5 Mini & 3 & $40.0 \pm 0.0$ & $34.0 \pm 0.0$ & $+1.7 \pm 2.9$ & $+0.00 \pm 0.87$ \\
OpenCode + Kimi K3 & 2.7 & $36.7 \pm 17.6$ & $21.3 \pm 6.7$ & $-10.8 \pm 11.3$ & $-2.83 \pm 2.25$ \\
OpenCode + GPT-5 Mini & 2.7 & $35.0 \pm 21.8$ & $24.3 \pm 15.4$ & $+0.8 \pm 8.8$ & $+1.17 \pm 4.25$ \\
OpenCode + GPT-5.6-SOL & 3.7 & $31.7 \pm 20.2$ & $18.0 \pm 6.9$ & $+10.0 \pm 17.3$ & $+3.00 \pm 6.06$ \\
OpenCode + Sonnet 5 & 3 & $20.0 \pm 0.0$ & $14.0 \pm 0.0$ & $+0.0 \pm 0.0$ & $-0.50 \pm 0.00$ \\
OpenCode + Gemini 3.7 Flash & 3 & $20.0 \pm 0.0$ & $14.0 \pm 0.0$ & $+0.0 \pm 0.0$ & $-0.50 \pm 0.00$ \\
\bottomrule
\end{longtable}
\endgroup
\begin{longtable}{@{}p{0.07\linewidth}p{0.87\linewidth}@{}}
\caption{\captionrevision{\revised{Plague: criteria for assessing rule discovery.}}}\label{tab:targets-definition-plague}\\
\toprule
\textbf{Target} & \textbf{Criterion} \\
\midrule
\endfirsthead
\toprule
\textbf{Target} & \textbf{Criterion (continued)} \\
\midrule
\endhead
T1 & \revised{Line and action economy: 20 people, 14 nights, four AP per day, three quarantine cells, severity 1--3, and death after severity \(>3\).} \\
T2 & \revised{Hidden-carrier dynamics: two asymptomatic carriers are the only spreaders, reshuffled into left/right halves, with radius \(1+\lfloor age/2\rfloor\), while symptomatic non-carriers do not infect anyone.} \\
T3 & \revised{Intervention semantics: quarantine blocks both catching and spreading, whereas treatment only reduces current severity and gives one-night immunity after cure.} \\
T4 & \revised{Carrier diagnosis from spatial evidence, especially separated sick pairs around a healthy center and disappearance of new cases after quarantining that center.} \\
T5 & \revised{AP triage: quarantine likely carriers early, treat only patients near death, and release/replace false quarantines without exceeding three cells.} \\
T6 & \revised{For a 100-point containment policy, identify both carriers by day 2--3 and quarantine them before their spread radius grows. Time treatments so that no other villager reaches the death threshold.} \\
\bottomrule
\end{longtable}
\begin{center}
\begin{minipage}{\linewidth}
\centering
\captionof{table}{\captionrevision{\revised{Plague: rule-discovery evidence. S: complete evidence in at least one trial; P: partial evidence only; --: insufficient evidence. Evidence may include episodes beyond the scoring window.}}}
\label{tab:rq1-targets-game-plague}
{
\appendixtable{%
\begin{tabular}{@{}l*{6}{c}@{}}
\toprule
\textbf{Configuration} & T1 & T2 & T3 & T4 & T5 & T6 \\
\midrule
\multicolumn{7}{@{}l}{\textbf{backbone comparison}} \\
Claude Opus~5 & S & S & S & S & S & S \\
Qwen 3.8 Max & S & S & S & S & S & S \\
Kimi~K3 & S & S & S & S & S & -- \\
Gemini~3.5 Flash & S & S & S & S & S & S \\
Gemini~3.7 Flash & S & S & S & S & S & S \\
GPT-5.6-SOL & S & S & S & S & -- & -- \\
DeepSeek V4 Pro & S & S & S & S & S & S \\
Claude Sonnet~5 & S & S & S & -- & S & -- \\
GPT-5 Mini & S & S & S & -- & S & -- \\
\midrule
\multicolumn{7}{@{}l}{\textbf{agent harness comparison}} \\
Claude Code + Opus~5 & S & S & S & S & S & S \\
Codex + GPT-5.6-SOL & S & S & S & S & S & S \\
\bottomrule
\end{tabular}
}
}

\end{minipage}
\end{center}

\subsection{\revised{Patch Reality}}
\label{app:game-patch_reality_moderate}

\revised{\textbf{Learning task.} Remember diagnosis--repair mappings, decoys, and prerequisites. Fixed game, scored seed 1, first 5 episodes, reference score $C=960$.}
\begin{center}
\includegraphics[width=0.95\linewidth]{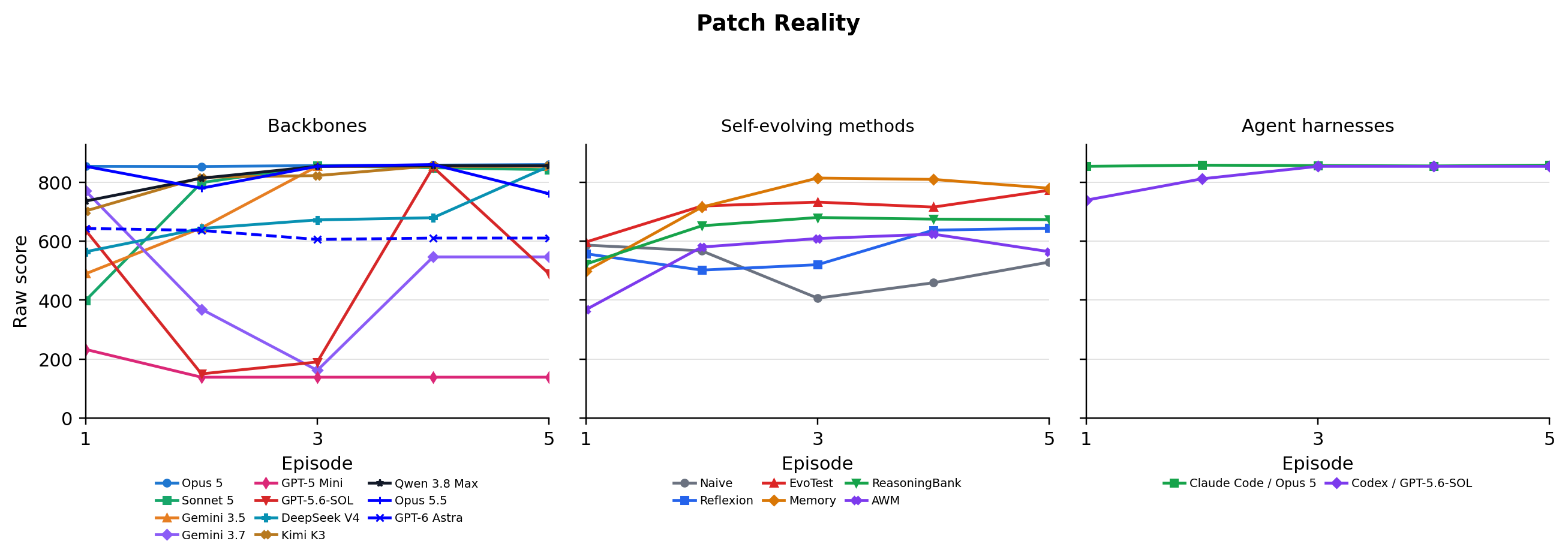}
\captionof{figure}{\revised{Patch Reality: raw-score learning curves within the scoring window.}}
\label{fig:game-curve-patch_reality_moderate}

\end{center}
\Needspace{10\baselineskip}
% Combined score table: normalized points and trial SD; human/First retained.
\begingroup
\setlength{\tabcolsep}{3pt}
\begin{longtable}{@{}lrrrrr@{}}
\caption{\captionrevision{Patch Reality: normalized performance and learning, with trial SD. Rows follow descending Max. First denotes the average episode at which a trial first reaches its highest score.}}
\label{tab:game-scores-patch_reality_moderate}\label{tab:trial-variability-patch-reality-moderate}\\
\toprule
Configuration & First & Max $\uparrow$ & Mean $\uparrow$ & LG $\uparrow$ & LS $\uparrow$ \\
\midrule
\endfirsthead
\toprule
Configuration & First & Max $\uparrow$ & Mean $\uparrow$ & LG $\uparrow$ & LS $\uparrow$ \\
\midrule
\endhead
EvoTest + Opus 5 & 3.7 & $89.7 \pm 0.4$ & $86.9 \pm 2.6$ & $+2.3 \pm 3.6$ & $+1.30 \pm 1.19$ \\
ReasoningBank + Opus 5 & 4.3 & $89.7 \pm 0.4$ & $85.2 \pm 2.7$ & $+6.1 \pm 0.2$ & $+2.44 \pm 0.10$ \\
Claude Code + Opus 5 & 2.7 & $89.6 \pm 0.0$ & $89.2 \pm 0.1$ & $+0.1 \pm 0.5$ & $+0.06 \pm 0.11$ \\
OpenCode + Opus 5 & 2.7 & $89.6 \pm 0.0$ & $89.2 \pm 0.1$ & $+0.6 \pm 0.3$ & $+0.17 \pm 0.05$ \\
Human Top-1 & 5 & 89.6 & 59.2 & -- & -- \\
\opusrevision{OpenCode + Opus 5.5} & \opusrevision{4} & \opusrevision{$89.6 \pm 0.0$} & \opusrevision{$85.6 \pm 3.3$} & \opusrevision{$-0.7 \pm 10.3$} & \opusrevision{$-1.10 \pm 3.81$} \\
Reflexion + Opus 5 & 3.7 & $89.5 \pm 0.4$ & $84.5 \pm 6.1$ & $+0.2 \pm 5.7$ & $+0.09 \pm 2.32$ \\
EvoTest + Kimi K3 & 3 & $89.3 \pm 0.0$ & $84.1 \pm 2.4$ & $+12.2 \pm 6.1$ & $+4.15 \pm 1.84$ \\
Memory + Opus 5 & 3 & $89.3 \pm 0.0$ & $82.5 \pm 4.9$ & $+4.8 \pm 8.3$ & $+1.94 \pm 3.33$ \\
Memory + Kimi K3 & 3 & $89.3 \pm 0.0$ & $71.5 \pm 9.2$ & $+43.8 \pm 23.1$ & $+14.83 \pm 6.93$ \\
OpenCode + Qwen 3.8 Max & 3 & $89.2 \pm 0.4$ & $85.8 \pm 1.4$ & $+8.4 \pm 3.2$ & $+2.93 \pm 0.64$ \\
OpenCode + Kimi K3 & 3 & $89.2 \pm 0.4$ & $84.5 \pm 2.7$ & $+10.0 \pm 6.8$ & $+3.60 \pm 2.07$ \\
OpenCode + Sonnet 5 & 3 & $89.2 \pm 0.4$ & $78.0 \pm 2.3$ & $+25.7 \pm 4.3$ & $+9.76 \pm 1.78$ \\
Codex + GPT-5.6-SOL & 2 & $89.0 \pm 0.0$ & $85.7 \pm 2.7$ & $+8.2 \pm 6.6$ & $+2.83 \pm 2.45$ \\
OpenCode + Gemini 3.5 Flash & 2.7 & $89.0 \pm 0.0$ & $77.0 \pm 12.6$ & $+29.8 \pm 31.5$ & $+9.76 \pm 9.09$ \\
OpenCode + DeepSeek V4 Pro & 2.7 & $89.0 \pm 0.0$ & $71.1 \pm 26.7$ & $+17.0 \pm 19.7$ & $+6.41 \pm 7.26$ \\
OpenCode + GPT-5.6-SOL & 1.3 & $88.7 \pm 0.5$ & $48.3 \pm 4.1$ & $+28.9 \pm 44.9$ & $+4.24 \pm 11.63$ \\
ReasoningBank + Kimi K3 & 3.7 & $85.4 \pm 6.7$ & $72.3 \pm 10.0$ & $+5.6 \pm 11.0$ & $+3.00 \pm 3.36$ \\
Naive + Opus 5 & 3.3 & $85.0 \pm 7.0$ & $67.4 \pm 2.5$ & $+9.2 \pm 13.3$ & $+3.59 \pm 3.35$ \\
OpenCode + Gemini 3.7 Flash & 1.3 & $84.1 \pm 8.5$ & $49.9 \pm 38.0$ & $-2.4 \pm 30.3$ & $-2.81 \pm 10.90$ \\
Reflexion + Kimi K3 & 4.7 & $80.9 \pm 6.1$ & $38.5 \pm 18.4$ & $+31.9 \pm 7.4$ & $+8.33 \pm 4.86$ \\
AWM + Opus 5 & 3.3 & $78.1 \pm 10.9$ & $67.1 \pm 12.5$ & $+6.7 \pm 12.0$ & $+1.55 \pm 3.63$ \\
AWM + Kimi K3 & 3 & $77.7 \pm 0.0$ & $62.5 \pm 3.1$ & $+37.3 \pm 7.7$ & $+14.52 \pm 2.40$ \\
Naive + Kimi K3 & 2.7 & $77.5 \pm 0.2$ & $43.1 \pm 6.0$ & $-20.5 \pm 9.6$ & $-5.14 \pm 1.82$ \\
Memory + GPT-5 Mini & 2.7 & $77.1 \pm 0.7$ & $72.2 \pm 3.1$ & $+10.1 \pm 7.6$ & $+3.76 \pm 2.30$ \\
\astrarevision{OpenCode + GPT-6 Astra} & \astrarevision{2.7} & \astrarevision{$67.4 \pm 38.4$} & \astrarevision{$64.7 \pm 42.0$} & \astrarevision{$-3.1 \pm 7.3$} & \astrarevision{$-0.95 \pm 2.15$} \\
Reflexion + GPT-5 Mini & 2.3 & $65.7 \pm 0.2$ & $55.9 \pm 5.1$ & $+2.6 \pm 10.2$ & $+1.23 \pm 4.01$ \\
Naive + GPT-5 Mini & 1.3 & $62.7 \pm 2.0$ & $48.9 \pm 2.8$ & $-14.6 \pm 7.7$ & $-5.44 \pm 0.86$ \\
EvoTest + GPT-5 Mini & 5 & $62.6 \pm 8.1$ & $50.1 \pm 3.9$ & $+12.4 \pm 1.7$ & $+5.42 \pm 1.98$ \\
ReasoningBank + GPT-5 Mini & 3.3 & $59.1 \pm 6.7$ & $42.7 \pm 5.4$ & $+15.4 \pm 22.2$ & $+4.67 \pm 7.20$ \\
AWM + GPT-5 Mini & 2 & $51.8 \pm 9.1$ & $42.0 \pm 13.6$ & $-6.5 \pm 7.7$ & $-2.42 \pm 1.77$ \\
OpenCode + GPT-5 Mini & 1.3 & $28.1 \pm 11.2$ & $16.4 \pm 7.1$ & $-4.9 \pm 13.8$ & $-1.97 \pm 5.51$ \\
\bottomrule
\end{longtable}
\endgroup
\begin{longtable}{@{}p{0.07\linewidth}p{0.87\linewidth}@{}}
\caption{\captionrevision{\revised{Patch Reality: criteria for assessing rule discovery.}}}\label{tab:targets-definition-patch-reality}\\
\toprule
\textbf{Target} & \textbf{Criterion} \\
\midrule
\endfirsthead
\toprule
\textbf{Target} & \textbf{Criterion (continued)} \\
\midrule
\endhead
T1 & \revised{Per-level action grammar: observe evidence, optionally audit risky records, diagnose source/type/strategy, then patch the root object.} \\
T2 & \revised{Levels 1--3 mappings: coin/drain uses environment-cover pavement friction, vibrating cup requires auditing residual patch then expiring it, and flickering key requires object physics repair by solidifying the key.} \\
T3 & \revised{Levels 4--5 mappings: hospital ledger is an identity-record trap solved by binding the wristband, while premature echo is a time-order conflict solved by aligning the acoustic log.} \\
T4 & \revised{Level 6 ordered record repair: reconcile the ledger before halting archive reindexing.} \\
T5 & \revised{Level 7 public-record exposure solved by camera desync.} \\
T6 & \revised{Level 8 identity-anchor pollution solved by reclaiming the name anchor.} \\
T7 & \revised{Audit discipline: audit only residual patch, hospital ledger, acoustic log, security log, broadcast, and agent offer when they gate evidence, because extra/decoy information costs score.} \\
T8 & \revised{860 campaign: clear all eight levels with the exact diagnosis triples and root patches, avoid reward traps and nonessential audits, and preserve focus through the multi-level sequence.} \\
\bottomrule
\end{longtable}
\begin{center}
\begin{minipage}{\linewidth}
\centering
\captionof{table}{\captionrevision{\revised{Patch Reality: rule-discovery evidence. S: complete evidence in at least one trial; P: partial evidence only; --: insufficient evidence. Evidence may include episodes beyond the scoring window.}}}
\label{tab:rq1-targets-game-patch-reality}
{
\appendixtable{%
\begin{tabular}{@{}l*{8}{c}@{}}
\toprule
\textbf{Configuration} & T1 & T2 & T3 & T4 & T5 & T6 & T7 & T8 \\
\midrule
\multicolumn{9}{@{}l}{\textbf{backbone comparison}} \\
Claude Opus~5 & S & S & S & S & S & S & S & S \\
Qwen 3.8 Max & S & S & S & S & S & S & S & S \\
Kimi~K3 & S & S & S & S & S & S & S & S \\
Gemini~3.5 Flash & S & S & S & S & S & S & S & -- \\
Gemini~3.7 Flash & S & S & S & S & S & S & S & -- \\
GPT-5.6-SOL & S & S & S & S & S & S & S & -- \\
DeepSeek V4 Pro & S & S & S & S & S & S & S & -- \\
Claude Sonnet~5 & S & S & S & S & S & S & S & S \\
GPT-5 Mini & S & S & S & -- & -- & -- & S & -- \\
\midrule
\multicolumn{9}{@{}l}{\textbf{agent harness comparison}} \\
Claude Code + Opus~5 & S & S & S & S & S & S & S & S \\
Codex + GPT-5.6-SOL & S & S & S & S & S & S & S & -- \\
\bottomrule
\end{tabular}
}
}

\end{minipage}
\end{center}

\subsection{\revised{Poisoner}}
\label{app:game-poisoner}

\revised{\textbf{Learning task.} Infer taste constraints from accepted and rejected dishes and apply them to new menus. Re-shuffled game, scored seed 2, first 5 episodes, reference score $C=100$.}
\begin{center}
\includegraphics[width=0.95\linewidth]{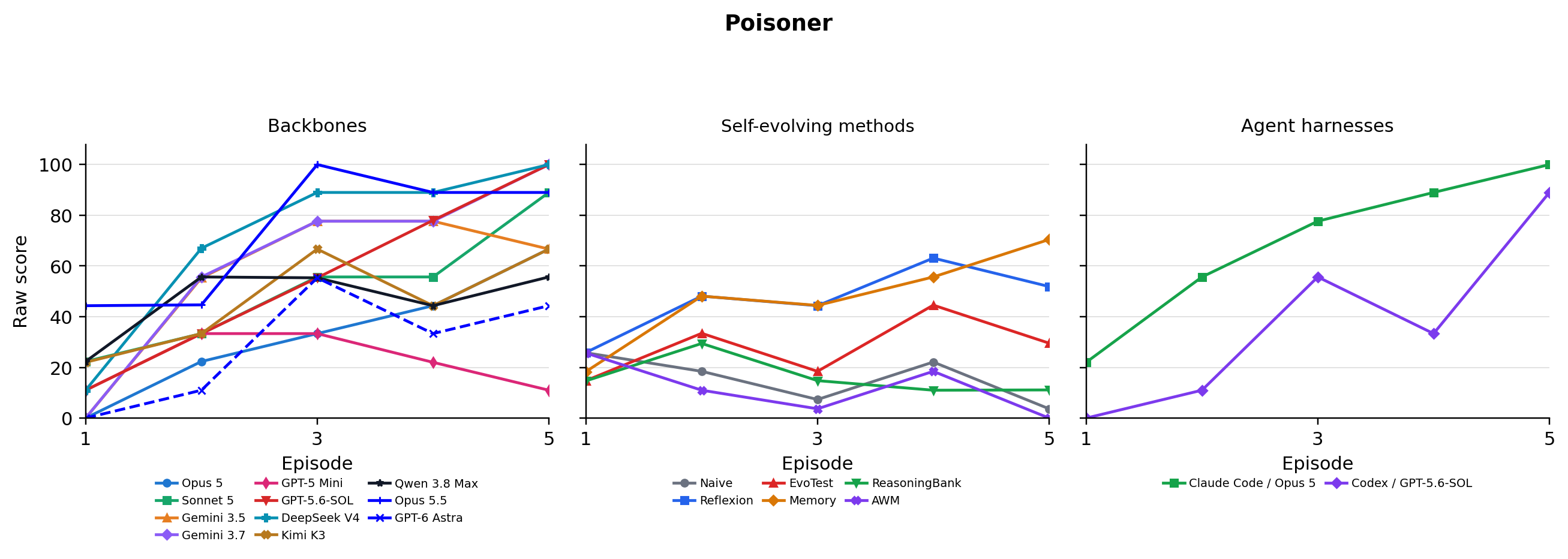}
\captionof{figure}{\revised{Poisoner: raw-score learning curves within the scoring window.}}
\label{fig:game-curve-poisoner}

\end{center}
\Needspace{10\baselineskip}
% Combined score table: normalized points and trial SD; human/First retained.
\begingroup
\setlength{\tabcolsep}{3pt}
\begin{longtable}{@{}lrrrrr@{}}
\caption{\captionrevision{Poisoner: normalized performance and learning, with trial SD. Rows follow descending Max. First denotes the average episode at which a trial first reaches its highest score.}}
\label{tab:game-scores-poisoner}\label{tab:trial-variability-poisoner}\\
\toprule
Configuration & First & Max $\uparrow$ & Mean $\uparrow$ & LG $\uparrow$ & LS $\uparrow$ \\
\midrule
\endfirsthead
\toprule
Configuration & First & Max $\uparrow$ & Mean $\uparrow$ & LG $\uparrow$ & LS $\uparrow$ \\
\midrule
\endhead
Human Top-1 & 3 & 100.0 & 66.8 & -- & -- \\
OpenCode + Gemini 3.7 Flash & 3.3 & $100.0 \pm 0.0$ & $62.2 \pm 25.3$ & $+61.0 \pm 9.5$ & $+22.20 \pm 1.91$ \\
OpenCode + DeepSeek V4 Pro & 3.7 & $100.0 \pm 0.0$ & $71.2 \pm 3.8$ & $+55.5 \pm 19.1$ & $+20.00 \pm 5.72$ \\
Claude Code + Opus 5 & 3.7 & $100.0 \pm 0.0$ & $68.9 \pm 19.3$ & $+55.7 \pm 9.8$ & $+18.93 \pm 3.87$ \\
OpenCode + GPT-5.6-SOL & 4.3 & $100.0 \pm 0.0$ & $55.5 \pm 21.4$ & $+66.8 \pm 16.8$ & $+22.27 \pm 5.10$ \\
\opusrevision{OpenCode + Opus 5.5} & \opusrevision{2.3} & \opusrevision{$100.0 \pm 0.0$} & \opusrevision{$73.4 \pm 24.0$} & \opusrevision{$+44.5 \pm 25.7$} & \opusrevision{$+13.37 \pm 8.88$} \\
Reflexion + Kimi K3 & 2.7 & $89.0 \pm 19.1$ & $57.8 \pm 36.7$ & $+22.2 \pm 58.7$ & $+8.87 \pm 16.81$ \\
EvoTest + Opus 5 & 3.7 & $89.0 \pm 19.1$ & $31.1 \pm 7.7$ & $+27.8 \pm 67.4$ & $+8.90 \pm 19.52$ \\
OpenCode + Sonnet 5 & 4.3 & $89.0 \pm 19.1$ & $51.2 \pm 25.3$ & $+44.5 \pm 10.0$ & $+15.57 \pm 5.16$ \\
Codex + GPT-5.6-SOL & 4.3 & $89.0 \pm 19.1$ & $37.8 \pm 4.0$ & $+55.7 \pm 19.2$ & $+20.03 \pm 3.30$ \\
EvoTest + Kimi K3 & 2.3 & $78.0 \pm 19.1$ & $37.9 \pm 31.5$ & $-16.8 \pm 60.3$ & $-3.37 \pm 18.63$ \\
Reflexion + Opus 5 & 2.7 & $78.0 \pm 19.1$ & $59.9 \pm 23.1$ & $+22.5 \pm 19.5$ & $+7.87 \pm 7.00$ \\
Memory + Opus 5 & 3.7 & $78.0 \pm 19.1$ & $55.5 \pm 27.0$ & $+28.2 \pm 9.7$ & $+10.13 \pm 3.30$ \\
OpenCode + Gemini 3.5 Flash & 2.3 & $77.7 \pm 38.7$ & $55.5 \pm 25.4$ & $+44.5 \pm 34.8$ & $+15.57 \pm 10.22$ \\
OpenCode + Kimi K3 & 3.7 & $77.7 \pm 38.7$ & $46.6 \pm 37.1$ & $+27.8 \pm 19.2$ & $+10.03 \pm 5.77$ \\
Reflexion + GPT-5 Mini & 3.3 & $67.0 \pm 0.0$ & $22.3 \pm 3.9$ & $+16.8 \pm 16.8$ & $+3.37 \pm 10.05$ \\
OpenCode + Qwen 3.8 Max & 2.3 & $66.7 \pm 33.5$ & $46.7 \pm 37.3$ & $+11.0 \pm 25.2$ & $+5.53 \pm 7.65$ \\
Memory + Kimi K3 & 2.7 & $66.7 \pm 57.7$ & $40.0 \pm 37.1$ & $+28.0 \pm 34.8$ & $+11.17 \pm 11.74$ \\
Memory + GPT-5 Mini & 3 & $66.7 \pm 57.7$ & $46.7 \pm 41.7$ & $+33.3 \pm 28.9$ & $+12.23 \pm 10.72$ \\
OpenCode + Opus 5 & 4.3 & $66.7 \pm 33.5$ & $33.3 \pm 35.3$ & $+44.3 \pm 25.5$ & $+15.53 \pm 8.41$ \\
EvoTest + GPT-5 Mini & 4.3 & $55.7 \pm 19.6$ & $15.5 \pm 7.7$ & $+27.8 \pm 9.8$ & $+6.67 \pm 0.06$ \\
\astrarevision{OpenCode + GPT-6 Astra} & \astrarevision{2.7} & \astrarevision{$55.3 \pm 38.7$} & \astrarevision{$28.8 \pm 21.6$} & \astrarevision{$+33.3 \pm 44.2$} & \astrarevision{$+11.10 \pm 13.91$} \\
ReasoningBank + GPT-5 Mini & 1.7 & $44.3 \pm 51.0$ & $15.5 \pm 21.4$ & $-16.7 \pm 16.8$ & $-3.33 \pm 3.35$ \\
Naive + Kimi K3 & 2 & $44.3 \pm 19.6$ & $13.3 \pm 11.7$ & $-22.3 \pm 42.1$ & $-7.80 \pm 11.75$ \\
Naive + GPT-5 Mini & 2.7 & $44.3 \pm 19.6$ & $13.3 \pm 6.7$ & $+11.2 \pm 25.4$ & $+2.23 \pm 7.65$ \\
ReasoningBank + Kimi K3 & 4 & $44.3 \pm 19.6$ & $15.5 \pm 10.2$ & $+0.2 \pm 16.5$ & $+2.27 \pm 5.06$ \\
OpenCode + GPT-5 Mini & 1.7 & $33.3 \pm 33.5$ & $22.1 \pm 20.3$ & $-5.7 \pm 25.5$ & $-1.13 \pm 8.36$ \\
AWM + GPT-5 Mini & 2 & $33.3 \pm 33.5$ & $8.9 \pm 10.2$ & $+5.7 \pm 25.5$ & $+0.03 \pm 6.65$ \\
AWM + Opus 5 & 1 & $33.0 \pm 0.0$ & $19.8 \pm 0.0$ & $-16.5 \pm 0.0$ & $-6.60 \pm 0.00$ \\
Naive + Opus 5 & 1 & $33.0 \pm 0.0$ & $19.8 \pm 0.0$ & $-16.5 \pm 0.0$ & $-6.60 \pm 0.00$ \\
ReasoningBank + Opus 5 & 1 & $33.0 \pm 0.0$ & $17.6 \pm 10.1$ & $-16.5 \pm 0.0$ & $-6.60 \pm 0.00$ \\
AWM + Kimi K3 & 1 & $33.0 \pm 0.0$ & $6.6 \pm 0.0$ & $-16.5 \pm 0.0$ & $-6.60 \pm 0.00$ \\
\bottomrule
\end{longtable}
\endgroup

\revised{\textbf{Seed-matched evidence.} Labels for T2, T4, and T5 use the three seed-2 trials used for score reporting. This seed-matched assessment covers these three targets, not the other labels, and does not determine the episode in which a target was discovered.}
\begin{longtable}{@{}p{0.07\linewidth}p{0.87\linewidth}@{}}
\caption{\captionrevision{\revised{Poisoner: criteria for assessing rule discovery.}}}\label{tab:targets-definition-poisoner}\\
\toprule
\textbf{Target} & \textbf{Criterion} \\
\midrule
\endfirsthead
\toprule
\textbf{Target} & \textbf{Criterion (continued)} \\
\midrule
\endhead
T1 & \revised{Three-banquet inference interface: eight dishes per banquet, one serve action per banquet, score \(0/33/67/100\), and only one labeled accept/reject outcome per menu.} \\
T2 & \revised{The exact five-feature conjunction for the evaluated seed 2: not sweet, spicy, meaty, fragrant, and soft are required. Sourness, saltiness, and temperature are irrelevant.} \\
T3 & \revised{Wildcard and near-miss handling: each menu contains at least one full match and at least one 4-of-5 bait, so four-feature theories fail.} \\
T4 & \revised{For each new menu, check the learned combination of required features. Ignore irrelevant traits and distinguish a full match from dishes satisfying only four requirements.} \\
T5 & \revised{3/3 regicide: carry the exact five locked features across banquets and serve a full match in every reshuffled menu.} \\
\bottomrule
\end{longtable}
\begin{center}
\begin{minipage}{\linewidth}
\centering
\captionof{table}{\captionrevision{\revised{Poisoner: rule-discovery evidence. S: complete evidence in at least one trial; P: partial evidence only; --: insufficient evidence. Evidence may include episodes beyond the scoring window.}}}
\label{tab:rq1-targets-game-poisoner}
{
\appendixtable{%
\begin{tabular}{@{}l*{5}{c}@{}}
\toprule
\textbf{Configuration} & T1 & T2 & T3 & T4 & T5 \\
\midrule
\multicolumn{6}{@{}l}{\textbf{backbone comparison}} \\
Claude Opus~5 & S & S & S & S & S \\
Qwen 3.8 Max & S & S & S & S & S \\
Kimi~K3 & S & S & S & S & S \\
Gemini~3.5 Flash & S & S & S & S & S \\
Gemini~3.7 Flash & S & S & S & S & S \\
GPT-5.6-SOL & S & S & S & S & S \\
DeepSeek V4 Pro & S & S & S & S & S \\
Claude Sonnet~5 & S & S & S & S & S \\
GPT-5 Mini & S & P & -- & P & -- \\
\midrule
\multicolumn{6}{@{}l}{\textbf{agent harness comparison}} \\
Claude Code + Opus~5 & S & S & S & S & S \\
Codex + GPT-5.6-SOL & S & S & S & S & S \\
\bottomrule
\end{tabular}
}
}

\end{minipage}
\end{center}

\subsection{\revised{Roommates}}
\label{app:game-hezu}

\revised{\textbf{Learning task.} Infer pairwise and nonlinear grouping rules from room-level feedback. Re-shuffled game, scored seed 1, first 5 episodes, reference score $C=7$.}
\begin{center}
\includegraphics[width=0.95\linewidth]{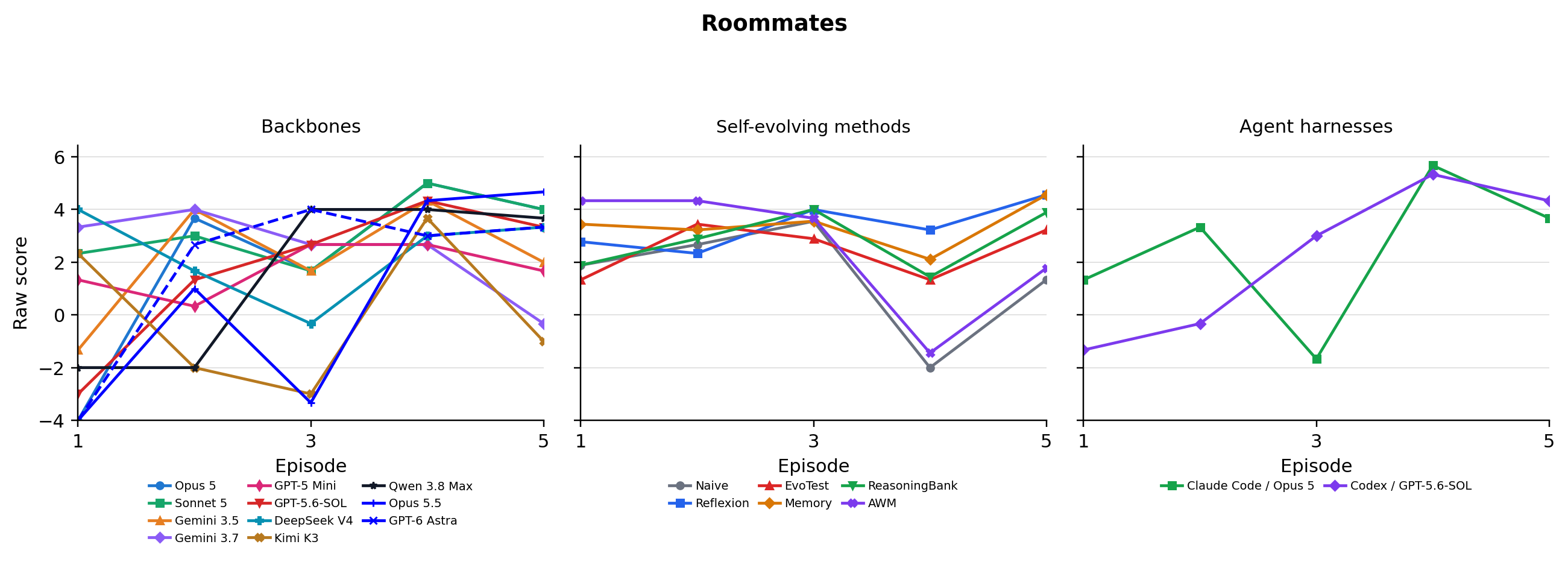}
\captionof{figure}{\revised{Roommates: raw-score learning curves within the scoring window.}}
\label{fig:game-curve-hezu}

\end{center}
\Needspace{10\baselineskip}
% Combined score table: normalized points and trial SD; human/First retained.
\begingroup
\setlength{\tabcolsep}{3pt}
\begin{longtable}{@{}lrrrrr@{}}
\caption{\captionrevision{Roommates: normalized performance and learning, with trial SD. Rows follow descending Max. First denotes the average episode at which a trial first reaches its highest score.}}
\label{tab:game-scores-hezu}\label{tab:trial-variability-hezu}\\
\toprule
Configuration & First & Max $\uparrow$ & Mean $\uparrow$ & LG $\uparrow$ & LS $\uparrow$ \\
\midrule
\endfirsthead
\toprule
Configuration & First & Max $\uparrow$ & Mean $\uparrow$ & LG $\uparrow$ & LS $\uparrow$ \\
\midrule
\endhead
Human Top-1 & 3 & 100.0 & -4.7 & -- & -- \\
Reflexion + Opus 5 & 3.7 & $100.0 \pm 0.0$ & $71.4 \pm 7.6$ & $+28.6 \pm 18.9$ & $+8.11 \pm 5.41$ \\
ReasoningBank + Opus 5 & 3.3 & $90.5 \pm 8.2$ & $55.2 \pm 1.6$ & $+42.8 \pm 31.1$ & $+13.31 \pm 5.02$ \\
Reflexion + GPT-5 Mini & 4.3 & $85.7 \pm 24.7$ & $39.0 \pm 36.7$ & $+9.5 \pm 14.9$ & $+6.19 \pm 5.77$ \\
Claude Code + Opus 5 & 4 & $81.0 \pm 8.2$ & $35.2 \pm 25.8$ & $+33.4 \pm 36.0$ & $+10.03 \pm 13.63$ \\
OpenCode + DeepSeek V4 Pro & 4.7 & $81.0 \pm 8.2$ & $33.3 \pm 8.7$ & $+4.7 \pm 36.0$ & $-0.01 \pm 13.09$ \\
Codex + GPT-5.6-SOL & 3.3 & $76.2 \pm 16.5$ & $31.4 \pm 33.0$ & $+80.9 \pm 66.0$ & $+24.26 \pm 17.55$ \\
OpenCode + Sonnet 5 & 3.7 & $76.2 \pm 16.5$ & $45.7 \pm 12.5$ & $+26.2 \pm 21.8$ & $+7.63 \pm 5.02$ \\
OpenCode + Opus 5 & 3.7 & $76.2 \pm 16.5$ & $29.5 \pm 11.5$ & $+66.6 \pm 16.5$ & $+24.76 \pm 5.77$ \\
OpenCode + Qwen 3.8 Max & 4 & $76.2 \pm 16.5$ & $21.9 \pm 23.3$ & $+83.4 \pm 32.2$ & $+24.77 \pm 5.95$ \\
\opusrevision{OpenCode + Opus 5.5} & \opusrevision{4.7} & \opusrevision{$76.2 \pm 21.8$} & \opusrevision{$7.6 \pm 14.4$} & \opusrevision{$+85.7 \pm 12.4$} & \opusrevision{$+29.52 \pm 1.65$} \\
AWM + Opus 5 & 1 & $71.4 \pm 24.7$ & $45.7 \pm 17.8$ & $-50.0 \pm 12.4$ & $-12.39 \pm 4.36$ \\
Memory + Kimi K3 & 3.7 & $71.4 \pm 14.3$ & $54.3 \pm 2.9$ & $-7.1 \pm 7.1$ & $-0.00 \pm 2.86$ \\
Memory + Opus 5 & 3.7 & $71.4 \pm 14.3$ & $44.8 \pm 4.4$ & $+11.9 \pm 10.9$ & $+3.80 \pm 2.97$ \\
EvoTest + Opus 5 & 5 & $71.4 \pm 0.0$ & $45.7 \pm 11.4$ & $+9.6 \pm 23.0$ & $+3.34 \pm 4.59$ \\
OpenCode + Kimi K3 & 2 & $66.7 \pm 16.5$ & $0.0 \pm 7.6$ & $+16.7 \pm 10.9$ & $-1.41 \pm 6.55$ \\
EvoTest + GPT-5 Mini & 2.7 & $66.7 \pm 16.5$ & $35.3 \pm 19.4$ & $+30.9 \pm 10.9$ & $+13.33 \pm 5.95$ \\
Memory + GPT-5 Mini & 3.7 & $66.7 \pm 8.2$ & $45.7 \pm 7.6$ & $-4.7 \pm 32.2$ & $+0.97 \pm 12.88$ \\
\astrarevision{OpenCode + GPT-6 Astra} & \astrarevision{3.3} & \astrarevision{$66.7 \pm 16.5$} & \astrarevision{$25.7 \pm 2.9$} & \astrarevision{$+54.8 \pm 10.9$} & \astrarevision{$+21.43 \pm 5.15$} \\
Naive + Opus 5 & 1.7 & $61.9 \pm 8.2$ & $32.4 \pm 1.6$ & $-64.3 \pm 0.0$ & $-17.14 \pm 0.00$ \\
ReasoningBank + GPT-5 Mini & 2.3 & $61.9 \pm 8.2$ & $27.6 \pm 6.6$ & $-35.7 \pm 49.5$ & $-5.71 \pm 19.79$ \\
OpenCode + Gemini 3.5 Flash & 2.7 & $61.9 \pm 8.2$ & $30.5 \pm 3.3$ & $+26.1 \pm 28.9$ & $+9.99 \pm 12.37$ \\
OpenCode + GPT-5 Mini & 2.7 & $61.9 \pm 8.2$ & $24.8 \pm 6.6$ & $+19.1 \pm 64.4$ & $+4.31 \pm 18.90$ \\
OpenCode + GPT-5.6-SOL & 2.7 & $61.9 \pm 8.2$ & $24.7 \pm 17.5$ & $+66.6 \pm 54.1$ & $+22.37 \pm 14.31$ \\
AWM + GPT-5 Mini & 1 & $57.1 \pm 0.0$ & $31.4 \pm 0.0$ & $-64.3 \pm 0.0$ & $-17.14 \pm 0.00$ \\
AWM + Kimi K3 & 1 & $57.1 \pm 0.0$ & $31.4 \pm 0.0$ & $-64.3 \pm 0.0$ & $-17.14 \pm 0.00$ \\
ReasoningBank + Kimi K3 & 1.3 & $57.1 \pm 0.0$ & $38.1 \pm 8.2$ & $+4.8 \pm 39.3$ & $+3.34 \pm 15.80$ \\
OpenCode + Gemini 3.7 Flash & 1.3 & $57.1 \pm 0.0$ & $35.3 \pm 5.9$ & $-35.6 \pm 12.4$ & $-12.36 \pm 5.95$ \\
Reflexion + Kimi K3 & 1.3 & $57.1 \pm 0.0$ & $34.3 \pm 4.9$ & $+19.1 \pm 25.1$ & $+4.77 \pm 9.18$ \\
EvoTest + Kimi K3 & 1.3 & $57.1 \pm 0.0$ & $23.8 \pm 13.2$ & $-45.2 \pm 33.0$ & $-9.51 \pm 13.20$ \\
Naive + Kimi K3 & 1.3 & $57.1 \pm 0.0$ & $20.0 \pm 11.4$ & $-21.4 \pm 37.8$ & $-3.33 \pm 12.15$ \\
Naive + GPT-5 Mini & 2 & $57.1 \pm 0.0$ & $11.4 \pm 18.7$ & $-26.1 \pm 33.0$ & $-4.27 \pm 11.69$ \\
\bottomrule
\end{longtable}
\endgroup
\begin{longtable}{@{}p{0.07\linewidth}p{0.87\linewidth}@{}}
\caption{\captionrevision{\revised{Roommates: criteria for assessing rule discovery.}}}\label{tab:targets-definition-roommates}\\
\toprule
\textbf{Target} & \textbf{Criterion} \\
\midrule
\endfirsthead
\toprule
\textbf{Target} & \textbf{Criterion (continued)} \\
\midrule
\endhead
T1 & \revised{Feedback decomposition: one-shot assignment of nine tenants into three triple rooms, then infer pairwise value from room-level harmony sums rather than opaque labels.} \\
T2 & \revised{Seed-1 hidden five-type ring: Extrovert--Homebody--Foodie--Night Owl--Neat Freak up to rotation/reversal, where adjacent types score \(+1\), distance-two types score \(-1\), and same type scores 0.} \\
T3 & \revised{Nonlinear trio chemistry: \(+5\) for Night-Owl/Extrovert/Homebody and Foodie/Foodie/Neat-Freak, and \(-5\) for Foodie/Neat-Freak/Homebody.} \\
T4 & \revised{Search the 280 possible assignments for the current tenants. Account for the part of each room's score that pairwise values alone cannot explain.} \\
T5 & \revised{7-point partition: combine the ring pair map with the three trio exceptions and pick all three rooms jointly, not greedily one room at a time.} \\
\bottomrule
\end{longtable}
\begin{center}
\begin{minipage}{\linewidth}
\centering
\captionof{table}{\captionrevision{\revised{Roommates: rule-discovery evidence. S: complete evidence in at least one trial; P: partial evidence only; --: insufficient evidence. Evidence may include episodes beyond the scoring window.}}}
\label{tab:rq1-targets-game-roommates}
{
\appendixtable{%
\begin{tabular}{@{}l*{5}{c}@{}}
\toprule
\textbf{Configuration} & T1 & T2 & T3 & T4 & T5 \\
\midrule
\multicolumn{6}{@{}l}{\textbf{backbone comparison}} \\
Claude Opus~5 & S & S & S & S & S \\
Qwen 3.8 Max & S & S & S & S & -- \\
Kimi~K3 & S & S & S & S & S \\
Gemini~3.5 Flash & S & S & S & S & -- \\
Gemini~3.7 Flash & S & S & S & S & S \\
GPT-5.6-SOL & S & S & S & S & S \\
DeepSeek V4 Pro & S & S & S & S & S \\
Claude Sonnet~5 & S & S & S & S & -- \\
GPT-5 Mini & S & -- & S & S & -- \\
\midrule
\multicolumn{6}{@{}l}{\textbf{agent harness comparison}} \\
Claude Code + Opus~5 & S & S & S & S & S \\
Codex + GPT-5.6-SOL & S & S & S & S & -- \\
\bottomrule
\end{tabular}
}
}

\end{minipage}
\end{center}

\subsection{\revised{RedEye}}
\label{app:game-redeye}

\revised{\textbf{Learning task.} Combine character evidence and execute the rescue steps in order. Fixed game, scored seed 2, first 10 episodes, reference score $C=100$.}
\begin{center}
\includegraphics[width=0.95\linewidth]{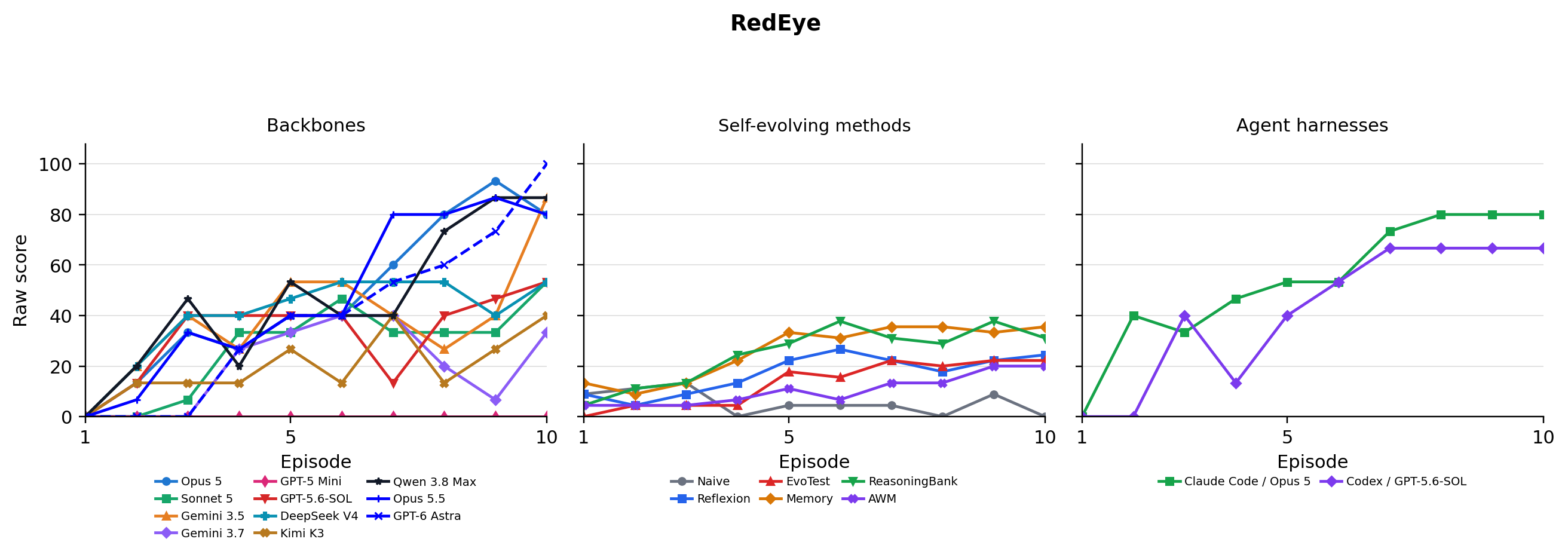}
\captionof{figure}{\revised{RedEye: raw-score learning curves within the scoring window.}}
\label{fig:game-curve-redeye}

\end{center}
\Needspace{10\baselineskip}
% Combined score table: normalized points and trial SD; human/First retained.
\begingroup
\setlength{\tabcolsep}{3pt}
\begin{longtable}{@{}lrrrrr@{}}
\caption{\captionrevision{RedEye: normalized performance and learning, with trial SD. Rows follow descending Max. First denotes the average episode at which a trial first reaches its highest score.}}
\label{tab:game-scores-redeye}\label{tab:trial-variability-redeye}\\
\toprule
Configuration & First & Max $\uparrow$ & Mean $\uparrow$ & LG $\uparrow$ & LS $\uparrow$ \\
\midrule
\endfirsthead
\toprule
Configuration & First & Max $\uparrow$ & Mean $\uparrow$ & LG $\uparrow$ & LS $\uparrow$ \\
\midrule
\endhead
OpenCode + Opus 5 & 8.3 & $100.0 \pm 0.0$ & $46.7 \pm 1.2$ & $+68.9 \pm 16.8$ & $+9.78 \pm 1.63$ \\
\astrarevision{OpenCode + GPT-6 Astra} & \astrarevision{9} & \astrarevision{$100.0 \pm 0.0$} & \astrarevision{$39.3 \pm 3.1$} & \astrarevision{$+77.8 \pm 3.8$} & \astrarevision{$+10.87 \pm 0.31$} \\
OpenCode + Qwen 3.8 Max & 6.7 & $86.7 \pm 23.1$ & $46.7 \pm 6.4$ & $+60.0 \pm 17.6$ & $+8.65 \pm 2.14$ \\
OpenCode + Gemini 3.5 Flash & 7.3 & $86.7 \pm 23.1$ & $38.7 \pm 15.1$ & $+31.1 \pm 34.2$ & $+5.41 \pm 4.58$ \\
\opusrevision{OpenCode + Opus 5.5} & \opusrevision{7.7} & \opusrevision{$86.7 \pm 23.1$} & \opusrevision{$47.3 \pm 11.7$} & \opusrevision{$+68.9 \pm 30.8$} & \opusrevision{$+10.14 \pm 4.70$} \\
Claude Code + Opus 5 & 5.3 & $80.0 \pm 34.6$ & $54.0 \pm 17.4$ & $+55.6 \pm 36.7$ & $+7.96 \pm 5.34$ \\
Codex + GPT-5.6-SOL & 5 & $66.7 \pm 30.6$ & $41.3 \pm 16.0$ & $+53.3 \pm 30.6$ & $+8.32 \pm 4.30$ \\
Human Top-1 & 2 & 60.0 & 38.0 & -- & -- \\
ReasoningBank + Opus 5 & 4 & $60.0 \pm 34.6$ & $40.0 \pm 18.3$ & $+35.6 \pm 23.4$ & $+4.69 \pm 2.68$ \\
OpenCode + DeepSeek V4 Pro & 3.7 & $53.3 \pm 11.5$ & $40.0 \pm 19.3$ & $+28.9 \pm 10.2$ & $+4.44 \pm 1.28$ \\
OpenCode + GPT-5.6-SOL & 5.3 & $53.3 \pm 23.1$ & $32.7 \pm 5.0$ & $+28.9 \pm 3.8$ & $+3.84 \pm 0.67$ \\
OpenCode + Sonnet 5 & 6 & $53.3 \pm 11.5$ & $27.3 \pm 13.3$ & $+37.8 \pm 23.4$ & $+5.21 \pm 3.30$ \\
EvoTest + Opus 5 & 6.7 & $53.3 \pm 11.5$ & $31.3 \pm 12.7$ & $+42.2 \pm 10.2$ & $+6.18 \pm 1.19$ \\
OpenCode + Gemini 3.7 Flash & 7.3 & $53.3 \pm 11.5$ & $20.0 \pm 6.9$ & $+20.0 \pm 20.0$ & $+2.99 \pm 2.43$ \\
Reflexion + Kimi K3 & 6 & $46.7 \pm 11.5$ & $28.7 \pm 4.2$ & $+37.8 \pm 10.2$ & $+5.29 \pm 1.24$ \\
Naive + Opus 5 & 1.3 & $40.0 \pm 0.0$ & $12.7 \pm 1.2$ & $-22.2 \pm 20.4$ & $-2.71 \pm 2.71$ \\
Memory + Opus 5 & 2 & $40.0 \pm 0.0$ & $32.0 \pm 4.0$ & $+15.6 \pm 23.4$ & $+2.02 \pm 2.92$ \\
Memory + Kimi K3 & 3.3 & $40.0 \pm 0.0$ & $31.3 \pm 7.6$ & $+26.7 \pm 23.1$ & $+3.52 \pm 3.05$ \\
ReasoningBank + Kimi K3 & 3.7 & $40.0 \pm 0.0$ & $27.3 \pm 5.0$ & $+31.1 \pm 15.4$ & $+4.24 \pm 1.58$ \\
AWM + Opus 5 & 4 & $40.0 \pm 20.0$ & $15.3 \pm 21.4$ & $+0.0 \pm 13.3$ & $+0.28 \pm 2.37$ \\
OpenCode + Kimi K3 & 4.7 & $40.0 \pm 0.0$ & $20.0 \pm 10.6$ & $+17.8 \pm 20.4$ & $+3.15 \pm 2.46$ \\
ReasoningBank + GPT-5 Mini & 6 & $40.0 \pm 0.0$ & $7.3 \pm 4.2$ & $+2.2 \pm 10.2$ & $+0.77 \pm 1.70$ \\
Reflexion + Opus 5 & 4 & $33.3 \pm 11.5$ & $15.3 \pm 21.4$ & $-2.2 \pm 10.2$ & $-0.36 \pm 1.67$ \\
AWM + GPT-5 Mini & 4.7 & $33.3 \pm 30.6$ & $6.0 \pm 5.3$ & $+13.3 \pm 13.3$ & $+1.58 \pm 1.71$ \\
Memory + GPT-5 Mini & 4 & $26.7 \pm 23.1$ & $15.3 \pm 13.3$ & $+26.7 \pm 23.1$ & $+3.92 \pm 3.39$ \\
Reflexion + GPT-5 Mini & 4 & $26.7 \pm 23.1$ & $7.3 \pm 9.5$ & $+6.7 \pm 11.5$ & $+1.25 \pm 1.96$ \\
Naive + Kimi K3 & 4.7 & $26.7 \pm 11.5$ & $4.0 \pm 2.0$ & $-2.2 \pm 13.9$ & $-0.00 \pm 1.83$ \\
AWM + Kimi K3 & 5.3 & $26.7 \pm 23.1$ & $10.0 \pm 8.7$ & $+26.7 \pm 23.1$ & $+3.76 \pm 3.26$ \\
EvoTest + GPT-5 Mini & 2.3 & $13.3 \pm 23.1$ & $2.0 \pm 3.5$ & $+0.0 \pm 0.0$ & $+0.04 \pm 0.07$ \\
EvoTest + Kimi K3 & 2.7 & $13.3 \pm 23.1$ & $6.7 \pm 11.5$ & $+13.3 \pm 23.1$ & $+2.02 \pm 3.50$ \\
Naive + GPT-5 Mini & 1 & $0.0 \pm 0.0$ & $0.0 \pm 0.0$ & $+0.0 \pm 0.0$ & $+0.00 \pm 0.00$ \\
OpenCode + GPT-5 Mini & 1 & $0.0 \pm 0.0$ & $0.0 \pm 0.0$ & $+0.0 \pm 0.0$ & $+0.00 \pm 0.00$ \\
\bottomrule
\end{longtable}
\endgroup
\begin{longtable}{@{}p{0.07\linewidth}p{0.87\linewidth}@{}}
\caption{\captionrevision{\revised{RedEye: criteria for assessing rule discovery.}}}\label{tab:targets-definition-redeye}\\
\toprule
\textbf{Target} & \textbf{Criterion} \\
\midrule
\endfirsthead
\toprule
\textbf{Target} & \textbf{Criterion (continued)} \\
\midrule
\endhead
T1 & \revised{Six-location time-loop map on the Sea-Azure, with the first loop as a forced prologue and later loops retaining player memory.} \\
T2 & \revised{Culprit evidence channels: oldman/backpacker saw graycoat track the suite owner, caller heard the insurance phone call, dining search finds over-insurance and forged seaworthiness evidence, and graycoat can supply documents.} \\
T3 & \revised{Role inversion: graycoat is the ex-engineer ally seeking the stern scuttling rig, while suit is the owner staging insurance fraud with an engine-room accomplice.} \\
T4 & \revised{Gating rules: confronting suit needs hard evidence plus at least two independent culprit clues. Calling coast guard needs evidence and must happen before the radio window closes at segment 24.} \\
T5 & \revised{Trust and device prerequisites: comfort graycoat only after suit is restrained and after learning his victim story, and disarm only after comfort plus device knowledge.} \\
T6 & \revised{Failure avoidance: restraining graycoat, accusing without evidence, comforting too early, or solo disarming ends the loop with only milestone credit.} \\
T7 & \revised{100 rescue script: gather two culprit clues and hard evidence, confront suit, call coast guard, learn/comfort graycoat, then seal the stern valve while the coast guard blocks the engine-room accomplice.} \\
\bottomrule
\end{longtable}
\begin{center}
\begin{minipage}{\linewidth}
\centering
\captionof{table}{\captionrevision{\revised{RedEye: rule-discovery evidence. S: complete evidence in at least one trial; P: partial evidence only; --: insufficient evidence. Evidence may include episodes beyond the scoring window.}}}
\label{tab:rq1-targets-game-redeye}
{
\appendixtable{%
\begin{tabular}{@{}l*{7}{c}@{}}
\toprule
\textbf{Configuration} & T1 & T2 & T3 & T4 & T5 & T6 & T7 \\
\midrule
\multicolumn{8}{@{}l}{\textbf{backbone comparison}} \\
Claude Opus~5 & S & S & S & S & S & S & S \\
Qwen 3.8 Max & S & S & S & S & S & S & S \\
Kimi~K3 & S & S & S & S & -- & -- & -- \\
Gemini~3.5 Flash & S & S & S & S & S & S & S \\
Gemini~3.7 Flash & S & S & S & S & S & -- & -- \\
GPT-5.6-SOL & S & S & S & S & -- & S & -- \\
DeepSeek V4 Pro & S & S & S & S & -- & -- & -- \\
Claude Sonnet~5 & S & S & S & S & S & -- & -- \\
GPT-5 Mini & S & S & S & S & -- & S & -- \\
\midrule
\multicolumn{8}{@{}l}{\textbf{agent harness comparison}} \\
Claude Code + Opus~5 & S & S & S & S & S & S & S \\
Codex + GPT-5.6-SOL & S & S & S & S & S & S & S \\
\bottomrule
\end{tabular}
}
}

\end{minipage}
\end{center}

\subsection{\revised{Butterfly}}
\label{app:game-butterfly}

\revised{\textbf{Learning task.} Infer inheritance laws to breed new target combinations. Re-shuffled game, scored seed 1, first 5 episodes, reference score $C=439$.}
\begin{center}
\includegraphics[width=0.95\linewidth]{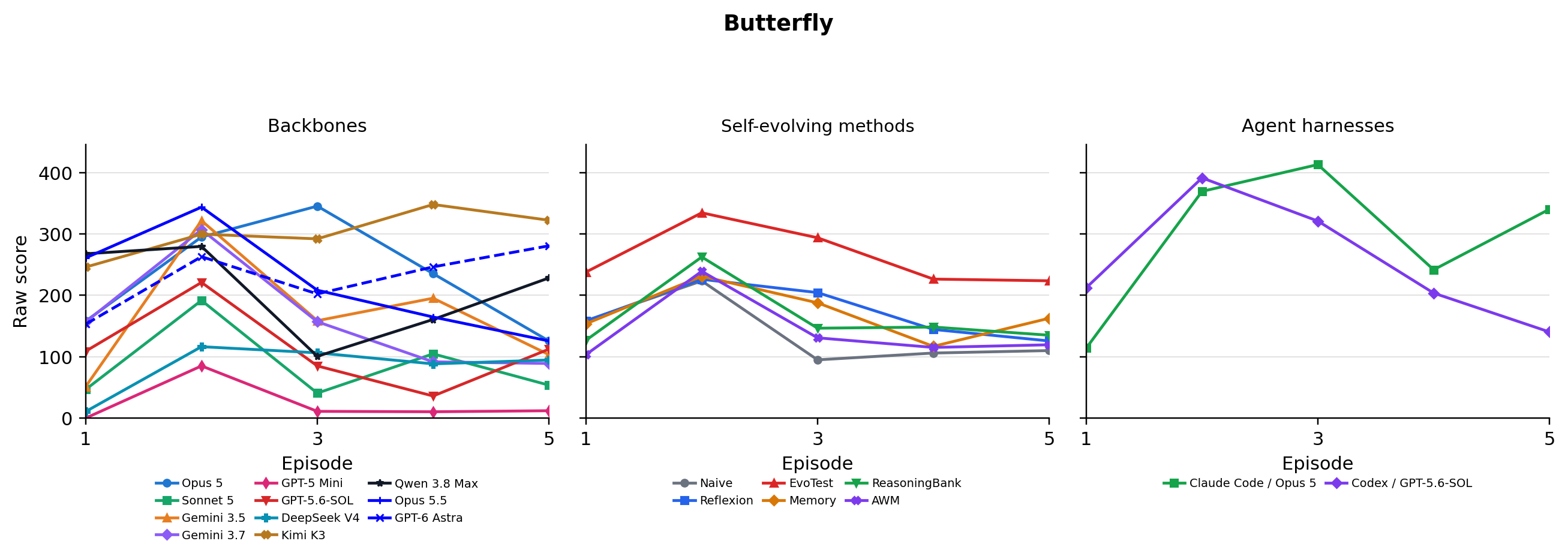}
\captionof{figure}{\revised{Butterfly: raw-score learning curves within the scoring window.}}
\label{fig:game-curve-butterfly}

\end{center}
\Needspace{10\baselineskip}
% Combined score table: normalized points and trial SD; human/First retained.
\begingroup
\setlength{\tabcolsep}{3pt}
\begin{longtable}{@{}lrrrrr@{}}
\caption{\captionrevision{Butterfly: normalized performance and learning, with trial SD. Rows follow descending Max. First denotes the average episode at which a trial first reaches its highest score.}}
\label{tab:game-scores-butterfly}\label{tab:trial-variability-butterfly}\\
\toprule
Configuration & First & Max $\uparrow$ & Mean $\uparrow$ & LG $\uparrow$ & LS $\uparrow$ \\
\midrule
\endfirsthead
\toprule
Configuration & First & Max $\uparrow$ & Mean $\uparrow$ & LG $\uparrow$ & LS $\uparrow$ \\
\midrule
\endhead
EvoTest + Opus 5 & 2.7 & $99.1 \pm 1.6$ & $80.9 \pm 16.8$ & $+0.0 \pm 7.0$ & $+0.93 \pm 3.22$ \\
Claude Code + Opus 5 & 4.3 & $98.2 \pm 3.2$ & $67.3 \pm 12.8$ & $+11.2 \pm 29.7$ & $+7.40 \pm 8.06$ \\
EvoTest + Kimi K3 & 2.3 & $96.3 \pm 4.3$ & $71.6 \pm 19.1$ & $-18.6 \pm 24.2$ & $-3.77 \pm 5.32$ \\
Human Top-1 & 4 & 91.6 & 62.0 & -- & -- \\
Codex + GPT-5.6-SOL & 2 & $89.1 \pm 1.4$ & $57.8 \pm 18.9$ & $-29.6 \pm 8.6$ & $-7.55 \pm 2.94$ \\
OpenCode + Kimi K3 & 4.7 & $83.8 \pm 28.0$ & $68.7 \pm 33.0$ & $+14.2 \pm 16.5$ & $+4.58 \pm 4.93$ \\
Naive + GPT-5 Mini & 2 & $81.1 \pm 3.4$ & $43.2 \pm 3.9$ & $-27.1 \pm 17.7$ & $-7.05 \pm 5.43$ \\
\opusrevision{OpenCode + Opus 5.5} & \opusrevision{1.7} & \opusrevision{$80.4 \pm 5.7$} & \opusrevision{$50.3 \pm 4.3$} & \opusrevision{$-35.8 \pm 17.3$} & \opusrevision{$-10.24 \pm 5.69$} \\
Reflexion + GPT-5 Mini & 2.3 & $79.5 \pm 7.1$ & $48.4 \pm 16.4$ & $-19.4 \pm 24.5$ & $-6.83 \pm 6.33$ \\
OpenCode + Opus 5 & 3 & $78.7 \pm 11.7$ & $52.8 \pm 11.6$ & $-10.5 \pm 8.6$ & $-2.83 \pm 5.30$ \\
OpenCode + Qwen 3.8 Max & 2.7 & $75.5 \pm 13.4$ & $47.2 \pm 8.6$ & $-18.0 \pm 13.9$ & $-4.50 \pm 6.15$ \\
Memory + GPT-5 Mini & 2.7 & $75.4 \pm 21.0$ & $49.4 \pm 14.0$ & $-14.9 \pm 20.1$ & $-4.43 \pm 4.77$ \\
ReasoningBank + Opus 5 & 3.3 & $74.6 \pm 22.4$ & $43.0 \pm 17.7$ & $+4.6 \pm 19.5$ & $+3.70 \pm 3.30$ \\
AWM + GPT-5 Mini & 2.3 & $73.7 \pm 15.5$ & $34.9 \pm 1.7$ & $-29.3 \pm 11.8$ & $-6.59 \pm 2.93$ \\
ReasoningBank + GPT-5 Mini & 1.7 & $73.5 \pm 4.3$ & $36.8 \pm 8.5$ & $-26.0 \pm 23.6$ & $-7.81 \pm 8.13$ \\
OpenCode + Gemini 3.5 Flash & 2 & $73.3 \pm 5.0$ & $37.8 \pm 4.2$ & $-8.5 \pm 21.2$ & $-0.51 \pm 7.88$ \\
AWM + Kimi K3 & 2.7 & $72.4 \pm 12.8$ & $40.2 \pm 6.5$ & $-1.4 \pm 24.0$ & $+0.36 \pm 5.23$ \\
OpenCode + Gemini 3.7 Flash & 2 & $69.9 \pm 30.0$ & $36.5 \pm 16.1$ & $-32.2 \pm 29.8$ & $-7.99 \pm 8.51$ \\
\astrarevision{OpenCode + GPT-6 Astra} & \astrarevision{3} & \astrarevision{$69.7 \pm 44.6$} & \astrarevision{$52.2 \pm 40.7$} & \astrarevision{$+12.6 \pm 20.3$} & \astrarevision{$+5.44 \pm 7.26$} \\
Reflexion + Kimi K3 & 2.7 & $66.4 \pm 30.8$ & $33.7 \pm 15.7$ & $-25.0 \pm 7.6$ & $-7.09 \pm 1.59$ \\
ReasoningBank + Kimi K3 & 2 & $62.6 \pm 21.9$ & $32.1 \pm 10.3$ & $-14.8 \pm 15.5$ & $-2.57 \pm 4.54$ \\
EvoTest + GPT-5 Mini & 2 & $59.5 \pm 22.1$ & $27.4 \pm 5.2$ & $-23.2 \pm 16.8$ & $-6.48 \pm 4.52$ \\
Memory + Kimi K3 & 3.3 & $58.5 \pm 26.3$ & $32.9 \pm 10.3$ & $-6.4 \pm 13.8$ & $+0.74 \pm 4.92$ \\
Memory + Opus 5 & 2 & $58.1 \pm 29.1$ & $34.2 \pm 13.1$ & $-14.7 \pm 5.9$ & $-2.89 \pm 2.24$ \\
Reflexion + Opus 5 & 3.7 & $55.9 \pm 8.4$ & $35.3 \pm 4.0$ & $+5.4 \pm 18.9$ & $+3.89 \pm 3.94$ \\
Naive + Opus 5 & 1.7 & $52.2 \pm 25.0$ & $27.4 \pm 6.2$ & $-19.0 \pm 22.1$ & $-5.17 \pm 7.92$ \\
OpenCode + GPT-5.6-SOL & 2 & $50.3 \pm 22.2$ & $25.7 \pm 4.4$ & $-20.7 \pm 15.0$ & $-4.06 \pm 3.40$ \\
OpenCode + Sonnet 5 & 1.7 & $43.7 \pm 46.1$ & $19.9 \pm 20.6$ & $-9.1 \pm 12.9$ & $-1.66 \pm 3.51$ \\
Naive + Kimi K3 & 2.7 & $42.9 \pm 18.2$ & $24.1 \pm 5.0$ & $-10.6 \pm 17.6$ & $-2.45 \pm 5.04$ \\
AWM + Opus 5 & 3 & $39.5 \pm 9.3$ & $21.7 \pm 2.7$ & $-6.2 \pm 12.1$ & $-0.05 \pm 3.32$ \\
OpenCode + DeepSeek V4 Pro & 2.7 & $28.2 \pm 44.5$ & $19.0 \pm 32.0$ & $+6.4 \pm 9.0$ & $+3.19 \pm 4.68$ \\
OpenCode + GPT-5 Mini & 1.3 & $19.4 \pm 33.5$ & $5.4 \pm 9.3$ & $-7.1 \pm 12.4$ & $-1.15 \pm 2.00$ \\
\bottomrule
\end{longtable}
\endgroup
\begin{longtable}{@{}p{0.07\linewidth}p{0.87\linewidth}@{}}
\caption{\captionrevision{\revised{Butterfly: criteria for assessing rule discovery.}}}\label{tab:targets-definition-butterfly}\\
\toprule
\textbf{Target} & \textbf{Criterion} \\
\midrule
\endfirsthead
\toprule
\textbf{Target} & \textbf{Criterion (continued)} \\
\midrule
\endhead
T1 & \revised{Breeding interface: four visible traits, deterministic offspring, six mate actions, permanent founders, and capacity-limited offspring management.} \\
T2 & \revised{Antenna dominance: seed-1 strength order has Feathery strongest, then Straight, Hooked, and Curled weakest.} \\
T3 & \revised{Size law: child size is the half-up average of parent indices, making extreme sizes require matching/extreme donors.} \\
T4 & \revised{Color law: seed-1 seven-color nontransitive ring Purple--Blue--Green--Yellow--Cyan--Orange--Crimson, with each color beating the next three and no global strongest color.} \\
T5 & \revised{Pattern law: normally max-complexity wins, but Curled antenna suppresses the child pattern to Plain.} \\
T6 & \revised{Scoring law: offspring gain by target-match count \((0,4,18,49,120)\), novelty bonus, duplicate decay, and first-exact unused-cross bonus.} \\
T7 & \revised{439 breeding sequence: use donor mapping, color bridge crosses, antenna/pattern backcrosses, and size averaging to hit the exact target early, then spend remaining crosses on novel high-match offspring.} \\
\bottomrule
\end{longtable}
\begin{center}
\begin{minipage}{\linewidth}
\centering
\captionof{table}{\captionrevision{\revised{Butterfly: rule-discovery evidence. S: complete evidence in at least one trial; P: partial evidence only; --: insufficient evidence. Evidence may include episodes beyond the scoring window.}}}
\label{tab:rq1-targets-game-butterfly}
{
\appendixtable{%
\begin{tabular}{@{}l*{7}{c}@{}}
\toprule
\textbf{Configuration} & T1 & T2 & T3 & T4 & T5 & T6 & T7 \\
\midrule
\multicolumn{8}{@{}l}{\textbf{backbone comparison}} \\
Claude Opus~5 & S & S & S & S & S & S & -- \\
Qwen 3.8 Max & S & S & S & S & S & S & -- \\
Kimi~K3 & S & S & S & S & S & S & S \\
Gemini~3.5 Flash & S & S & S & S & -- & S & -- \\
Gemini~3.7 Flash & S & -- & S & S & S & S & -- \\
GPT-5.6-SOL & S & -- & -- & S & S & S & -- \\
DeepSeek V4 Pro & S & S & S & S & S & S & -- \\
Claude Sonnet~5 & S & S & S & S & S & -- & -- \\
GPT-5 Mini & S & -- & -- & -- & -- & S & -- \\
\midrule
\multicolumn{8}{@{}l}{\textbf{agent harness comparison}} \\
Claude Code + Opus~5 & S & S & S & S & S & S & S \\
Codex + GPT-5.6-SOL & S & S & S & S & S & S & S \\
\bottomrule
\end{tabular}
}
}

\end{minipage}
\end{center}

\subsection{\revised{Model Auditor}}
\label{app:game-modelkeeper_expert}

\revised{\textbf{Learning task.} Learn how to probe a new stateful machine and predict its outputs. Re-shuffled game, scored seed 1, first 5 episodes, reference score $C=100$.}
\begin{center}
\includegraphics[width=0.95\linewidth]{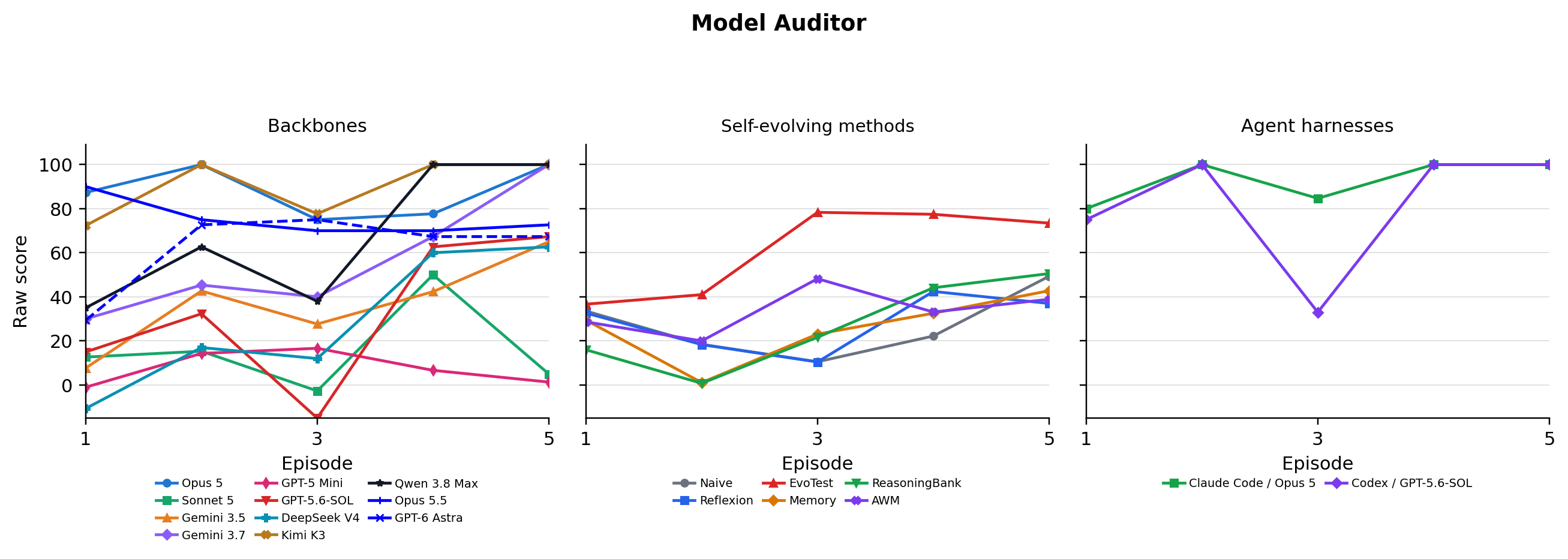}
\captionof{figure}{\revised{Model Auditor: raw-score learning curves within the scoring window.}}
\label{fig:game-curve-modelkeeper_expert}

\end{center}
\Needspace{10\baselineskip}
% Combined score table: normalized points and trial SD; human/First retained.
\begingroup
\setlength{\tabcolsep}{3pt}
\begin{longtable}{@{}lrrrrr@{}}
\caption{\captionrevision{Model Auditor: normalized performance and learning, with trial SD. Rows follow descending Max. First denotes the average episode at which a trial first reaches its highest score.}}
\label{tab:game-scores-modelkeeper_expert}\label{tab:trial-variability-modelkeeper-expert}\\
\toprule
Configuration & First & Max $\uparrow$ & Mean $\uparrow$ & LG $\uparrow$ & LS $\uparrow$ \\
\midrule
\endfirsthead
\toprule
Configuration & First & Max $\uparrow$ & Mean $\uparrow$ & LG $\uparrow$ & LS $\uparrow$ \\
\midrule
\endhead
Naive + Kimi K3 & 1 & $100.0 \pm 0.0$ & $54.5 \pm 15.7$ & $-1.2 \pm 28.4$ & $-1.93 \pm 5.69$ \\
OpenCode + Kimi K3 & 1.7 & $100.0 \pm 0.0$ & $90.0 \pm 9.2$ & $+13.8 \pm 15.1$ & $+5.53 \pm 6.05$ \\
EvoTest + Kimi K3 & 1.7 & $100.0 \pm 0.0$ & $79.7 \pm 8.7$ & $+33.7 \pm 21.8$ & $+7.77 \pm 9.07$ \\
EvoTest + Opus 5 & 1.7 & $100.0 \pm 0.0$ & $78.1 \pm 18.0$ & $+24.2 \pm 28.3$ & $+9.67 \pm 11.32$ \\
Claude Code + Opus 5 & 2 & $100.0 \pm 0.0$ & $92.9 \pm 3.8$ & $+10.0 \pm 4.3$ & $+4.00 \pm 1.73$ \\
OpenCode + Opus 5 & 2 & $100.0 \pm 0.0$ & $88.0 \pm 4.4$ & $-4.8 \pm 18.4$ & $+0.30 \pm 3.53$ \\
Codex + GPT-5.6-SOL & 2 & $100.0 \pm 0.0$ & $81.6 \pm 4.7$ & $+12.5 \pm 11.8$ & $+5.00 \pm 4.73$ \\
Memory + Kimi K3 & 2 & $100.0 \pm 0.0$ & $65.7 \pm 20.3$ & $+38.8 \pm 27.6$ & $+9.07 \pm 10.02$ \\
AWM + Kimi K3 & 2.3 & $100.0 \pm 0.0$ & $73.4 \pm 17.6$ & $+20.8 \pm 40.1$ & $+4.07 \pm 13.02$ \\
OpenCode + Qwen 3.8 Max & 2.7 & $100.0 \pm 0.0$ & $67.1 \pm 19.5$ & $+51.2 \pm 46.7$ & $+16.73 \pm 12.20$ \\
ReasoningBank + Kimi K3 & 3.3 & $100.0 \pm 0.0$ & $60.6 \pm 14.7$ & $+66.5 \pm 26.0$ & $+18.10 \pm 7.45$ \\
OpenCode + Gemini 3.7 Flash & 3.3 & $100.0 \pm 0.0$ & $56.5 \pm 38.3$ & $+46.0 \pm 46.8$ & $+16.20 \pm 14.86$ \\
\opusrevision{OpenCode + Opus 5.5} & \opusrevision{1} & \opusrevision{$90.0 \pm 17.3$} & \opusrevision{$75.5 \pm 42.4$} & \opusrevision{$-11.2 \pm 19.3$} & \opusrevision{$-3.97 \pm 6.87$} \\
Reflexion + Kimi K3 & 3 & $83.0 \pm 29.4$ & $46.3 \pm 21.2$ & $+38.5 \pm 8.3$ & $+9.13 \pm 4.63$ \\
OpenCode + DeepSeek V4 Pro & 3.3 & $77.7 \pm 38.7$ & $28.2 \pm 28.8$ & $+58.2 \pm 73.3$ & $+18.97 \pm 21.04$ \\
OpenCode + Gemini 3.5 Flash & 3.7 & $75.0 \pm 43.3$ & $37.1 \pm 30.9$ & $+28.5 \pm 19.2$ & $+11.43 \pm 8.75$ \\
\astrarevision{OpenCode + GPT-6 Astra} & \astrarevision{2} & \astrarevision{$75.0 \pm 43.3$} & \astrarevision{$62.3 \pm 47.2$} & \astrarevision{$+16.3 \pm 34.5$} & \astrarevision{$+7.07 \pm 13.29$} \\
EvoTest + GPT-5 Mini & 3.7 & $72.3 \pm 30.3$ & $26.3 \pm 26.3$ & $+52.0 \pm 38.7$ & $+15.56 \pm 8.33$ \\
Human Top-1 & 1 & 70.0 & 19.2 & -- & -- \\
OpenCode + GPT-5.6-SOL & 2.3 & $67.3 \pm 56.6$ & $32.5 \pm 34.9$ & $+41.3 \pm 45.0$ & $+13.50 \pm 12.99$ \\
Reflexion + GPT-5 Mini & 4.3 & $57.0 \pm 37.5$ & $22.6 \pm 24.0$ & $+9.2 \pm 32.1$ & $+0.40 \pm 12.02$ \\
OpenCode + Sonnet 5 & 2.3 & $55.3 \pm 41.5$ & $16.1 \pm 10.7$ & $+13.5 \pm 24.2$ & $+1.93 \pm 5.52$ \\
Naive + GPT-5 Mini & 4.7 & $54.0 \pm 39.8$ & $10.4 \pm 21.8$ & $+35.5 \pm 13.5$ & $+12.23 \pm 2.06$ \\
AWM + GPT-5 Mini & 4.3 & $51.3 \pm 42.3$ & $12.3 \pm 9.9$ & $+19.2 \pm 11.0$ & $+5.70 \pm 2.85$ \\
ReasoningBank + GPT-5 Mini & 4 & $47.0 \pm 33.0$ & $10.7 \pm 14.0$ & $+34.5 \pm 32.5$ & $+10.27 \pm 8.27$ \\
ReasoningBank + Opus 5 & 3.7 & $44.0 \pm 34.4$ & $8.5 \pm 7.9$ & $+15.8 \pm 23.1$ & $+5.37 \pm 8.53$ \\
Memory + Opus 5 & 1.7 & $40.3 \pm 38.7$ & $6.1 \pm 23.8$ & $+13.0 \pm 45.6$ & $+2.23 \pm 13.84$ \\
AWM + Opus 5 & 5 & $31.0 \pm 0.0$ & $15.6 \pm 0.0$ & $-5.0 \pm 0.0$ & $+0.30 \pm 0.00$ \\
Naive + Opus 5 & 5 & $31.0 \pm 0.0$ & $15.6 \pm 0.0$ & $-5.0 \pm 0.0$ & $+0.30 \pm 0.00$ \\
Reflexion + Opus 5 & 5 & $31.0 \pm 0.0$ & $15.6 \pm 0.0$ & $-5.0 \pm 0.0$ & $+0.30 \pm 0.00$ \\
Memory + GPT-5 Mini & 5 & $28.3 \pm 13.2$ & $5.7 \pm 5.1$ & $+15.3 \pm 8.6$ & $+6.13 \pm 2.31$ \\
OpenCode + GPT-5 Mini & 2 & $16.7 \pm 14.4$ & $7.6 \pm 7.2$ & $-2.7 \pm 2.4$ & $-0.30 \pm 0.36$ \\
\bottomrule
\end{longtable}
\endgroup
\begin{longtable}{@{}p{0.07\linewidth}p{0.87\linewidth}@{}}
\caption{\captionrevision{\revised{Model Auditor: criteria for assessing rule discovery.}}}\label{tab:targets-definition-model-auditor}\\
\toprule
\textbf{Target} & \textbf{Criterion} \\
\midrule
\endfirsthead
\toprule
\textbf{Target} & \textbf{Criterion (continued)} \\
\midrule
\endhead
T1 & \revised{Each episode has a new machine with ternary inputs. What transfers is the method for testing and modeling the machine, not the previous exam answers.} \\
T2 & \revised{The machine combines four mechanisms: GATE thresholds signed current inputs. MIX sums modulo 3. HOLD maintains a bounded up/down state. ECHO delays its input by one step.} \\
T3 & \revised{A two-input GATE/MIX component feeds a shared hidden HOLD/ECHO state. One output is a reversible MIX view of that state. The other uses GATE/MIX and may also depend on an additional state component.} \\
T4 & \revised{Reset between probe sequences. Single-input ramps or pulses, pairs of active inputs, and graded mixed inputs distinguish GATE from MIX and HOLD from ECHO.} \\
T5 & \revised{Fit both outputs jointly within the 40-probe budget. Fitting each output separately can miss the state they share.} \\
T6 & \revised{For full exam accuracy, fit a depth-3 machine consistent with the observed transcript and check that all consistent candidates predict the same outputs for the unseen ten-step exam. Submit every output without relying on feedback before the exam ends.} \\
\bottomrule
\end{longtable}
\begin{center}
\begin{minipage}{\linewidth}
\centering
\captionof{table}{\captionrevision{\revised{Model Auditor: rule-discovery evidence. S: complete evidence in at least one trial; P: partial evidence only; --: insufficient evidence. Evidence may include episodes beyond the scoring window.}}}
\label{tab:rq1-targets-game-model-auditor}
{
\appendixtable{%
\begin{tabular}{@{}l*{6}{c}@{}}
\toprule
\textbf{Configuration} & T1 & T2 & T3 & T4 & T5 & T6 \\
\midrule
\multicolumn{7}{@{}l}{\textbf{backbone comparison}} \\
Claude Opus~5 & S & S & S & S & S & S \\
Qwen 3.8 Max & S & S & S & S & S & S \\
Kimi~K3 & S & S & S & S & S & S \\
Gemini~3.5 Flash & S & S & S & S & S & S \\
Gemini~3.7 Flash & -- & S & S & S & S & S \\
GPT-5.6-SOL & -- & S & S & S & S & S \\
DeepSeek V4 Pro & S & S & S & -- & -- & -- \\
Claude Sonnet~5 & -- & S & S & S & S & S \\
GPT-5 Mini & S & S & -- & S & -- & -- \\
\midrule
\multicolumn{7}{@{}l}{\textbf{agent harness comparison}} \\
Claude Code + Opus~5 & S & S & S & S & S & S \\
Codex + GPT-5.6-SOL & S & S & S & S & S & S \\
\bottomrule
\end{tabular}
}
}

\end{minipage}
\end{center}

\end{document}